\documentclass[a4paper,11pt,oneside]{book}
\usepackage[english]{babel} 
\usepackage[utf8]{inputenc} 
\pdfoutput=1 

\usepackage{blindtext}
\usepackage{tocbibind}
\usepackage[colorlinks=false]{hyperref} 

\usepackage{color} 
\usepackage{amsfonts} 
\usepackage{amsmath}
\usepackage{amssymb}
\usepackage{amsmath} 
\usepackage{graphicx} 
\usepackage{geometry} 
\usepackage{acronym} 
\usepackage{float} 
\usepackage{parskip} 
\graphicspath{{./immagini/}}
\newcommand*{\fullref}[1]{\hyperref[{#1}]{\autoref*{#1} \nameref*{#1}}} 

\usepackage{fancyhdr} 
\usepackage[labelfont=bf,small]{caption} 

\usepackage{booktabs} 
\usepackage{subcaption}  
\usepackage[table,xcdraw]{xcolor} 

\usepackage{enumerate}

\begin{document}

\begin{titlepage}
\center

\begin{figure}[ht]
\center \includegraphics {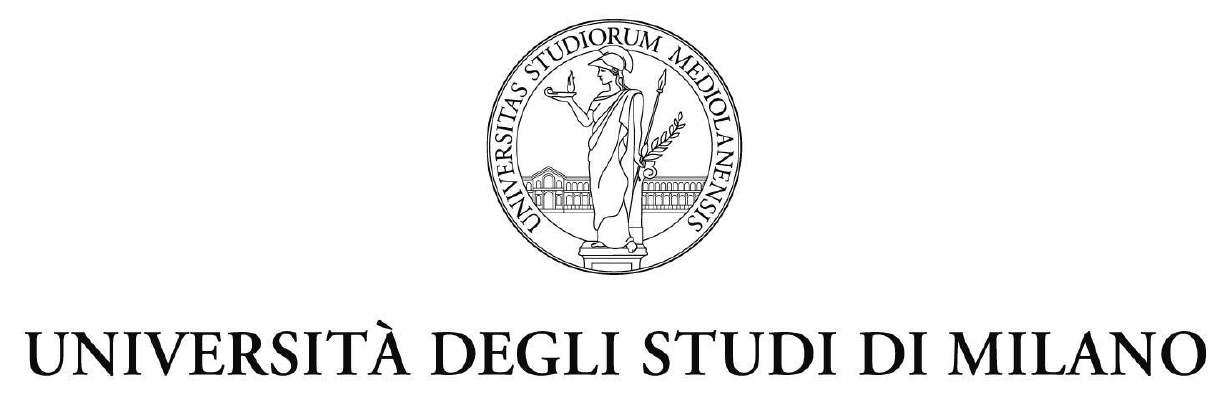}\\
\end{figure}
\vspace{0.5cm}

\begin{large}
\textbf{Faculty of Science and Technology}\\
\vspace{0.2cm}
\textbf{Master Degree in Molecular Biotechnology and Bioinformatics}
\end{large}

\vspace{4cm}

\begin{LARGE}
\textbf{Network-based methods for outcome prediction in the ``sample space''}
\end{LARGE}

\vspace{3.5cm}

\begin{flushleft}
\textbf{Advisor:\hspace{0.9cm}Prof. Giorgio VALENTINI {\small(University of Milan)}}\\
\textbf{Co-Advisor: Prof. Alberto PACCANARO {\small(Royal Holloway, University of London)}}\\
\textbf{\hspace{2.51cm}Prof. Matteo RE {\small(University of Milan)}}
\end{flushleft}

\vspace{2cm}

\begin{flushright}
\large
\textbf{Candidate:}\\
\textbf{Jessica GLIOZZO}\\
\textbf{Student ID 846946}
\end{flushright}

\vspace{3cm}

\begin{center}
\large
\textbf{ACADEMIC YEAR 2015-2016}
\end{center}

\end{titlepage}

\newpage\null\thispagestyle{empty}\newpage

\frontmatter

\tableofcontents
\cleardoublepage

\listoffigures
\cleardoublepage

\listoftables
\cleardoublepage


\chapter*{Summary}\label{Summary}
\markboth{\MakeUppercase{Summary}}{\MakeUppercase{Summary}}
\addcontentsline{toc}{chapter}{Summary}

Phenotype and clinical outcome prediction are common tasks in computational biology \cite{dupuy2007critical}. State-of-the-art methods are usually based on inductive supervised models, which exploit sets of biomarkers, e.g. gene expression signatures selected by feature selection methods, to predict the phenotype under study \cite{guyon2002gene}. However, recent works highlighted that the results of this approach are often not statistically significant \cite{venet2011most} or not reliable due to unsound experimental techniques adopted for the evaluation of the performances of the methods \cite{ambroise2002selection, michiels2005prediction}.  
Moreover these methods exploit the selected signatures to classify patients in different classes, but they do not consider the relationships between individuals.  
This is also true for semi-supervised methods that do not explicitly take into account the functional or the genetic relationships between patients.

Systems Biology analyses the relationships between the components of biological systems~\cite{Kitano15}, and in this context the new field of the \textit{Network Medicine} provides a new framework to study the molecular mechanisms underlying diseases \cite{barabasi2011network, Pacca15}.
According to this approach, we can study the biology of an organism through a graph structure, where nodes represent bio-molecules (e.g. genes, proteins) and edges represent functional or genetic relationships between the biological/biomolecular entities under study. In the last years, several network-based methods have been developed with the purpose of finding out novel biomarkers. These methods exploit the topological relationships between nodes in the biomolecular network and semi-supervised learning algorithms can infer novel candidate disease genes or can select markers for specific diseases \cite{valentini2014extensive, winter2012google}. 

In this thesis we consider the complementary ``sample space'' where the components of the system are the samples (representing e.g. the biomolecular profile of a patient) and the functional/genetic relationships between them. 
That is our Network Medicine-based approach moves from  the ``biomarker space'', which has been already  extensively studied in literature, to the ``sample space'' to exploit the advantages of graph-based methods for the analysis and prediction of the outcome, or phenotype or response to treatment of patients. 
Only a few previous works explored the ``sample space''~\cite{park2014integrative, wang2014similarity}, since both unsupervised and semi-supervised Network Medicine methods have been mainly applied to the ``biomarker space'' to discover e.g. genes associated with diseases or to discover proteins involved in specific metabolic networks~\cite{moreau2012computational}.

To both consider patients' specific biomolecular information and to exploit the representation of systems as networks in the ``sample space'', we developed \textbf{P-Net}, a novel semi-supervised network-based algorithm. We explicitly consider the functional/genetic relationships between individuals, while, at the same time, we can exploit the available “a priori” phenotypic information about them. The peculiar and innovative characteristic of this method is that P-Net constructs a network of samples (rather than bio-molecules), where the edges represent functional or genetic relationships between individuals (e.g. similarity of expression profiles) to predict the phenotype/outcome of interest. 

The proposed algorithm is able to provide a rank or a classification of the samples with respect to a specific property (e.g. phenotype, outcome or response to treatment), by exploiting the similarity between the biomolecular profiles of the patients and the topology of the networks in the ``sample space''.

To this purpose, we apply P-Net on three different publicly available datasets from patients afflicted with a specific type of tumour: pancreatic cancer, melanoma and ovarian cancer dataset, by using the data and following the experimental set-up proposed in two recently published papers~\cite{barter2014network, winter2012google}. 
We show that network-based methods in the sample space can achieve results competitive with classical supervised inductive systems.
Moreover the graph representation of the samples can be easily visualized through networks, and can be used to gain visual clues about the relationships between samples, taking into account the phenotype associated or predicted for each sample.

To our knowledge this is one of the first works that proposes graph-based algorithms working in the sample space of the biomolecular profiles of the patients to predict their phenotype or outcome, thus contributing with a novel research line in the framework of the Network Medicine.

\vspace{0.5cm}
This thesis has been organised in the following way. In chapter \ref{Intro} a detailed explanation of the outcome prediction problem and state-of-the-art approaches commonly employed to address this task are introduced. A brief description of the main features, and the position of our proposed method in the current research scenario is also provided. Chapter \ref{Methods} describes the theoretical background and implementation of P-Net, the network-based method that exploits the ``sample space'' to rank and classify patients with respect to the phenotype under study. 
Moreover, it presents a short overview about the feature selection methods, both in general and with a focus on the methods used in this work. Chapter \ref{Res} provides the experimental set-up followed to assess the performances of P-Net, and the presentation and discussion of all the experiments and results. Then we introduce some approaches to visualise the \textit{graph of patients} which is built by P-Net and we explain how this visualisation can be used to analyse the functional or genetic relationships between patients. Chapter \ref{Concl} concludes the thesis, by outlining the main contributions of this work and possible future research developments.

\mainmatter

\chapter{Introduction}\label{Intro}

In this chapter we introduce the problem of the phenotype/outcome prediction from biomolecular data, the main state-of-art methods proposed in this context and the positioning of this work with respect to the current research scenario.

\section{Phenotype prediction and P4 medicine} \label{outcome_pred}
Phenotype and outcome prediction are well-known problems in the context of computational biology, where the purpose is to build a model able to predict the phenotypic class of unseen samples. The concept of phenotype is really general and it is defined as: \textit{the observable expression of an individual's genotype as appearances, signs and symptoms of disease} \cite{wojczynski2008definition}. 
Among the categories of phenotypes, the clinical outcome of a patient afflicted with a specific disease and in particular the prediction of cancer outcomes are one of the most explored \cite{kourou2015machine, ressom2008classification} in literature. Moreover, outcome prediction is usually achieved using a set of selected biomarkers (e.g. gene expression signatures or allelic configurations of SNPs) \cite{dupuy2007critical, shipp2002diffuse, ye2003predicting} obtained by a suitable feature selection method, which is able to select the most significant features for the task. The selection of a gene signature is useful not only to create a classifier with better performances but also to obtain a list of genes, which should be the most discriminative for the specific outcome under study. In this way, we can identify the most important molecular characteristics related to a specific disease state and we can exploit them in different ways. For example, we can use a signature to develop a diagnostic kit to stratify the patients with different forms and degree of severity for the same disease and to provide to them personalised treatments. This is the case of the MammaPrint\textsuperscript{TM} assay that is a prognostic test for women under 61 years able to discriminate the patients between high-risk and low-risk of disease recurrence, where a low-risk means that the likelihood to develop metastasis for the patient is low. This microarray-based assay evaluates the gene expression levels of a 70-gene signature. This commercialised assay can spot the patients with a good prognosis so we can avoid to expose them to unnecessary toxic chemotherapy treatments \cite{slodkowska2009mammaprint}.

Outcome prediction is one of the problems considered by the new trend of \textit{P4 medicine}, which is a term that describes a system approach to medicine. It involves four principal essential features for medicine: \textit{predictive}, \textit{personalized}, \textit{preventive} and \textit{participatory}. The term ``P4 medicine'' was introduced for the first time more than 10 years ago and, over the time, it acquired a great importance because of the increment in the amount of different data sources available for a single patient and the development of new analytical technologies, which give us the possibility to decipher the data associated with each patient. There are different types of data we can retrieve for each individual, exploiting suitable technologies (e.g. DNA sequence, gene expression levels, epigenetic changes, level of proteins, protein-protein networks, etc.), and all these data can be integrated to model the various hierarchical levels of the biological information \cite{hood2011predictive, tian2012systems}.\\
In the context of phenotype prediction, the available data from a cohort of patients can be used to build a model (also called predictor or classifier). At a later stage, the predictor is used to assign a new sample, which was not used to build the model, to an already-defined class. There are manifold different phenotypes we can predict: response to a specific drug \cite{riddick2011predicting, ruderfer2009using}, prognosis \cite{chen2014risk, das2015encapp}, classification of cancer subtypes \cite{podolsky2016evaluation}, recurrence \cite{bartsch2016use, kim2012development}, to name a few. In all these situations, we can reduce our problem to a classification task where our goal is to assign to each patient the right \textit{class label}. In the simplest case, we want to say if a given sample belongs or not to the group of samples with the phenotype of interest. This is a binary classification problem, but it can be easily extended to a multi-class classification problem.  

\section{Machine learning methods}
The exponential increase in the amount and complexity of biological data available during the last decades is the reason why the machine learning algorithms are applied in computational biology. Indeed, machine learning methods give us the possibility to extract useful information from these data and also to exploit these data to obtain models of the system under study. Once we obtain a model, we are able to use it to make predictions on the system \cite{larranaga2006machine}. 
Machine learning (a branch of the Artificial Intelligence field) develops algorithms to learn from the data in order to infer new knowledge from the information available for the phenomenon under study. Machine learning methods are roughly divided into the three following subtypes:

\begin{itemize}
\item \textbf{Supervised learning methods}\\ 
The learner receives as input a set of \textit{n} labelled examples, which are the training data:

\begin{equation}
\lbrace (x_{1},y_{1}),..., (x_{n},y_{n}) \rbrace
\end{equation}

where $x_{i}$ is the vector of features of the $i^{th}$ example and $y_{i}$ is the corresponding label. The goal of a supervised method is to fit a model that relates each example, which is characterised by its vector of features, to the right label. In other words, we search for a generic function:

\begin{equation}
g: X \rightarrow Y
\end{equation}

where $X$ is the input space containing all the feature vectors $x_{i}$ and $Y$ is the output space of the labels $y_{i}$.\\
Once we obtain the model, it can be used to make predictions for all the unseen observations that are the examples not used to train the model.

\item \textbf{Unsupervised learning methods}\\ 
The learner receives as input a set of \textit{n} examples $(x_{1},..., x_{n})$ that have not an associated label $y_{i}$. Basically, the training set consists of a vector of features $x_{i}$ for each observation $i^{th}$. We can refer to this kind of situation as \textit{unsupervised} because we lack the vector of \textit{true labels} that can supervise our analysis. In this case, we search for the relationships between the variables or between the observations with the purpose of finding out structures and patterns that characterize the phenomenon under study.

\item \textbf{Semi-supervised learning methods}\\ 
There are various problems that fall into the supervised or unsupervised learning paradigms. However, it is difficult sometimes assign a problem to one of these categories. Indeed, there are situations where we have a set of \textit{n} examples and only for \textit{m} of these examples (with $m<n$) we have the vector of features $x_{i}$ and the corresponding label $y_{i}$. For the remaining $m-n$ examples we have the feature vector but we do not have the corresponding label. We refer to this situation as \textit{semi-supervised} because we have just some \textit{true labels} to supervise our analysis. In this case, the learner tries to exploit both labelled and unlabelled data to make predictions for the unseen observations.
\end{itemize} 

So, it is straightforward that the aforementioned subtypes of algorithms differ mainly in the type of training data available for the training phase of the predictor \cite{james2013introduction, mohri2012foundations}.\\
Moreover, the tasks solved using machine learning techniques can be categorized into the following three classes with respect to the output of the learner:
\begin{enumerate}[a)]
\item \textbf{Classification:} refers to a problem that involves a qualitative output (in other words, categories or classes) and the learner splits the data into a set of finite classes.

\item \textbf{Regression:} refers to a problem that involves a quantitative output and, for each new sample, the learner predicts the quantitative value of a variable. 

\item \textbf{Clustering:} the initial observations are split into relatively distinct groups and the observations in the same group share some similar characteristics. This is a typical unsupervised problem because we do not know \textit{a priori} which are the groups. 
\end{enumerate} 

The classification task includes the outcome prediction problem, which is discussed in the precedent section \ref{outcome_pred}.

\subsection{Supervised and unsupervised learning methods} \label{sup_unsup}
There are several supervised learning methods commonly applied to outcome prediction with remarkable results (e.g. Artificial neural networks, Bayesian statistics, Decision tree learning, Support Vector Machines, Random Forest, etc).\\  
For example in the paper \cite{ahmad2013using}, the authors apply three different supervised machine learning methods (SVM, DT and ANN) to a cohort of 547 patients afflicted with breast cancer, where for each patient a series of clinical variables is collected. The 2-year recurrence rate of the patients, under a 10-fold cross-validation framework, is predicted to find which is the best classifier on this specific task. This work shows that the best model is the SVM, which achieves the highest sensitivity, specificity and accuracy. However, all the methods obtain interesting results with an accuracy of 0.936, 0.947 and 0.957 for DT, ANN and SVM respectively.

It is important to note that state-of-the-art methods are largely based on supervised models that use lists of selected biomarkers, usually represented as vectors, to predict the phenotype or outcome of interest and several works clearly showed the effectiveness of these methods (see, e.g. \cite{colombo2011microarrays}, \cite{guyon2002gene}, \cite{wang2008hybrid}).
The term biomarker stands for ``biological marker'' and the National Institutes of Health Biomarkers Definitions Working Group defined a biomarker as \textit{a characteristic that is objectively measured and evaluated as an indicator of normal biological processes, pathogenic processes, or pharmacologic responses to a therapeutic intervention} \cite{strimbu2010biomarkers}. The biomarker selection is a particular subtype of feature selection problem (see section \ref{feat_sel} for more details about the feature selection methods) where the purpose is to find the most significant features to distinguish between specific conditions (e.g. good prognosis and poor prognosis in patients afflicted with a specific type of cancer). We expect that if we apply different biomarker selection methods on the same dataset they should find the same set of biomarkers or, at least, there should be a significant overlap between the lists of biomarkers selected by different methods. Nevertheless, some studies recently pointed out that these results are in several cases not statistically significant \cite{venet2011most} or unreliable for the unsound experimental techniques adopted for the performance evaluation \cite{ambroise2002selection, michiels2005prediction} in the same datasets and across different datasets \cite{bernau2014cross}. In this connection, in the paper \cite{venet2011most} the authors tested the association with breast cancer outcome of 1000 random signatures and 47 published signatures. The results show that most of the published breast cancer signatures are not more strongly associated with breast cancer outcome than sets of random genes. The issue of the stability of feature selection methods is becoming pressing and only a few works take into account this problem \cite{dessi2013comparative, haury2011influence}. 

Another drawback of the supervised methods is that they do not explicitly take into account the functional or genetic relationships between individuals, in the sense that they use vectors of selected biomarkers to discriminate between patients and do not directly exploit the set of relationships between them. A possible solution is the use of a supervised learning algorithm with a prior causal network defined from literature \cite{zarringhalam2014robust}. This approach successfully incorporates causal prior knowledge and achieves robust performances across independent datasets. 

On the contrary unsupervised methods usually explicitly consider the relationships between samples.
e.g. clustering of patients on the basis of their expression profiles \cite{golub1999molecular}, but in that case the available \textit{a priori} information about the phenotype is lost, thus resulting in a significant decay of performances. 

Traditionally, the outcome prediction task from biomedical data is accomplished by supervised learning methods. The chosen computational algorithm searches for patterns among the features that model a specific outcome variable. The purpose is to build a classifier that maps the input data (features) to the output data (outcome variable) of interest. A domain expert is required to define the specific learning task, to decide the features and samples to use as input and prepares the outcome values for training and testing the model. However, this approach shows some weak points:

\begin{enumerate}[a)]
\item It has limited capacity to generalize beyond individual models developed to predict or explain pre-specified outcomes.

\item The selection of the features used to train the model is very important to determine the accuracy of a supervised method and also the step of preparation of the outcome variables, which are usually binary classes. However, these selection steps are time-consuming and challenging. 

\item Another limitation of supervised methods for outcome prediction is that they find only the patterns we choose to look for through the definition of inputs and output variables. Basically, supervised methods are a good choice if we want to find a model to predict phenotypes for which we know enough to label in advance.
\end{enumerate}

All these limitations of supervised methods are overcome when we use unsupervised methods, which can discover phenotypes from data without \textit{a priori} information \cite{lasko2013computational}. Nevertheless, if some \textit{a priori} information is available, we are not able to exploit them using this kind of machine learning technique. In the phenotype prediction context, unsupervised methods are often exploited to cluster patients in different groups with common biological characteristics and to understand if there are new different phenotypes for the same disease \cite{guan2016unsupervised, siroux2011identifying}. In this way, we can discover new latent phenotypes emerging from the data instead of continuing to use predefined classifications. The employed features are usually clinical data and the purpose is to highlight the heterogeneity of the disease and to characterise better each individual. The knowledge about patients' specific characteristics is helpful to develop personalised and more effective treatments. 

\subsection{Semi-supervised methods} \label{SSL}
Semi-supervised learning (SSL) algorithms were developed to address the necessity to exploit both labelled and unlabelled data in the training phase of a classifier. More in-depth, this kind of machine learning algorithms use the unlabelled data to either modify or reprioritise hypotheses obtained from the labelled data alone.\\ 
Supervised learning algorithms employ only the labelled data to train a predictor and, following this approach, we loss the unlabelled data that are usually more than the labelled ones in many situations. Indeed, labelled data are often difficult, expensive or time consuming to collect because they require the effort of human annotators \cite{zhu05survey}. For example, we can imagine we want to build a classifier to predict the prognosis of patients afflicted with a specific disease based on gene expression data. We need the expression profile of each sample and its time of survival to label each individual (e.g. patient with a time of survival higher than 5 years are labelled as ``good prognosis'' otherwise the label is ``poor prognosis''); these are the input data required to train the predictor. Implicitly, we assume that each patients had a follow-up of at least 5 years but this is not always true. To the end of obtaining a 5-year follow-up, oncologist consultation fees must be paid to confirm survivability and a confidentiality agreement is required to exploit patient's information. Moreover, censored data are common in survival analysis because when the patients' data are not updated recently, they remain unlabelled \cite{kim2013breast}.

The following question arises when we are considering to use an SSL method to address a classification problem: \textit{Does the addition of unlabelled data always lead to an improvement in the performances or it can lead to their degradation?}\\
This is a not trivial question but some general tips can be useful to understand when the application of a semi-supervised learning method can be the better choice. The general rule is that the distribution of the examples, which the unlabelled data will help to elucidate, has to be relevant for the classification problem. In other words, if we have a series of examples $x$ and some of them has an associated label $y$, the semi-supervised learning yields to an improvement over supervised learning only if the knowledge on $p(x)$, which one gains through the unlabelled data, carries useful information for the inference of $p(y|x)$. If this is not the case, it might even happen that using the unlabelled data degrades the prediction accuracy by misleading the inference. It is important to note that labelled data are exponentially more effective in reducing classification error than unlabelled data and they always improve the performances.\\
In this context, we have to consider also the so-called \textit{Finite Sample Effects}. There are many situations where only a few samples are available to estimate the predictor. In this case, especially if we have manifold attributes, the estimated predictor is likely to have an high variance and an high classification error. Furthermore, the variance becomes bigger with the increase of the number of parameters to estimate. At this point, the addition of unlabelled data is a positive move because it leads to decrease the variance of the estimator and to improve the classification performances \cite{chapelle2006semi}.

A growing trend is visible in the studies published in the last years, which applied SSL techniques for modelling cancer survival and, more in general, for predicting cancer disease states. It has been proved that they enhance the estimated performance compared to existing supervised techniques and, coupled with other approaches (e.g. decision trees \cite{shin2014coupling}), it can provide more interpretable results.\\
The paper \cite{shi2011semi} shows a successful application of SSL methods for the outcome prediction task. The authors develop and evaluate a semi-supervised classifier for human cancer, called LDS (low density separation) algorithm, and they compared it with respect to the supervised method SVM. The algorithm was employed to predict the 5-years recurrence of patients from five colorectal cancer datasets and two breast cancer datasets. They proved that the integration of unlabelled data, either from the same dataset or different datasets, enhances the classification performances with respect to SVM and the results improve with an higher size of unlabelled data. Moreover, the LDS method presents the possibility to use unlabelled data from 
different institutes which is a way to solve the small sample size issue in microarray data (especially if compared with the high number of features) that can lead to over-fitting.\\ 
Another application of an SSL method is presented in the paper \cite{halder2014semi}, where a novel semi-supervised fuzzy K-NN method is applied for classifying the patients from five different cancer datasets. The datasets contain the gene expression profiles of patients afflicted with a brain tumour, colon cancer, leukaemia, lymphoma and prostate cancer; brain tumour and lymphoma present multiple classes. The proposed approach is compared with the supervised counterparts K-NN and fuzzy K-NN classifiers but also with SVM and Naive Bayes classifiers. In all the cases the semi-supervised algorithm achieves the best accuracy confirming the potentiality of the SSL methods.\\
In addition, semi-supervised clustering methods are developed to identify biologically relevant clusters associated with a given outcome variable (e.g. survival time) exploiting both genetic and clinical outcome information. We have to remember that unsupervised learning is the approach commonly applied to identify clusters, but this kind of algorithm cannot use the outcome variables information and may fail in the discovery of clusters associated with the outcome, finding clusters unrelated to it \cite{bair2013semi}. A typical example of semi-supervised clustering procedure involves two steps. First, we exploit the outcome variable of interest to select the features that are strongly associated with the variable and then, there is the application of the chosen clustering method to the selected features subset \cite{koestler2010semi}. 
  
\section{Machine Learning applications to Network Medicine} \label{network}
\textit{\textbf{Network-based methods}}, which belong to the rising field of \textit{\textbf{Network Medicine}}, are employed to carry out a considerable part of the studies related to the outcome prediction task and we consider both the topics in the current section of this thesis. 

The term \emph{Network Medicine} was presented for the first time by Albert-László Barabási in a paper published in ``The New England Journal of Medicine'' in 2007 \cite{barabasi2007network}. This relatively new field is rooted in the assumption that networks pervade all the aspects of human health and we have to consider them to better understand the biological mechanisms underlying every aspect of an individual. This premise implies that a disease phenotype is seldom related to a single gene product but it reflects the effect of multiple processes, which interact in a complex net. Hence, an organism is described through the so-called \textit{human interactome}, a huge network involving the whole set of molecular interactions in a specific cell. The nodes represent cellular components (e.g. genes, proteins, metabolites, etc.) and the edges represent interactions between cellular components, both physical and indirect connections. The complexity of this network is outstanding: the number of nodes is more than 100,000 and the number of functionally relevant links is expected to be much larger. However, in the interactome we can identify more specific kinds of sub-networks depending on the nature of the considered nodes and edges:

\begin{itemize}
\item \textbf{protein-protein interaction networks:} where the nodes represent proteins connected through physical interactions;
\item \textbf{regulatory networks:} describe the relationships between transcription factors and genes or post-translational modifications; 
\item \textbf{metabolic networks:} the nodes are metabolites and an edge between two metabolites means that they are involved in the same biochemical reaction; 
\item \textbf{RNA networks:} capture the role of non-coding RNAs in the regulation of gene expression;
\item and so on. 
\end{itemize}

All these networks have their natural representation in \textit{graph structures} and we can employ the tools provided by the \textit{graph theory} to study them.
 
We have to consider the level of accuracy of the maps we want to employ when we decide to use a network-based approach. Indeed, a remarkable problem in this kind of methods is that the interactome network is still incomplete and noisy even if, in the last years, there is a systematic effort to improve the quality of the available maps \cite{barabasi2011network}.\\
Beyond this limitation, network-based methods became really popular and many papers started to propose new algorithms based on this framework. The reason underlying the spread of these methods is related to the documented advantages that they can provide in the research areas of ``outcome prediction'' and ``biomarker selection''.\\
Even if an extensive analysis of the literature about network-based methods is out of the goals of this work, in the remaining part of the current section we provide few relevant examples that highlight their strong points.

\subsection{Network-based methods in the feature/biomarker space} 
A relevant example is the NetRank algorithm \cite{winter2012google}, which is a semi-supervised network-based method able to exploit both expression and network information to rank genes with respect to their prognostic relevance (see section \ref{summary} for further details about the method and the experimental set-up followed in this work). In the paper, they exploited the gene expression profiles from patients afflicted with pancreatic cancer and a transcription factor-target network to select seven prognostic biomarkers for the outcome under study. There are some remarkable improvements in this work. The signature derived by NetRank is more predictive than classic clinical prognostic factors. An interesting feature of NetRank is that it evaluates the relevance of a gene as biomarker considering not only its own expression but also the expression level of its neighbours in the network. In this way, NetRank can spot and avoid the markers that are related to the outcome by chance or noisy measures, but that are not due to an underlying biological causality. In other words, network-based methods can identify biomarkers that are probably more truly relevant for a given outcome. However, the use of \textit{a priori} information to select more robust and biologically significant signatures leads to a bias in the procedure: NetRank tends to promote the selection of genes highly connected in the net over the other ones. Hence, the highly characterised genes are preferred during the selection because they present more connections in the net but this is not necessarily related to their biological relevance. \textit{Roy et al.} \cite{roy2012network} evaluated the performances of NetRank on 25 different cancer datasets from 13 kinds of tumour (e.g. breast cancer, lymphoma, lung cancer, melanoma, ovarian cancer) in comparison with some traditional feature selection methods (t-test, fold change and random). The study shows that NetRank achieves better performances with respect to the random selection of signatures in all the datasets and it is better than t-test and fold change in 23/25 datasets. So, the network-based outcome prediction is reliable on different kinds of cancer and network information can improve the classification accuracy. Moreover, in this work the overlap of the signatures published by the different authors of the datasets is assessed. There is no overlap between the published signatures, even if we consider the same kind of tumour, but there is a significant overlap between the network-based signatures and this is a prove that network information allows the selection of more consistent signatures.\\
In the work of \textit{Rapaport et al.} \cite{rapaport2007classification}, it is showed that the use of network information, which considers \textit{a priori} knowledge about the gene interactions, leads to better classification performances and it improves the interpretability of the results. Indeed, the results of experiments on microarray data are often summarised by a list of genes, which are differentially expressed between two conditions or phenotypes. It is necessary to interpret these signatures from a biological point of view and there are many methods able to achieve this goal \textit{a posteriori} (e.g. based on the identification of the pathway or biological process that involves the genes).
A different approach to the same problem takes into account gene expression networks and the assumption is that genes close in the net have probably a similar expression. This can allow to remove part of the noise from the measures by the extraction of the \textit{low frequency} components from the gene network. In that paper, they use two gene expression datasets from S. cerevisiae cultures, where the samples in the first dataset are not irradiated and the samples in the second one are exposed to low irradiation doses. The gene network data employed to carry out the analysis derive from the KEGG database of metabolic pathways. The approach is based on the spectral decomposition of the gene expression measures considering the network as a graph. The following step is the attenuation of the high-frequency components in the gene expression vectors, related to the topology of the network. At a later time, the application of unsupervised clustering algorithms and supervised classification algorithms leads to classifiers that are easily interpretable in terms of pathways.

The intrinsic nature of the biological systems also justifies the decision to exploit network data. These systems are dynamics and they evolve over time to respond to the genetic and environmental changes. We cannot avoid to consider the dynamic nature of biological systems to understand how they work. To achieve this goal, it is possible to carry out an experimental mapping of the networks under different conditions, times and species. We can apply the so-called \textit{differential mapping} where the stronger and more relevant differential interactions are the interactions with the highest change between the explored conditions. The genetic differential interactions are the consequence of the relevant cellular processes in the considered condition \cite{ideker2012differential}.\\
Many works pointed out that the alterations in the physical interaction networks can be powerful prognostic indicators for breast cancer. Among these works there is the paper of \textit{Chuang et al.} \cite{chuang2007network} that shows some potential advantages derived from the use of network-based methods in general:

\begin{enumerate}[a)]
\item the markers are organised in sub-networks and they can provide models of mechanisms underlying the development of metastasis; 

\item we can identify biomarkers which cannot be spot by the traditional methods but we know are related to metastasis; 

\item the sub-networks are more reproducible across different cohorts of breast cancer than gene signatures identified without network information; 

\item there is an improvement in the prediction accuracy when we carry out a cross-datasets analysis with respect to individual genetic markers obtained by classic methods. 
\end{enumerate}  

There is also the development of methods, as ENCAPP \cite{das2015encapp}, that combines the human protein interactome with gene expression data to predict the prognosis of different kinds of cancer. This method recognises functional modules which are differentially expressed in patients with good and poor prognosis. Then, it exploits the modules to build a regression model able to predict the prognosis of patients. The method is also able to identify prognostic biomarkers.

Other \textit{graph-based semi-supervised methods} have already been employed with success for different tasks (e.g. disease gene prioritisation, biomarker discovery and outcome prediction) in literature~\cite{valentini2014extensive}.\\
In the work of \textit{Cai et al.} \cite{cai2011bassum}, the authors present a new Bayesian semi-supervised algorithm, called BASSUM, to select features with important classification information taking into account the lack of labelled samples in the datasets. The authors evaluated the performances of BASSUM on real and synthetic datasets, both partially labelled, and they assessed also some supervised and semi-supervised feature selection algorithms (including Spectral, HITON and SVM-RFE). The classification accuracy was evaluated under the 10-fold cross-validation framework using different classifiers for each feature selection method tested. This work shows that generally, BASSUM achieves the highest classification accuracy with the smallest number of selected features. Hence, this approach can minimise the redundancy of the selected subset of features. Furthermore, a series of experiments was performed to assess the impact of the number of unlabelled and labelled data on the classification accuracy, simply changing their amount. The classification accuracy of BASSUM increases with the number of unlabelled data in all the datasets, which is the expected behaviour for a semi-supervised learning method and it proves that BASSUM exploits the unlabelled data.\\ 
In the work of \textit{Nguyen et al.} \cite{nguyen2012detecting}, a new SSL algorithm, which uses different kinds of biological data to unravel disease genes, is proposed. An interesting point is the integration of different kinds of data, which are used as features to predict the label of the unlabelled genes in an SSL framework. The authors integrate the data from six different databases that provide genomic, proteomic and topological features of the genes connected in a chosen human PPI network. In the network, the nodes are proteins and the edges are interactions between proteins while the labelled data are the gene products known to cause diseases (or known to be not associated with any disease) and the remaining genes represent the unlabelled data. Eight different features are extracted for each node from multiple public databases as binary, numerical or categorical data. Binary and categorical data are transformed and normalised in numerical data using specific score functions for each kind of feature and finally, the labelled and unlabelled data are exploited to build a classifier. The learning problem is formulated in terms of a Gaussian random field on the weighted graph representing the data, where the mean of the field is characterised in terms of harmonic functions. The output of the model is a set of new putative disease genes because the algorithm assigns to the unlabelled data one of the two possible labels ``disease gene'' and ``non-disease gene''. The SSL algorithm is compared with other two supervised methods (k-NN and SVM) and it achieved better performances in general. Then, the accuracy of SSL and k-NN methods is evaluated on datasets with a different number of labelled data and the SSL achieves the best results when the number of labelled data is limited. Moreover, the error rate of the SSL method with a different combination of data features is evaluated and it clearly shows that the combination of all the features provides the best results. Summarising, graph-based SSL methods are particularly suitable to predict disease genes when the labelled data are limited, they are competitive with the supervised methods and they provide a framework to integrate different kinds of biological data to improve the prediction performances.

It is important to note that \emph{data integration} is a key step when we want to discover the associations between genotype and phenotype. With the development of a series of high-throughput omic measurements, we have the possibility to exploit a huge amount of different data (e.g. gene expression from microarray and RNA sequencing; epigenetic variations from methylation arrays, methylation sequencing or chromatin immunoprecipitation followed by sequencing (ChIP-seq); protein variation by metabolomic or proteomic studies). We can consider the different types of data independently but their combination can compensate for missing or unreliable information at every single level and it can reduce the number of false positives in the results if there are many sources of evidence pointing to the same gene or pathway.\\
One of the strategies to integrate multiple data types is the \textit{transformation-based integration} \cite{ritchie2015methods} where each type of data is transformed into an intermediate form (e.g. a graph or a kernel matrix) and then, the data can be merged. The combined dataset is then exploited to elaborate a specific model and make, for example, outcome predictions or prioritise disease genes. \textit{Graph-based semi-supervised learning} methods are one of the ways to carry out the transformation-based integration and this approach has the following advantages:

\begin{enumerate}[a)]
\item it preserves the properties of the specific types of data when they are transformed into an intermediate representation;
\item it can be used to integrate many kinds of data (e.g. categorical, numerical, sequences) but they have to share a unifying feature; 
\item it is robust to different data measurement scales.
\end{enumerate} 

The use of network-based methods provides automatically a way to represent the data as graphs $G=\langle V,E \rangle$, where the set of vertices $V$ corresponds to the components (e.g. genes, proteins) and the set of edges $E$ represent the functional relationships between nodes (e.g. correlation between expression profiles, physical interactions between proteins, etc.). The graph structure of the data gives us an easy way to visualise the data through tools like \emph{Cytoscape} \cite{shannon2003cytoscape} and \emph{NAVIGaTOR} \cite{brown2009navigator}, which allow the visualisation of protein-protein networks and the analysis of the network by a user-friendly interface. These tools are useful also to develop and interpret cancer signatures \cite{fortney2011integrative}.

\subsection{Network-based methods in the sample/patient space}
Recently there is an increasing effort in the study of methods based on \textit{patient-patient networks}, which can offer novel insights for understanding cancer \cite{kim2016understanding}, and our proposed algorithm falls exactly in this category of approaches. P-Net builds the network starting from sample-specific molecular information but the nodes of the resulting net are not cellular components but samples in general (if we consider as the phenotype of interest some clinical outcome each sample correspond to one patient in the initial dataset). In this ``patient space'', the edges of the network are functional or genetic similarities (e.g. similarity between their expression profiles or allelic configurations of SNPs).\\
The construction of patient-patient networks can be used, for example, to consider cancer heterogeneity and the presence of multiple molecular subtypes in the same tumour. An example in this context is in the paper \cite{cho2013dissecting}, where a new probabilistic model is proposed to unravel the relationships between genotype and disease subtypes with the purpose of handle the heterogeneous nature of tumours. In this approach, the individual's representation is a mixture of subtypes, which are not predefined but discovered by the model building. The probabilistic model finds the genetic alterations that are the cause of the different subtypes and, more in general, of the similarities between patients. The patient network is obtained considering as edges the Pearson correlation coefficient between the gene expression profiles of patients. Then, a set of features (mutations, gene CNVs, miRNA dysregulation) is associated with each patient and it is exploited to explain the phenotypic similarities in the network. Moreover, these data allow the identification of the cancer subtypes where the distribution of the probabilities of the attributes defines each of them (in other words, there is a specific model for each subtype). So, a patient has some probability to belong to each subtype depending on the values of its attributes, which are a mixture of the genetic features belonging to each subtype. The final probabilistic model is the integration of the subtype models and, at the end, a link between two patients is defined by the similarity of the probabilistic subtype assignment between them. This approach guarantees that patients connected by an edge in the network have similar underlying features and that the subtypes definition is the probability distribution of these features. This method successfully found some already known subtypes in a dataset from samples of glioblastoma multiforme and the corresponding features, which explain the subtypes. The method is able to find not predefined subtypes and to unravel novel relationships between attributes and subtypes. Another interesting network-based approach, which considers the cancer heterogeneity issue, is showed in the paper \cite{jahid2014personalized} but in this case, the purpose is the clinical outcome prediction of patients. Assuming that each patient belongs to a different subtype, it is possible to construct multiple, personalised, predictive models for each individual belonging to the network. The patient-patient graph is built using the Euclidean distance between patients' gene expression profiles, which is followed by the application of random walk kernel with restart to compute the similarity between patients. Then, there is the evaluation of the models for their ability to predict the outcome of the specific individual (in this case, the development of metastasis in patients having breast cancer) and we retain only the best models. Finally, a new patient can be classified exploiting only the predictive models obtained from patients with similar molecular characteristics compared to the new patient. The final decision results from the predictions of all the selected models through a committee classification approach. This method can handle the heterogeneity of the disease without the use of prior pre-defined subtype information and it reaches good performances in both intra-dataset and cross-dataset predictions. 

Beyond the use of networks in the ``patient space'' to consider the intrinsic heterogeneous nature of cancer, recent works showed the opportunity to use this kind of graphs as a powerful way for the integration of multiple types of data. The goal is to improve the prediction accuracy of disease phenotypes. We can find an interesting application in the paper \cite{wang2014similarity}, where the similarity network fusion (SNF) approach is presented to integrate different kinds of data by the construction of a different network of samples for each kind of data based on similarity measures between patients. The chosen data types are gene expression, DNA methylation and miRNA expression data. Then, the different networks are fused into a single similarity network, using a nonlinear combination method, and the merged net can capture both shared and complementary information from the different kinds of data. This approach can effectively identify subtypes by the application of clustering methods (like spectral clustering) on the net and it reduces the noise by the aggregation of the data. Moreover, the data integration with SNF has better performances compared to some popular integrative approaches and a network-based approach that uses individual sources of data. The method addresses the outcome prediction task in two steps: the first step is the construction of the fused network with SNF and then, a label propagation approach is used to predict the label of a new patient. The work \cite{kim2012synergistic} presents a graph-based semi-supervised method to integrate multiple genomic information (CNA, methylation, gene expression and miRNA) and, at a later stage, to predict the clinical outcomes in brain cancer and ovarian cancer datasets. This method has the advantages to be scalable but also to preserve the level-specific properties from the different sources of data. The first step is the construction of different patient networks from each considered data type, where the edges represent the similarity between two patients. Then, the multiple graphs are integrated into a single network by finding the linear combination coefficient for the individual graph. In the two evaluated datasets, the authors exploit the integrated data to predict five sets of binary classification problems. A semi-supervised classifier is applied on the graph and it returns as output a vector with a real value for each sample to predict, which is transformed in a label choosing a proper threshold. Both the individual graphs and the integrated graphs are used to make predictions for comparison purpose. The results show that the integration of multiple levels of information leads to better performances for the outcome prediction and that the employment of a patient-patient network is a good framework for data integration and outcome prediction task. Moreover, in the paper \cite{park2014integrative} the authors develop a novel graph-based SSL algorithm for cancer prognosis. It exploits the integration of PPI data and gene expression data to obtain a network of samples. After the data integration, there is the application of an SSL algorithm that applies graph regularisation to predict the labels of the unlabelled samples. The method is applied to the prediction of cancer recurrence on three different cancer datasets and it improves the prediction accuracy compared to supervised (SVM, Naive Bayesian, RF) and semi-supervised methods (TSVM - the semi-supervised version of SVM) commonly applied in the state-of-the-art.\\
The use of a patient-patient network is possible also without the necessity of data integration for the cancer classification tasks and, for example, the network can derive directly from the patients' gene expression profiles \cite{tran2015normalized}.        

\section{The novel P-Net algorithm in the context of state-of-the-art network-based methods}
In this work, we develop and assess the performances of a novel network-based algorithm called \textbf{\textit{P-Net}}. This \textit{semi-supervised network-based} method can both prioritise and classify the samples with respect to a specific phenotype of interest. This method is designed to carry out the \textit{clinical outcome or phenotype prediction task} and to this end, it constructs a \textit{network of patients}, where each node corresponds to an individual and each edge represents functional relationships between patients. More precisely it constructs a network in the ``sample space'' considering the functional or genetic similarities between samples and then applies semi-supervised techniques based on the kernelization of the sample space to exploit the overall topology of the network in the prediction process~\cite{Vale16a}. 
We can exploit different kinds of data to build the network but in our work we focus on gene expression data obtained by microarray technology and we have a vector representing the gene expression profile for each sample. Hence, we can build the edges of the network by the computation of the correlation between the expression profiles of pairs of patients. Then, we apply on the ``sample network'' the semi-supervised algorithm based on the recently proposed \textit{Kernelized score functions} \cite{Vale16a} to compute a score for each sample, which allows us to rank the patients according to their likelihood to show the phenotype under study. The kernelized score functions assign to each sample a real number exploiting the weight of the edges that connect the specific sample to its neighbours and the overall topology of the underlying graph (see chapter \ref{Methods} for a detailed presentation of the P-Net algorithm).

The main advantages of the proposed approach can be summarised as follows: 

\begin{itemize}
\item It is not based on the selection of discriminative biomarker signatures that are often not statistically significant \cite{venet2011most} or unreliable due to the unsound experimental techniques adopted for the evaluation of the performances \cite{ambroise2002selection, michiels2005prediction}.

\item As all the semi-supervised methods, it can exploit both labelled and unlabelled data \cite{james2013introduction, mohri2012foundations}. This is an advantage because the unlabelled data are easier to collect \cite{kim2013breast, zhu05survey} and they can improve the classification performances \cite{halder2014semi, shi2011semi}, especially when we consider datasets containing few samples and several features \cite{chapelle2006semi}. This is the most common situation we can find in the datasets containing gene expression profiles obtained by microarray technology. In this kind of data, there is usually a large difference between the number of samples and the number of features and the accuracy of a predictor tends to decrease in this context. We can refer to this situation as ``curse of dimensionality'' \cite{wang2008approaches}. A possible solution is the use of the unlabelled data, together with the labelled ones, to reduce this difference and to provide more information for the classification. 

\item There are studies (some of them presented in the section \ref{SSL}) that show how the use of semi-supervised learning can improve the classification accuracy of outcome prediction tasks with respect to the supervised methods, which are traditionally employed in this context. 

\item P-Net is a network-based method so it can consider not only the \textit{a priori} phenotypic information retrieved from the clinical label of the data but, at the same time, we can explicitly exploit the functional relationships between the samples. Besides, some studies prove that the use of network information can boost the classification performances (see section \ref{network}). 

\item P-Net shifts from the feature/biomarker space to the sample/patient space to predict the phenotype, thus exploiting the overall functional similarities between samples, according to a transductive Network Medicine approach. Indeed no model is constructed for the predictions, but it is the topology of the network instead that is used to predict the labels.

\item The use of network information automatically provides a visual representation of the data in the form of graphs and there are publicly available tools to visualize the network and to further analyse the net (see section \ref{network}). 
\end{itemize}

\chapter{Methods}\label{Methods}

This chapter introduces the semi-supervised algorithm  \textbf{P-Net}. This is a network-based method that exploits the ``sample space'' to rank, and at a later stage classify, individuals with respect to a specific phenotype of interest (see \textit{\fullref {Intro}} for more details about network-based methods and semi-supervised algorithms).

In the first section of the current chapter a general overview on the main logical steps of P-Net is showed and it is followed by additional details about each of them.

\section{The P-Net algorithm} \label{pnet_general}

\textit{P-Net} is an acronym that stands for \textit{PatientNet}: Network-based ranking of patient with respect to a given phenotype.
The algorithm produces a predictor which is able to assign to each patient a score related to its odds to show a specific phenotype we want to investigate (e.g. a clinical outcome, the response to a treatment, ...). The scores can be used to rank the patients and, if we choose a proper ``score threshold'', also to binary classify the patients in the two broad classes ``with phenotype'' and ``without phenotype''.

The aforementioned predictor is obtained starting with a set of patients that can be described as a graph $G=<V,E>$, where the set of vertices \textit{V} is a set of patients and the set of edges \textit{E} corresponds to functional relationships between patients. There are different types of functional relationships that can be used to build the edges of the graph: correlation of expression profiles, similarity of the allelic configuration of their SNPs, clinical data or miRNA data are few examples. Each vertex $\textit{v} \in \textit{V}$ is identified by a natural number $\textit{i} \in \lbrace 1, ..., \vert V \vert \rbrace$, where $\vert V \vert$ is the cardinality of the set \textit{V}. It is also necessary define a subset $ V_{C} \in V $ of patients with a given phenotype of interest \textit{C} (e.g. patients having a poor prognosis or patients responsive to a given treatment).
From this starting point we can apply some \textit{kernel methods} and \textit{Kernelized score functions} \cite{6268265, re2012cancer, Vale16a} to obtain a score for each patient, and as a result a patients ranking, to predict which are the patients belonging to the phenotype \textit{C} under study.

We can summarize the \textit{main logical steps} of P-Net as follows:
\begin{enumerate}
\item \nameref{step1}
\item \nameref{step2}
\item \nameref{step3}
\item \nameref{step4} embedding the kernel matrix
\end{enumerate}

\begin{figure}[H]
\centering
\includegraphics[scale=0.7]{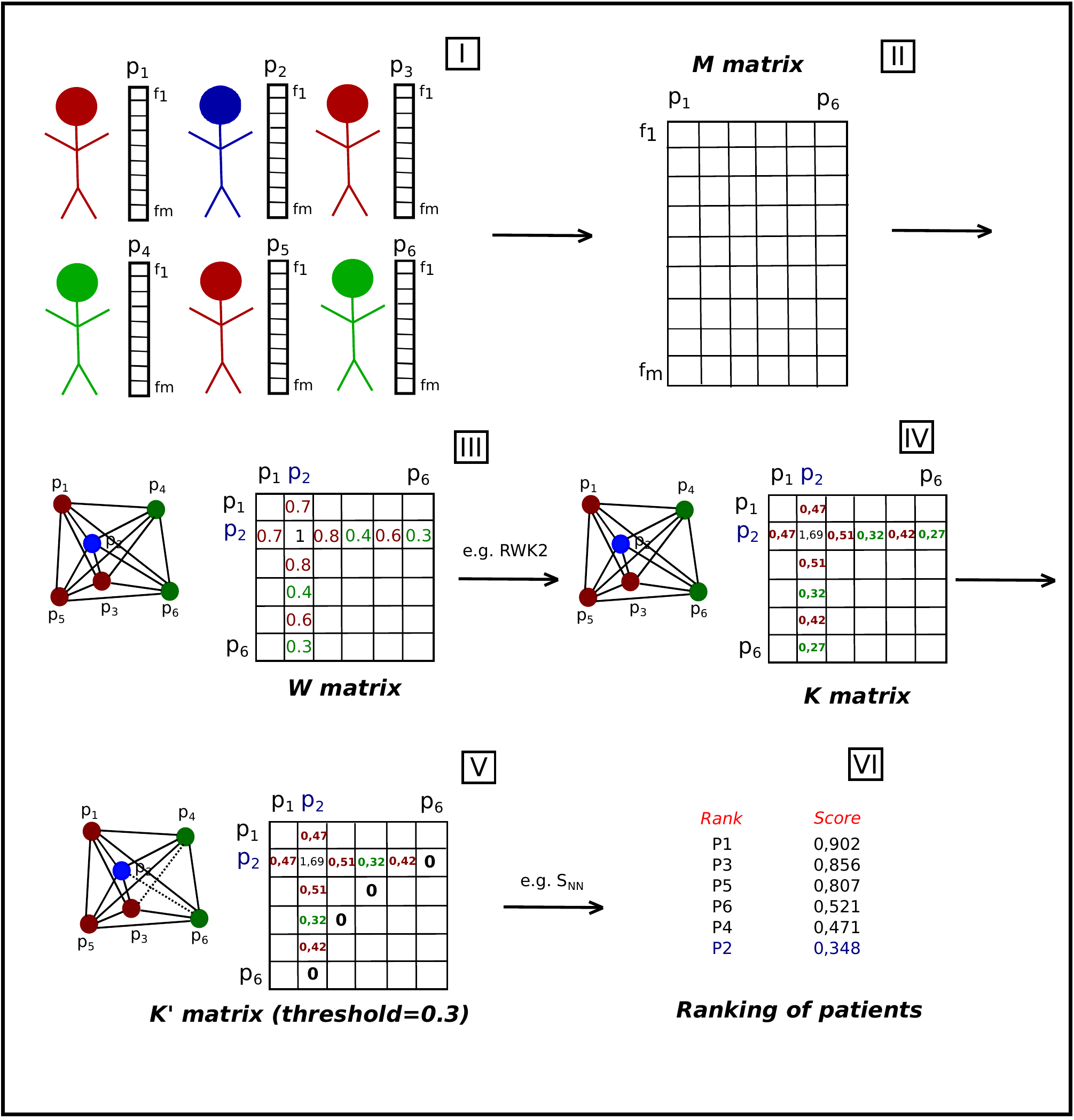}
\caption[Main logical steps of the P-Net algorithm]{\textbf{Main logical steps of the P-Net algorithm.} For the sake of simplicity in this picture the graph has only six nodes/patients. I) Top left: Each patient is represented by a vector of \textit{m} features (e.g. expression levels of the genes). The patients are labelled and in this context red patients/nodes have the phenotype \textit{C} under study, green patients/nodes do not belong to the \textit{C} group, blue patient/node is not labelled because our goal is to give a label to it based on the ranking obtained from P-Net. II) Top right: The vectors that represent the patients are grouped in an $\textit{m}\times\textit{n}$ matrix where $n=1,...,6$. III) Middle left: From the expression profile matrix we compute a W correlation matrix where each element $w_{ij}$ is the correlation between the patients \textit{i} and \textit{j}. We show only the correlation values for patient $p_{2}$ that is the unlabelled one. The W matrix can be represented as a complete graph. IV) Middle right: From the W matrix we can compute the corresponding K kernel matrix applying a proper kernel (e.g. 2-step Random Walk Kernel - RWK2). In this way the weight of the edges is different and we can try to capture the topological features of the graph. V) Bottom left: the K matrix is filtered and we remove all the edges with a weight lower than the chosen threshold (2 edges in this case). VI) Bottom right: We use a \textit{kernelized score function} (e.g. Nearest Neighbour score) to compute a score for each patient and to rank them according to these scores. The $p_{2}$ patient has a low score so we can conclude it has not the phenotype \textit{C}.} 
\label{fig_steps_pnet}
\end{figure}

\subsection{Computation of the similarity matrix between patients} \label{step1}
As previously stated, a set of patients can be described using a graph $G=<V,E>$, where each vertex $v_i$ is a patient and each edge is a functional relationship between two patients. We focus on correlation between expression profiles as functional relationships to build the edges of the graph because this is the kind of relationship used in the \textit{\fullref{Res}}.

Our initial data are collected in the $\textit{m}\times\textit{n}$ \textit{\textbf{M}} matrix that contains the expression profiles of all the patients; where \textit{m} rows refer to features (in other words gene expression levels) and \textit{n} columns to patients. As we can see in the top part of figure \ref{fig_steps_pnet} each patient is represented by a vector of gene expression levels and all these vectors are grouped in the matrix \textbf{\textit{M}}.  

A similarity matrix \textit{\textbf{W}} can be computed from matrix \textit{\textbf{M}} applying different correlation measures between the expression profiles of patients:
\begin{enumerate}
\item Pearson product-moment correlation coefficient
\item Spearman' s rank correlation coefficient
\item Kendall rank correlation coefficient
\end{enumerate}

The similarity matrix \textit{\textbf{W}} is an $\textit{n}\times\textit{n}$ matrix where each element is a coefficient that describes the correlation between the expression profiles of two patients. We can see the matrix \textit{\textbf{W}} as a \textit{\textbf{weighted graph}} $G=<V,E,w>$ where the vertices ${v_{i}}$ are the patients, the edges \textit{E} express the presence of a correlation between expression profiles and \textit{w(e)} is the value of the correlation coefficient associated to each edge. Indeed, we define as weighed graph a graph $G$ where each edge $e$ is associated to a real number $w(e)$, called \textit{weight} \cite{bondy1976graph}.

\subsubsection{Correlation measures: Pearson product-moment correlation coefficient}
The Pearson correlation coefficient \cite{dubitzky2013encyclopedia} is a coefficient to measure the dependence of two variables. Given two random variables \textit{X} and \textit{Y}, Pearson correlation coefficient is defined as the normalized form of their covariance:

\begin{equation}
PCC(X,Y)=\frac{cov(X,Y)}{\sigma_{X}\sigma_{Y}}=\dfrac{E[(X-\bar{X})(Y-\bar{Y})]}{\sigma_{X}\sigma_{Y}}
\end{equation}

where $\bar{X}$ and $\bar{Y}$ are the average values of X and Y; $\sigma_{X}$ and $\sigma_{Y}$ are the corresponding standard deviations; $E$ is the expectation.\\
The value of the correlation coefficient is always between -1 and +1. When $PCC>0$ we say that the variables X and Y are \textit{positively correlated} and when $PCC<0$ we say that the variables are \textit{negatively correlated}. If $PCC\approx0$ there is \textit{no correlation}. \cite{ross2014introduzione}

In our specific case $X$ and $Y$ are two vectors and each vector contains the expression profile of one patient. Pearson correlation coefficient is computed between all the possible pairs of patients and at the end we obtain a similarity matrix \textit{\textbf{W}} with dimensions $\textit{n}\times\textit{n}$ where the elements are the Pearson correlation coefficients between the expression profiles of two patients \textit{X} and \textit{Y}. 

\subsubsection{Correlation measures: Spearman' s rank correlation coefficient}

The Spearman correlation coefficient is another kind of similarity measure that can be used to compute the similarity matrix \textit{\textbf{W}} and it measures the strength of the statistical dependence between two ranked variables.

Let be $X$ and $Y$ the two variables observed. We can assign to both variables a ranking (increasing or decreasing) and we obtain $n$ sets of paired ranks. The correspondent ranking is denoted in the following way: $ R_{i}=rank(X_{i})$ and $ S_{i}=rank(Y_{i}) $. The $n$ paired ranks are: $ (r_{1},s_{1}),(r_{2},s_{2}),...,(r_{n},s_{n}) $.\\
Briefly, we can compute the Pearson correlation coefficient on these paired ranks and the resulting measure is the Spearman' s rank correlation coefficient that measures the degree of correspondence between ranks and not values, as Pearson correlation coefficient does. Indeed, the Spearman coefficient $r_{s}$ is computed from:

\begin{equation}
r_{s}=\rho_{R_{i},S_{i}}=\dfrac{cov(R_{i},S_{i})}{ \sigma_{R_{i}}\sigma_{S_{i}}}
\end{equation}

where $\rho$ denotes the Pearson correlation coefficient applied to the ranked variables, $cov(R_{i},S_{i})$ is the covariance of the ranked variables, $\sigma_{R_{i}}$ and $\sigma_{S_{i}}$ are the standard deviations of the ranked variables.

If the ranked variables are distinct integers we can simplify the formula:

\begin{equation}
r_{s}=1-\dfrac{6\sum^{n}_{i=1}D^{2}_{i}}{n(n^{2}-1)}
\end{equation}

where we assume, without loss of generality, that the $n$ sample pairs are labelled in accordance with increasing magnitude of the $X$ component. In this way $R_{i}=i$ for $i=1,2,...,n$ and $S_{i}$ is the rank of $Y$ which is paired with the rank $i$ of the sample $X$. According to this notation $D_{i}=i-S_{i}$ \cite{gibbons2011nonparametric}.

The Spearman coefficient is between -1 and +1; the interpretation of its value is really similar to the interpretation of Pearson correlation coefficient. 
The Spearman correlation coefficient is a distribution-free test. In other words, we do not make assumptions about the shape of the distribution from which the sample data are drawn. In the Pearson correlation we have to assume that the observations have been drawn from normally distributed populations. So, if one or both the variables under study show a non-normal distribution we have to use the Spearman coefficient. Furthermore, Pearson correlation is appropriate only if there is a linear relationship between two variables (linearity is one of the assumption in the Pearson correlation). While Pearson's correlation assesses linearity, Spearman' s correlation assesses monotonic relationships and not only the linear ones \cite{lehman2005jmp}.

In the context of the P-Net algorithm, and in particular to compute the similarity matrix \textbf{\textit{W}} from the initial matrix \textbf{\textit{M}}, we have to compute the Spearman' s correlation between each pair of ranked expression profiles. In other words, the variables $X$ and $Y$ in this application are two vectors which contain the expression profiles of two patients and the vectors are ranked before the computation of the coefficient. At the end the similarity matrix is a square matrix (patients $\times$ patients) and a single element in the matrix is the Spearman correlation coefficient between two patients.

\subsubsection{Correlation measures: Kendall rank correlation coefficient}
The Kendall rank correlation coefficient \cite{kendall1938new} evaluates the degree of similarity between two sets of ranks given to the same set of objects.

Let \textit{S} be a general set of \textit{N} objects: $S= \lbrace a,b,...x,y \rbrace $. For the sake of simplicity, we can set $N=4$ and obtain $ S= \lbrace a,b,c,d \rbrace $.\\
At this point, the order of the elements of the set is taken into account and we obtain an ordered set $\vartheta_{1}=[a,c,b,d]$ which can be described through the rank order $R_{1}=[1,3,2,4]$ assigned to the elements of the set. The ordered set $\vartheta_{1}$ can be split in all the possible ordered pairs. The number of ordered pairs is equal to $\dfrac{1}{2}N(N-1)$. In this way $\vartheta_{1}$ presents the following ordered pairs: $ P_{1}= \lbrace [a,c],[a,b],[a,d],[c,b],[c,d],[b,d] \rbrace $. If we have another ordered set $\vartheta_{2}$ (and of course its rank order $R_{2}$) we can associated to it another set of ordered pairs $P_{2}$.\\
The goal of the Kendall correlation coefficient is to give a measure to compare these two ordered sets of the same set of objects. In order to do this, the Kendall' s approach is to take into consideration the ordered pairs $P_{1}$ and $P_{2}$ and to count the number of pairs that belong to only one set. In this way, we compute a distance between the two sets which is called \textit{symmetric difference distance}. The symmetric difference distance between $P_{1}$ and $P_{2}$ is $d_{\Delta}(P_{1},P_{2})$. The formula to compute the Kendall rank correlation coefficient is the following:

\begin{equation}
\tau=\frac{1/2 N(N-1)-d_{\Delta}(P_{1},P_{2})}{1/2N(N-1)}= 1-\dfrac{2[d_{\Delta}(P_{1},P_{2})]}{N(N-1)}
\end{equation}

As shown in the above formula, the Kendall correlation coefficient is computed by normalizing the symmetric difference between -1 and +1. When $\tau=-1$ we have the largest possible distance between the two ordered sets and the order of one set is the opposite of the other one, $\tau=+1$ we have the minimum possible distance, equal to zero, and the ordered sets have the same order \cite{abdi2007kendall}. 

After the precedent explanation about the Kendall coefficient, it is easier understand how we can apply it on the initial matrix \textbf{\textit{M}} to compute the similarity matrix \textbf{\textit{W}}. Basically, we compute the coefficient between each pair of patients described by a vector of expression profiles. In our case the set of objects $S$ is given by the expression levels of the genes and for each patient we compute an ordered set $\vartheta_{i}$. For example, if we have 30 patients we obtain 30 ordered set and a similarity matrix \textbf{\textit{W}} (dimensions 30$\times$30) where each element \textit{w} of the matrix is the Kendall correlation coefficient between two patients. 

\subsection{Computation of the kernel matrix from the similarity matrix} \label{step2}

The next step in the P-Net algorithm is the computation of the kernel matrix \textit{\textbf{K}} derived from the similarity matrix \textit{\textbf{W}}. This is achieved by choosing a suitable \textit{kernel}~\cite{Shawe04} able to capture the topological characteristics of the underlying graph. Even if in principle any suitable kernel can be used, the current implementation of P-Net gives us the possibility to choose different kinds of kernel that are able to capture the topology of a given network with a different effectiveness: 
\begin{enumerate}
\item p-step Random Walk Kernel
\item Identity Kernel
\item Linear Kernel
\item Gaussian Kernel
\item Laplacian Kernel
\item Cauchy Kernel
\item Inverse Multiquadric Kernel
\item Polynomial Kernel 
\end{enumerate}

The computation of the kernel matrix \textit{\textbf{K}} allows a better investigation of the relationships between nodes exploiting the topological features of the graph, which are not considered in the \textbf{\textit{W}} matrix.

The kernel matrix \textit{\textbf{K}} is an $\textit{n}\times\textit{n}$ matrix  where each element is a real number that describes, better than before, the relationship between two patients. In the graph $G$, which describes our structured data, the application of the kernel on our similarity matrix \textit{\textbf{W}} involves changes in the value of the weights \textit{\textbf{w(e)}} associated to each edge of the graph. In other words, higher is the weight \textit{\textbf{w(e)}} associated to an edge \textbf{\textit{e}} and more similar are the two patients connected through that edge.

\subsubsection{An overview on Kernels}
In the following text there is a general introduction about the concepts of \textit{kernel} and \textit{kernel methods} which are necessary in order to understand what is a \textit{kernel matrix} and what is a \textit{kernelized score function}.

\textbf{\textit{Kernels}} are a popular method to represent data and to deal with the data representation problems~\cite{Shawe04, smola2003kernels}. We usually have a set of data \textit{S} with \textit{n} objects $S=(x_{1},...,x_{n})$ and $x_{i} \in \textit{X}$. We have to find a way to represent this set \textit{S} and, in order to achieve this goal, the first thing we can do is to define a representation for each individual object. Formally, this means that a representation $\phi(x) \in \textit{F}$ is defined for each object $x_{i} \in \textit{X}$. For example, if our dataset consists of oligonucleotides we can define as representation for each oligonucleotide a sequence of letters which represent the succession of nucleotides. The dataset can be represented as the following set of individual objects: $ \phi(\textit{S})=(\phi(x_{1}),...,\phi(x_{n}))$.

In the \textit{kernel methods} we have the possibility to apply a different approach where \textit{data are not represented individually but by a set of pairwise comparisons}. In this way, instead of using the usual mapping $\phi:\textit{X}\rightarrow \textit{F}$ to represent each object $x \in \textit{X}$ by $\phi(x) \in \textit{F}$, we can use the following \textit{\textbf{comparison function}}: 

\begin{equation}
k:\textit{X}\times \textit{X} \rightarrow \mathbb{R}
\end{equation}

and the dataset \textit{S} is represented by an $\textit{n}\times\textit{n}$ matrix of pairwise comparisons $k_{i,j}=k(x_{i},x_{j})$.

Many kernel methods can process only square matrices, in particular \textit{symmetric positive definite square matrices}. In other words, if \textit{k} is an $\textit{n}\times\textit{n}$ matrix made up by pairwise comparisons, it has to be:

\begin{enumerate}
\item \textbf{symmetric}: $k_{i,j}=k_{j,i}$ for any  $1\leq i,j \leq n $
\item \textbf{positive definite}: $\textbf{c}^\top\textit{k}\textbf{c}\geq 0$ for any $\textbf{c} \in \mathbb{R}^{n}$
\end{enumerate}

It is worthwhile to note that every kernel can be always seen as an inner product between the vectors representing the two general objects $\textbf{x}$ and $\textbf{x}^\prime$
to compare. Indeed, a really systematic way to define kernels for any object $\textbf{x} \in X$ involves:

\begin{enumerate}
\item Representing each object $\textbf{x} \in X$ as a vector $\phi(\textbf{x}) \in \mathbb{R}^{p}$ where $p\geq0$ is the number of elements of the vector. 
\item Define a kernel $k$ for any $\textbf{x},\textbf{x}^\prime \in X$ by: $k(\textbf{x},\textbf{x}^\prime)=\phi(\textbf{x})^\top\phi(\textbf{x}^\prime)$
\end{enumerate}

From these logical steps we derive the following theorem:

\textbf{Theorem}: For any kernel \textit{k} on a space \textit{X}, there exists a Hilbert space \textit{F} and a mapping $\phi:X \rightarrow F$ such that:

\begin{equation}
k(\textbf{x},\textbf{x}^\prime)=\langle \phi(\textbf{x}),\phi(\textbf{x}^\prime) \rangle 
\end{equation} , for any \textbf{x},\textbf{x}$^\prime$ $\in X$. $\langle u,v \rangle$ represents the dot product in the Hilbert space between any pair of points $u,v \in F$.

In this context, the use of a kernel consists in the representation of each object $\textbf{x} \in X$ as a vector $\phi(\textbf{x}) \in F$ and in the computation of the dot product between vectors. However, there is a meaningful difference with respect to the explicit representation of object as vectors seen at the beginning of this subsection: the representation $\phi(\textbf{x})$ does not need  to be computed explicitly for each object in the dataset \textit{S} because the only important thing to take into account is the pairwise dot product. This is the core of what we usually call \textit{\textbf{kernel trick}}: \textit{Any algorithm for vectorial data that can be expressed only in terms of dot products between vectors can be performed implicitly in the feature space F associated with any kernel, by replacing each dot product by a kernel evaluation}.\\
Another important feature of kernels is that kernels are often represented as \textit{similarity measures}. In other words, when $k(\textbf{x},\textbf{x}^\prime)$ is high the objects \textbf{x} and \textbf{x}$^\prime$ are more similar.

Once the concept of kernel is clear we define a \textit{kernel method} as an algorithm that uses as input a similarity matrix defined by a kernel (e.g. of kernel methods are Support Vector Machines, Principal Component Analysis) \cite{vert2004primer}.

\subsubsection{Types of kernels} \label{type_kernel}
There are different types of kernels we can use to  represent our data and map a dataset \textit{S} in the feature space \textit{F}. In this subsection we  want to explain briefly the different kernels we can choose in the P-Net algorithm.

The \textbf{Identity kernel} is a really trivial kernel which derives from the definition of kernel as dot product:
\begin{equation}
k(\textbf{x},\textbf{x}^\prime)=\langle \textbf{x},\textbf{x}^\prime \rangle
\end{equation}

and it leaves the initial similarity matrix \textbf{\textit{W}} unchanged in the P-Net algorithm.

The \textbf{Linear kernel} \cite{vert2004primer} is the simplest kind of kernel we can meet. Basically, it is the inner product between the two objects $\textbf{x},\textbf{x}^\prime \in X$ plus an optional constant \textit{c}: 

\begin{equation}
k(\textbf{x},\textbf{x}^\prime)=\textbf{x}^\top \textbf{x}^\prime+c
\end{equation}

In the context of classification problems with Support Vector Machines, it has been proved that if the number of features is larger than the number of instances we do not need to map our data in an higher dimensional space and the application of a linear kernel is sufficient. Basically, we do not need to apply non-linear mapping (e.g. Gaussian kernel) and the linear kernel performs as good as Gaussian kernel but we can avoid to tune the additional parameter $\sigma$ \cite{hsu2003practical}. This is meaningful because the data we use in the \fullref{Res} are obtained from microarray technology and between the number of patients and features there is a difference of two orders of magnitude.

The \textbf{Gaussian radial basis function kernel} is another kind of kernel function:

\begin{equation}
k(\textbf{x},\textbf{x}^\prime)= exp\left(\dfrac{d(\textbf{x},\textbf{x}^\prime)^{2}}{2\sigma^{2}}\right)
\end{equation}

where $\sigma$ is a parameter and \textit{d} is the Euclidean distance $\Vert \textbf{x}-\textbf{x}^\prime \Vert^{2} $. The value of Gaussian kernel decreases with the distance and its value is always between 0 and +1 (when $\textbf{x}=\textbf{x}^\prime$) \cite{vert2004primer}.

The \textbf{Polynomial kernel} is another possible option:

\begin{equation}
k(\textbf{x},\textbf{x}^\prime)=(\alpha\textbf{x}^\top \textbf{x}^\prime+c)^{d}
\end{equation}

where \textit{d} is the degree of the polynomial, \textit{c} is a constant and $\alpha$ is the slope. As we can see the linear kernel is a particular case of polynomial kernel.

The \textbf{Laplacian kernel} is related to the Gaussian kernel: 

\begin{equation}
k(\textbf{x},\textbf{x}^\prime)= exp\left(\dfrac{d(\textbf{x},\textbf{x}^\prime)}{\sigma}\right)
\end{equation}

where $\sigma$ is a parameter and \textit{d} is the Euclidean distance. 

The \textbf{Cauchy kernel} has the following formula:

\begin{equation}
k(\textbf{x},\textbf{x}^\prime)= \dfrac{1}{1+\dfrac{\Vert \textbf{x} - \textbf{x}^\prime \Vert^{2}}{\sigma^{2}}}
\end{equation}

where $\sigma$ is a parameter and $\Vert \textbf{x}-\textbf{x}^\prime \Vert^{2} $ is the Euclidean distance.
 
The \textbf{Inverse multiquadric kernel}:

\begin{equation}
k(\textbf{x},\textbf{x}^\prime)= \dfrac{1}{\sqrt{\Vert\textbf{x}- \textbf{x}^\prime\Vert^{2}+c^{2}}}
\end{equation}

where $c$ is a parameter and $\Vert \textbf{x}-\textbf{x}^\prime \Vert^{2} $ is the Euclidean distance.

The \textbf{p-step Random Walk Kernel} \cite {smola2003kernels} is defined as:

\begin{equation}
K=((a-1)I+D^{- \frac{1}{2}}WD^{- \frac{1}{2}})^{p}     
\end{equation}

with $p\geq 0$ and $a\geq0$. \textit{I} is the identity matrix, \textit{W} is the weight matrix of an undirected graph, \textit{D} is the diagonal matrix with $D_{ii}=\sum_{j}W_{ij}$, \textit{a} is a parameter related to the probability of remain on the same vertex (the higher is \textit{a} higher is the probability), \textit{p} is the number of steps of the Random Walk.

The p-step Random Walk Kernel is the kernelized version  of \textit{Markov Random Walks} which exploits random trajectories across the graph to analyse the relationships between nodes. In the Markov Random Walks we consider the usual graph $G=\langle V,E \rangle$ and we start from an initial vertex $v_{0}$ (this vertex can be fixed or we can drawn it from an initial distribution $P_{0}$). Let $p$ the prefixed number of steps, we start from the initial vertex $v_{0}$ and then we move to one of its neighbours randomly. Then from this new selected vertex we move to another one randomly and so on. The number of times we pass from a vertex to another is decided by the number of step $p$. At the $p^{th}$ step we are in the $v_{p}$ node and there is a probability $1/d(v_{p})$ we move to neighbours of $v_{p}$. The sequence of random nodes  of the trajectory is a Markov Chain.\\
At the end, we obtain a matrix where each element $p_{ij}^{t}$ is the probability that, starting from the node \textit{i} we reach the node \textit{j} in \textit{p} steps \cite{asz80random}.

\subsection{Filtering of the kernel matrix} \label{step3}
The filtering step of the similarity matrix \textbf{\textit{W}} or the kernel matrix \textbf{\textit{K}} is fundamental in the P-Net algorithm. Indeed, when the \textbf{\textit{W}} matrix is computed by correlation between the expression profiles of patients the corresponding graph is usually complete (or quasi-complete). In a complete graph every pair of nodes is connected by an unique edge and the graph is undirected.

In this context the normal filtering methods used for genes in \cite{6268265,re2012cancer} or drugs in \cite{re2012large} have poor performances. In order to overcome this situation, we need to find a suitable \textit{threshold} so we can remove the irrelevant edges and retain only the relevant ones. Basically, if the weight of the edge $w(e)<threshold$ we can ``cut'' the correspondent edge and remove it. This means that in the similarity matrix \textbf{\textit{W}} or in the kernel matrix \textbf{\textit{K}} we can put $w(e)=0$ and remove the irrelevant edge. If $w(e)>threshold$ the edge is retained.

\textbf{\textit{How can the threshold be selected?}} In order to find the threshold we can use the \textit{cross-validation (or leave-one-out) techniques}. In general the \textit{cross-validation techniques} are methods to asses the performances of an algorithm. In the cross-validation techniques the observations present in the dataset are randomly divided in $k$ groups, or folds, of approximately
equal size. One of these folds is used as \textit{test set} and the remaining $k-1$ folds define the \textit{training set}. The procedure is repeated $k$ times and each time a different fold is used as \textit{test set}. For each round, the algorithm is trained on the training set and the obtained predictor is used to predict a response on the new observations belonging to the test set.\\
The \textit{leave-one-out cross-validation} is an extreme case of cross-validation: in this case the test set involves a single observation $(x_{1} , y_{1})$ and the remaining observations $\lbrace (x_{2},y_{2}),...,(x_{n},y_{n}) \rbrace$ make up the training set. The predictor is trained on the training set and the obtained model is used to predict the single observation in the test set. The procedure is repeated $n$ times changing every time the individual observation in the test set. In this way we obtain a threshold for each patient. For more details about the cross validation techniques see \cite{james2013introduction}.

The use of cross-validation techniques to obtain the threshold is possible because the patient-networks are usually small and because of the relatively low computational complexity of the applied ranking algorithms described in the \fullref{implementation}. Briefly, in the P-Net algorithm an \textit{internal LOO} is used to select the optimal threshold. A vector of thresholds is tested and the threshold which achieves the best AUC (Area Under the Curve) is selected. All the edges below the threshold are removed from the graph. At the end of this procedure we obtain a threshold for each patient. We can decide to exploit these patient-specific thresholds or we can average the thresholds to get a single value. Another solution to obtain a single value is to compute the median of the thresholds obtained by internal loo. 

\subsection{Ranking of patients using kernelized score functions} \label{step4}
The properly filtered kernel matrix \textbf{\textit{K}} can be used to compute the \textit{kernelized score functions} \cite{Vale16a}. These functions allow the ranking of the patients with respect to the specific phenotype under study.\\
On the kernel matrix we can apply one of the kernelized score functions $S:V\rightarrow \mathbb{R}$ which associates to every vertex a real number. In this way, we can use directly the scores $S(v)$ to rank the vertices/patients. The higher is the score, higher is the probability that a patient belongs to the group $V_{C}$ of patients having a specific phenotype \textit{C}.

Once we computed the kernel matrix \textbf{\textit{K}} by the application of a proper kernel (e.g. Random Walk Kernel), the elements of the kernel matrix are described as $k_{ij}$ where the positive integers \textit{i},\textit{j} represent nodes. An element $k_{ij}$ is in the $i^{th}$ row and $j^{th}$ column of the matrix \textbf{\textit{K}}. $\textbf{\textit{K}}_{i}$ represents the $i^{th}$ row of the corresponding matrix and $V_{C}\subset V$ is the subset of ``positive vertices'' (in other words, patients with the phenotype \textit{C} of interest).

According with the precedent notation, we can compute the following \textit{kernelized score functions} \cite{Vale16a}:

{\em Average score}:
\begin{equation}
S_{AV}(i,\textbf{K}_i, V_C) = \frac{1}{|V_C|} \sum_{j \in V_C} k_{ij} 
\label{eq:SAV}
\end{equation}

{\em Nearest Neighbour score}:
\begin{equation}
S_{NN}(i,\textbf{K}_i,V_C) =   \max_{j \in V_C}   k_{ij}
\label{eq:SNN}
\end{equation}

{\em k-Nearest Neighbour score}:
\begin{equation}
S_{kNN}(i,\textbf{K}_i, V_C) =   \sum_{j \in I_k(i)}  k_{ij}
\label{eq:SKNN}
\end{equation}
where  
$I_k(i) = \{j | j \in V_C \wedge \text{rank}(k_{ij}) \leq k \}$.

Patients can be ranked according to scores \ref{eq:SAV}, \ref{eq:SNN} and \ref{eq:SKNN}.

Until this point to compute the scores we considered only the set of ``positive'' patients $V_C$. In order to consider all the available patients we can compute the following \textit{kernelized score function}:

{\em Total score}:
\begin{equation}
S_{TOT}(i,\textbf{K}_i, V_C) = \frac{ \sum_{j \in V_C} k_{ij}}{\sum_{j \in V_C} k_{ij} +\sum_{j \in V \setminus V_C} k_{ij} }
\label{eq:STOT}
\end{equation}

where $V \setminus V_C$ represents all the patients with a different outcome from $V_C$, indeed $V_C \cup V \setminus V_C=V$.

Other possible scores that take into account the contribution of the neighbourhood patients belonging to both $V_C$ and $V \setminus V_C$ are: 

{\em Differential score}:
\begin{equation}
S_{Diff}(i,\textbf{K}_i, V_C) = \sum_{j \in V_C} k_{ij} - \sum_{j \in V \setminus V_C} k_{ij} 
\label{eq:Sdiff}
\end{equation}

{\em Differential normalized score}:
\begin{equation}
S_{Dnorm}(i,\textbf{K}_i, V_C) = \frac{\sum_{j \in V_C} k_{ij} - \sum_{j \in V \setminus V_C} k_{ij}} {\sum_{j \in V_C} k_{ij} + \sum_{j \in V \setminus V_C} k_{ij}} 
\label{eq:Sdiffnorm}
\end{equation}

The score functions \ref{eq:STOT}, \ref{eq:Sdiff} and \ref{eq:Sdiffnorm} allow us to exploit the information that comes from the whole neighbourhood of the investigated node \textit{i}. This can be an advantage if we have to deal with a little dataset, which contains few nodes, and every node is connected with few ``positive'' nodes. This situation is really common in the microarray datasets, as we can see in \fullref{Res}.

\section{Implementation of P-Net} \label{implementation}
The main logical steps of P-Net (described in the precedent \fullref{pnet_general}) can be implemented exploiting a \textit{double leave-one-out procedure}. We use: 
\begin{enumerate}
\item An \textbf{external loo} to assess the performances of the algorithm;
\item An \textbf{internal loo} to select the optimal threshold for the network.
\end{enumerate}
The use of an external loo is not the only possibility to evaluate the performances of the algorithm. In this section some alternative implementations of P-Net are shown and these variants involve the use of \textit{k-Fold cross-validation} or \textit{held-out} replacing the external loo, with the same goal, and an internal loo to select the network threshold. 

\subsection{Implementation of the internal leave-one-out} \label{efficient_loo}
In the leave-one-out technique we usually have to run the assessed algorithm $m$ times, where $m$ is the number of rows or columns in the kernel matrix \textbf{\textit{K}} in this case ($m$ is also the number of nodes in the corresponding patients graph). This is computationally expensive and in this situation we can avoid to run the P-Net algorithm $m$ times. It is sufficient to set to zero the diagonal of the \textbf{\textit{K}} matrix and in this way we can run only once the algorithm. Indeed, the following fact is true in this specific case:

\textbf{Fact 1}: \textit{Given a kernel matrix \textbf{K} computed from the weighted adjacency matrix \textbf{W} of a graph $G=<V,E>$, with nodes $v\in V$ denoted with $i \in \lbrace 1,...,\vert  V\vert \rbrace$ and $ V_{C} \subset V $ the subset of nodes with a given phenotype, when the leave-one-out procedure is applied through P-Net the following fact is true:}

\begin{center}
$ k_{ii}=0$ $\Leftrightarrow$ \textit{i is left out}
\end{center}

\textit{Proof}

\begin{enumerate}
\item $k_{ii}=0 \Rightarrow$ node \textit{i} is left out

If $\forall i$ $k_{ii}=0$ only the following two conditions are possible:
\begin{itemize}
\item $i \notin V_{C}$: in this case $k_{ii}$ is not considered in the computation of the score function $S(i)$ when $S$ is one of the score functions $S_{AV}$, $S_{NN}$, $S_{kNN}$. $k_{ii}$ is always included in the computation of the score functions $S_{TOT}$, $S_{Diff}$, $S_{Dnorm}$, but by hypothesis $k_{ii}=0$.
\item $i \in V_{C}$: in this case $k_{ii}$ is considered in the computation of the score functions, but by hypothesis $k_{ii}=0$.
\end{itemize}
In both the conditions, the score of the node $i$ is computed without considering the label of the node $i$ and this is logically equivalent to left out the node $i$.
\item $k_{ii}=0 \Leftarrow$ node \textit{i} is left out

If $i$ is left out, $k_{ii}$ is not used in the computation of the score $S$ (this is independent from the label of the node $i$) and it is equivalent to set $k_{ii}=0$ and to use it in the computation of $S(i)$.
\end{enumerate}

Once explained the trick which allows us to run just once the P-Net algorithm we can introduce the procedure \textit{Optimize\_thresh\_by\_loo} which exploits the efficient implementation of the internal loo to select the optimal network threshold.

\begin{figure}[H]
\centering
\includegraphics[scale=0.7]{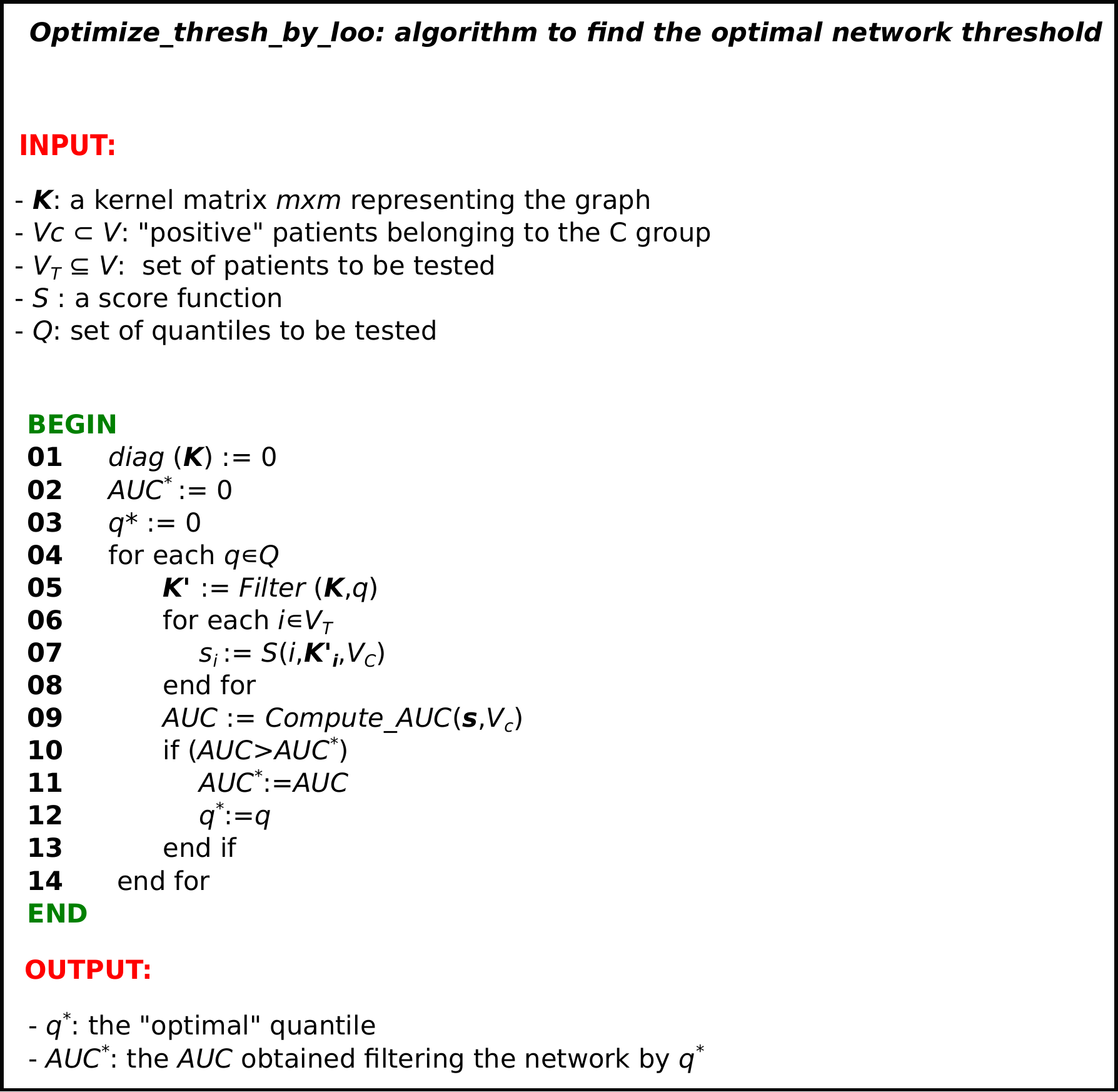}
\caption[Pseudo-code of the algorithm \textit{Optimize\_thresh\_by\_loo}]{\textbf{Pseudo-code of the algorithm \textit{Optimize\_thresh\_by\_loo}.}} 
\label{fig_opt_thresh}
\end{figure}

In the algorithm (see figure \ref{fig_opt_thresh}) the diagonal of the kernel matrix \textbf{\textit{K}} is set to zero (row 1). It is worthwhile to note that the input matrix in the algorithm can be or a kernel or a correlation matrix, but it has to be a symmetric matrix representing similarities between patients. In the rows 02-03 we initialize the optimal Area Under the Curve ($AUC^{*}$) and the optimal quantile $q^{*}$ to zero. In the main \textit{for loop} (rows 04-14) a set of pre-selected quantiles $q \in Q$ is tested to find the optimal quantile $ q^{*} $ which is the quantile that gives the best performances in term of AUC when it is used as threshold to filter the graph. In this loop, the algorithm compute a different filtered Kernel matrix \textbf{\textit{K}}$^\prime$ at each repetition by choosing a different quantile \textit{q} each time. The \textit{Filter} function removes all the edges with a weight below the threshold corresponding to the selected quantile and it changes their values to zero. In the following inner \textit{for loop} (rows 06-08) the score function $S$ (one of the functions in \fullref{step4}) is applied on the filtered kernel matrix to compute the score $s_{i}$ of each node $i$. Only the nodes included in the subset $V_{T}$ are tested. At the row 09 the function \textit{Compute\_AUC} computes the AUC exploiting the vector of scores \textit{\textbf{s}} obtained from the precedent \textit{for loop}. The computed AUC and the corresponding quantile \textit{q} used to filter the kernel matrix are stored. Then if the stored AUC is higher of the current maximal Area Under the Curve (AUC*), the stored AUC becomes the new $AUC^{*}$ and the corresponding quantile becomes $q^{*}$ (rows 10-13). At the end of the procedure the ``optimal'' quantile and the corresponding AUC achieved are returned.

The complexity of the algorithm {\em Optimize\_thresh\_by\_loo} is $\mathcal{O}(m^2)$ if $|Q| << m$, otherwise the complexity is $\mathcal{O}(m^3)$. Indeed, within the \textit{for each} loop the most expensive procedures are \textit{Filter} (row $05$) and the \textit{for loop} at rows 06-08, both with complexity $\mathcal{O}(m^2)$.

\subsubsection{The overall procedure of P-Net: double leave-one-out}
\begin{figure}[H]
\centering
\includegraphics[scale=0.7]{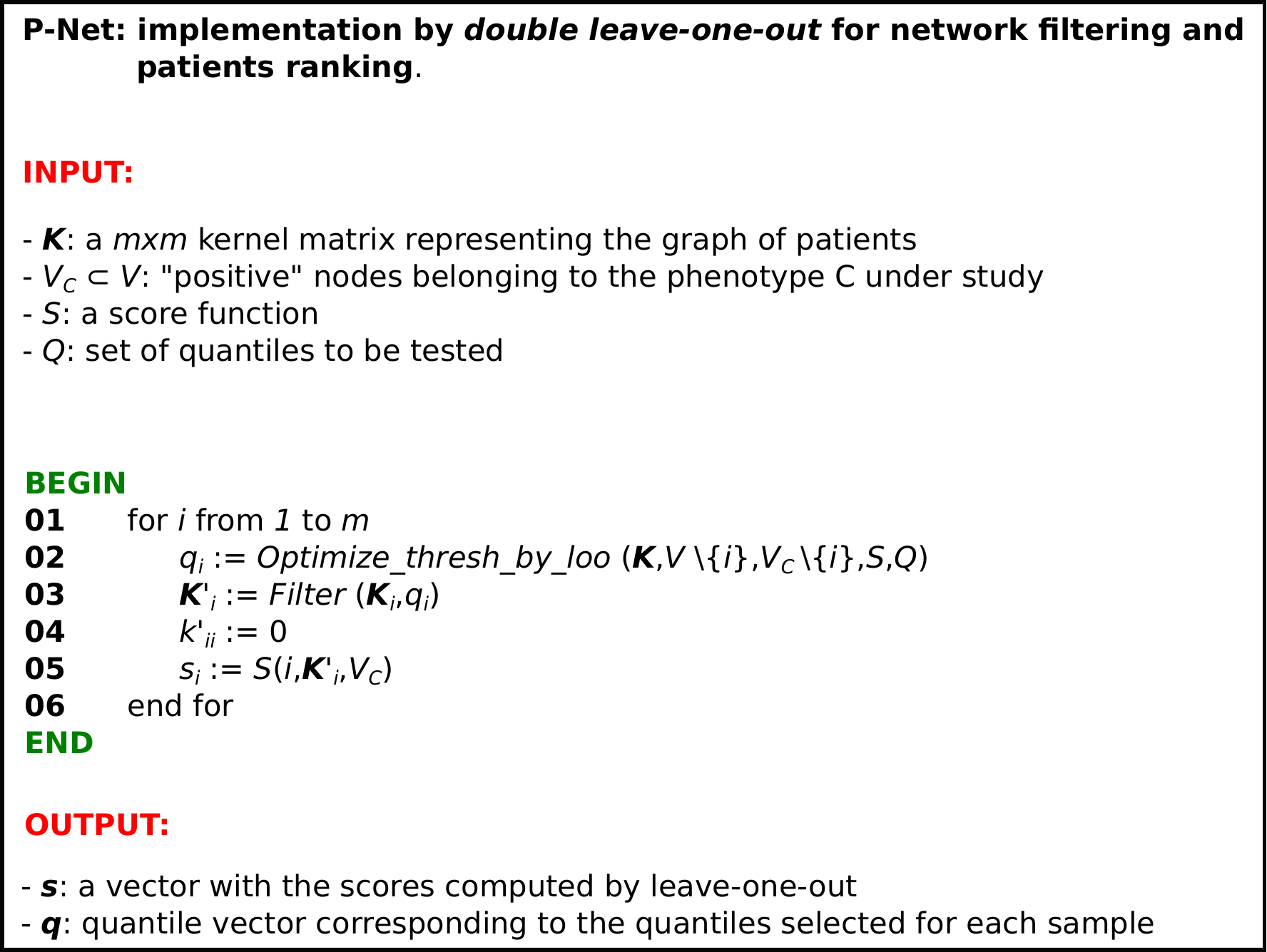}
\caption[Pseudo-code of the P-Net algorithm with a double leave-one-out]{\textbf{Pseudo-code of the P-Net algorithm with a double leave-one-out.}} 
\label{fig_double_loo}
\end{figure}

In the figure \ref{fig_double_loo} it is showed the pseudo-code of the P-Net algorithm. Basically, there is a main \textit{for loop} (row 02-06) necessary to obtain a score for each node \textit{i} of the input matrix \textbf{\textit{K}}. So, the P-Net algorithm computes the ``optimal'' quantile $q_{i}$ and the score $s_{i}$ for each node applying the same steps node by node. In the row 02 the function \textit{Optimize\_thresh\_by\_loo} implements the \textit{internal loo} to select the ``optimal'' quantile for the $i^{th}$ node. In this loo the node $i$ is left out and it is not considered in the computation. Then the quantile $q_{i}$ is exploited to filter the $i^{th}$ row of the matrix \textit{\textbf{K}}. In the $i^{th}$ row all the elements below to $q_{i}$ are set to zero and this is equivalent to cut the edges incident to the node \textit{i} in the corresponding graph. In the 04 row the element $k'_{ii}$ (it is the element in the diagonal of the filtered matrix \textbf{\textit{K}}$^\prime$) is changed to zero. This is necessary to implement the \textit{external loo} through the main for loop (rows 01-06). The last step (row 05) is the computation of the score $s_{i}$ using the filtered row $K'_{i}$. Then the external for loop starts again considering the next node $i_{th}+1$ and at each iteration we have the following steps:
\begin{enumerate}
\item the optimal quantile $q_{i}$ is computed by internal loo exploiting all the available information except the node $i_{th}$;
\item the row $K_{i}$ is filtered using the optimal quantile
\item the score for the node \textit{i} is computed
\end{enumerate}
The outputs of this procedure are: a vector \textbf{\textit{s}} which contains the scores for each node computed by external loo; a vector \textbf{\textit{q}} which contains all the ``optimal'' quantiles selected for each patient by internal loo. 

It is worthwhile to note that the \textit{double loo procedure} seems to be computationally intensive but we have to consider the use of an efficient implementation of the loo technique, described in \ref{efficient_loo}. At the end, the overall complexity of \textit{P-Net} is $\mathcal{O}(m^3)$. Indeed, the \textit{Optimize\_thresh\_by\_loo} procedure has complexity $\mathcal{O}(m^2)$ when $|Q| << m$ is iterated \textit{m} times. 

\subsection{Cross-validation based implementation of P-Net}
Another efficient implementation of P-Net exploits:
\begin{itemize}
\item an \textbf{internal loo} to select the ``optimal'' threshold for the network 
\item an \textbf{external k-fold cross-validation} to assess the capabilities of generalization of the algorithm
\end{itemize}

\begin{figure}[H]
\centering
\includegraphics[scale=0.8]{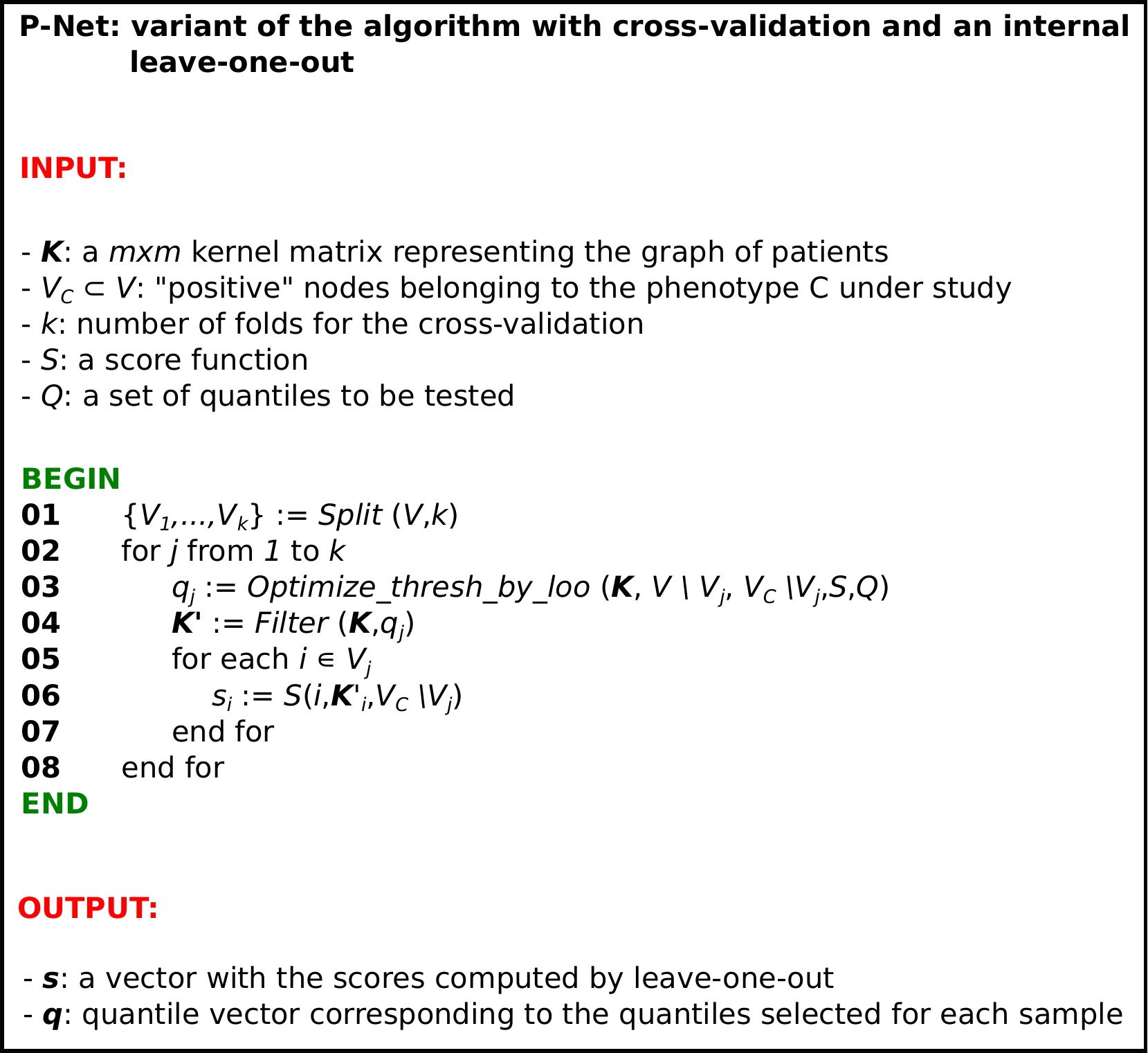}
\caption[Pseudo-code of the P-Net algorithm with an external cross-validation]{\textbf{Pseudo-code of the P-Net algorithm with an external cross-validation.}} 
\label{fig_pnet_cv}
\end{figure}

In the pseudo-code (see figure \ref{fig_pnet_cv}) we can see how the main structure of P-Net remains the same but we use a \textit{cross-validation technique} instead of an external loo. In the row 01 the function \textit{Split} divides randomly the vertices \textit{V} in \textit{k} folds. As shown previously, there is a main \textit{for loop} which considers at each iteration a different fold (rows 02-08). Inside the for loop,  there is the computation of the quantile $q_{j}$ (row 03) related to the $j^{th}$ fold using the \textit{Optimize\_thresh\_by\_loo} function. The function computes the quantile $q_{j}$ by internal loo on the vertices belonging to $V \backslash V_{j}$. Then the matrix \textbf{\textit{K}} is filtered using the optimal quantile for the fold $V_{j}$. Finally, in the last loop (rows 05-07) the score for each node belonging to the test fold $V_{j}$ is computed. In general, the kernelized score function is computed \textit{k} times on the test fold $V_{j}$. 

It is easy to see that the {\em held-out} version of {\em P-Net}, by which data are split in training and test set, can be obtained from the cross-validation version of Fig.~\ref{fig_pnet_cv}.
Indeed we can substitute the {\em Split} function at row $01$ of Fig.~\ref{fig_pnet_cv} with a simpler splitting of the vertices $V$ in training and test set. Then the outer  \textit{for loop} can be eliminated: it is sufficient a single iteration of the \textit{for loop} at rows 02-08, and finally $V_j$ can be substituted with the test set.

\section{Feature selection methods} \label{feat_sel}

\textit{Feature selection} (also known as \textit{variable selection} or \textit{subset selection}) is a process widely used in machine learning. It consists in the selection of a subset of relevant features, which are the most significant to build a model/predictor.\\
These methods are widely applied in the data mining of datasets involving thousands of features (e.g. text processing, microarray data). The advantages achieved with the application of feature selection methods can be summarized in the following points:

\begin{itemize}
\item The models became easier to interpret by the researchers \cite{james2013introduction}
\item These methods improve the prediction performances of the predictors defying the curse of dimensionality 
\item They provide faster and more cost-effective predictors and a better understanding of the underlying process that generated the data \cite{guyon2003introduction}
\end{itemize}

When we use a feature selection method, we assume that the data contains many features which are \textit{redundant} or \textit{irrelevant} and we can remove these features with a negligible loss of information. This is necessary to avoid the reduction of the performances of the predictor related to the problem of over-fitting, which is common when the number of features is high compared to the number of examples \cite{duangsoithong2009relevant}. Basically, the algorithm is trained on the selected training set data and we obtain a model, which should be able to predict the desired output on an unseen example. If the examples in the training set are too little with respect to the number of variables, the learner algorithm produce a model that is too complex and has too many parameters. The model is adapted to some characteristics that are specific of the training set and its generalization capabilities are reduced.\\
The feature selection methods \cite{guyon2003introduction} can be categorized into three main classes:

\begin{enumerate}
\item \textbf{Filter methods:} 
These filters are used as a pre-selection methods and they are independent from the chosen machine learning algorithm, applied at a later time. These methods are usually based on univariate test statistics and some examples are: Fisher score, Pearson correlation, Chi-square test. Basically, each feature is ranked according to its statistical value and the subset of features can be selected after the choice of a suitable statistical cut-off. The features with a statistical value below the cut-off are the least interesting and they are not considered for the construction of the final selected subset.\\
Filter methods are usually fast and computationally simple because these approaches require a single pass to the whole set of features, so they can easily applied to high-dimensional datasets. Moreover, they are independent from the used classifier. On the other hand, they are usually univariate methods and this means that they consider each feature independently from the others. In this way, filter methods do not consider the relationship between variables and this may lead to a reduction of the performances because we do not take into account the combined effect of multiple features together. Indeed, a variable that is useless by itself (and it is not considered after the application of a filter method) can provide a significant improvement in the performances when taken with others. Moreover, two variables that are useless by themselves can be useful together. 

\item \textbf{Wrapper methods:}
The method considers the learner as a black box and it uses the performances of the predictor to evaluate the goodness of subsets of features. These methods involve three elements. At the beginning a \textit{search algorithm} is used to explore the space of all the possible feature subsets. Then, a \textit{score function} is chosen to assess the performances of the subsets based on the  \textit{learning algorithm} around which the selection procedure is wrapped. 
In this case, there is interaction between the search of a feature subset and the model construction. These methods take into account the relationships between features. The most important limitation of these methods is that an exhaustive search of the subsets space can be carry out only if the number of variables is not too large. Indeed, the time complexity of the methods grows exponentially with the number of screened features. To handle this issue, heuristic search techniques are often exploited \cite {janecek2008relationship}. 

\item \textbf{Embedded methods:}
These methods incorporate variable selection during the training phase of a predictor and they are specific for a given learning machine. In this way, the trained predictive model depends only on a subset of the original variables. The advantage respect to the wrapper methods is that they are faster and less computationally intensive.  
\end{enumerate}

Once we explained the different types of variable selection methods, we continue talking briefly about the \textit{Welch's t-test} and \textit{Moderated t-statistic method}, which are the two chosen filter methods used in the experimental workflows presented in the subsections \ref{setup_ggc} and \ref{setup_mel_ov}, respectively.   

\subsection{Welch's t-test}

The \textbf{Welch's t-test} \cite{welch1947generalization} is a statistical test to determine if the difference between the means of two groups is statistically significant. This is a common statistical test employed as feature selection method (more precisely, it is a filter method). For the sake of simplicity, the following explanation about the Welch's t-test refers to the application on differential expression of genes because we use microarray data in our work (see subsection \ref{setup_ggc}, where we used the Welch's t-test two-sided with a confidence level of 0.95).\\
In this context, when we use the t-test the purpose is to understand if the genes are differentially expressed between two different conditions (e.g. two different kinds of tumours, patients with \textit{good prognosis} vs. \textit{poor prognosis}). In the Welch's t-test, we assume that the two groups of samples are drawn from a normal distribution and the variance between the two groups is unequal. Indeed, in the Student t-test we assume that the same feature in two different groups has similar variance. But, this is not common in the microarray experiments and the Welch's t-test is more appropriate for this kind of data \cite{sreekumar2008statistical}. The unequal variance t-test should be used whenever the variances are unknown, and/or the sample size is small (not greater than 30 samples), and/or you wish to be conservative in the inferences that you were drawn. The main difference between t-test and Welch's t-test is that the computation of the degrees of freedom is more complicated in the latter method \cite{boslaugh2012statistics}.\\
So, we start from a dataset with \textit{n} samples and \textit{m} genes. The samples are split into the two groups $G_{1}$ and $G_{2}$, which have different phenotypes. Then, we carry out a statistic test for each gene/feature $f=1,...,m$ and at the end of the test we obtain a different \textit{p-value} for each gene. For each gene, we test the null hypothesis ($H_{0}$) that the expected expression level for a given gene is equal between the two groups, against the alternative hypothesis ($H_{1}$) that they are different. Welch's t-test defines the statistic \textit{t} as:

\begin{equation}
t=\dfrac{\overline{X}_{1} - \overline{X}_{2}}{\sqrt{\dfrac{s_{1}^{2}}{N_{1}}+{\dfrac{s_{2}^{2}}{N_{2}}}}}
\end{equation}

where $\overline{X}_{1}$ is the sample mean, $s_{1}^{2}$ is the sample variance and $N_{1}$ is the sample size (all related to the first group $G_{1}$). The degrees of freedom $v$ are computed in the following way: 

\begin{equation}
v \approx \dfrac{ \left( \dfrac{s_{1}^{2}}{N_{1}}+{\dfrac{s_{2}^{2}}{N_{2}}}\right) ^2}{ \dfrac{s_{1}^{4}}{N_{1}^{2}v_{1}}+{\dfrac{s_{2}^{4}}{N_{2}^{2}v_{2}}}}
\end{equation}

where $v_{1}=N_{1}-1$ and $v_{2}=N_{2}-1$ are the degrees of freedom related to the first and second variance estimated. 

After the computation of $v$ and $t$, these values can be used with the t-distribution to test the null hypothesis that the two means are equal and we obtain a p-value for each gene. The genes are ranked with respect to the p-value and we can choose a threshold to select the most differentially expressed features (e.g. the first 1000 features with the smallest p-value) and to obtain a subset of significant features. 

\subsection{Moderated t-statistic}

The \textbf{moderated t-statistic} \cite{berkeley2004linear} is another filter feature selection method and it can be carry out using the \textit{limma} package in R. Basically, it takes a gene expression matrix \textit{m}$\times$\textit{n} as input where each row represents a gene and each column represents a sample. The method fits a linear model for each gene \textit{g} (in other words, for each row of the expression matrix):

\begin{equation}
E(y_{g})=X\beta_{g}
\end{equation}
 
where $y_{g}$ is the vector of expression values for the gene $g$, $X$ is the design matrix that relates these values to some regression coefficients of interest $\beta_{g}$.\\
The purpose of these models is to assess the differential expression of the genes computing log-ratios (for two-channel data) or log-intensities (for single-channel data) between two or more samples. The differential expression can be evaluated also between two groups of samples having different phenotypes. The regression coefficients and the standard errors are computed and they estimate the comparison of interest. 
After the computation of the \textit{m} linear models, we can apply the parametric empirical Bayes, a class of statistical methods which is a way to consider the information not only from one gene at a time but from all the genes in the matrix. In this way, the estimated variance for each gene is a compromise between the gene-wise estimator, computed using the data of a single gene, and the global variability across all the genes. In other words, the variance is moderated across genes. This method is particularly suitable for experiments with small numbers of samples because the inference remains reliable and stable even in this case. Moreover, the use of global parameters, to take into account the information from all the genes to compute the variances, allows to the moderated t-test to handle unequal variances. The empirical Bayes moderated t-statistic and their associated p-values are used to assess the significance of the observed expression changes. Smaller is the p-value, higher is the statistical significance of the expression change. At the end of the procedure, we obtain a p-value for each gene and we can rank them and select a subset of features choosing a proper threshold \cite{ritchie2015limma}. 

\chapter{Results}\label{Res}
In this chapter the new semi-supervised algorithm P-Net is tested in order to evaluate its performances. To this end we compared P-Net with several baseline methods, with a particular focus on state-of-the-art methods that apply a network-based approach  in the context of outcome/phenotype prediction problems. More precisely we selected two state-of-the-art methods~\cite{barter2014network,winter2012google} that adopt network-based learning strategies to learn and predict the phenotype associated to a specific patient/sample on the basis of its specific bio-molecular profile. It is worth noting that these methods, as usual in systems biology, work on the ``feature'' network space (i.e. on networks where nodes are represented by e.g. genes or proteins), while P-Net works on the ``sample'' space, i.e. on networks where the samples/patients represent nodes and edges their similarities based on the biomolecular profile of samples/patients. 
To our knowledge, only a few network-based methods working on the ``sample space'' have been proposed in literature \cite{park2014integrative, wang2014similarity}. 

There are different experimental strategies to compare a new proposed algorithm with state-of-the-art methods. For example, we can choose some publicly available datasets and apply state-of-the-art algorithms to these data. Then we can use on the same data P-Net and we can proceed with a comparison of the experimental results.
In this work we decided to follow a different strategy and to compare P-Net with  experimental results already published on scientific journals. Basically, we selected two papers \cite{barter2014network, winter2012google} and in each one a new clinical outcome predictor is presented and it is compared with some standard methods. We compared our method with the results showed in these papers and, to achieve this goal, we strictly followed the experimental set-up applied in the papers using the same datasets. In this way we can safely compare the published results with those obtained by P-Net.

All the tests presented in this work are carried out using R \cite{venables2016introduction} as programming language (unless explicitly specified otherwise).

\section{Summary of strategies and methods applied in the reference papers} \label{summary}

In this section, there is a summary of the methods and the experimental set-ups presented in the two papers selected for comparison purposes.\\
In \cite{winter2012google}, the authors proposed \textit{\textbf{NetRank}}, based on the well-known \textit{Google's PageRank} algorithm~\cite{ilprints422}.
PageRank exploits the information about the hyperlinks between web pages, to better select which are the most relevant ones. Similarly, NetRank exploits the information about biological interactions between gene products to decide which genes are more relevant for the outcome prediction task. NetRank assigns a score to one gene which is influenced by the score of the connected genes. The interactions between genes are created using a transcription factor-target network. At the beginning, NetRank assigns to each gene the absolute correlation between the gene expression level and the patient survival time. Then, the network is exploited to spread this correlation to its neighbours and beyond. The algorithm is applied to gene expression profiles of pancreatic ductal adenocarcinoma samples, which are split into the two classes \textit{good prognosis} and \textit{poor prognosis} based on their survival time. In this way, the authors can rank the genes and they found seven candidate genes to be prognostic biomarkers. Five methods were tested on the same dataset to prioritise genes with respect to their ability to discriminate between the two prognostic groups. These methods are: \textit{fold-change}, \textit{t-statistic}, \textit{Pearson and Spearman correlation coefficients} between gene expression and survival time of each patient, \textit{SAM method} (Significance Analysis of Microarray), \textit{NetRank} algorithm and the \textit{random selection} of genes as control method. The methods were used to rank the genes and the first 5-10 genes for each method were used to train a different SVM classifier. The authors computed the accuracy under 1000 rounds of Monte Carlo cross-validation to assess the classification error rates and the procedure was repeated with different training and test sizes (see subsection \ref{setup_ggc} for further details on the experimental workflow). Then, the signature selected by NetRank was validated using immunohistochemistry on an independent set of 412 pancreatic cancer samples. The validation step proofs that the selected signature is more predictive than the classic clinical parameters (e.g. grade, tumour size, nodal status). Moreover, the signature selected by NetRank is more accurate than signatures from other computational methods or from literature. 

In the paper \cite{barter2014network}, the authors compare some classic approaches to select gene expression signatures (single-gene \textit{Moderated t-statistic} and gene-set \textit{Median expression} methods) with respect to network-based methods. The methods are employed to distinguish between \textit{good prognosis} and \textit{poor prognosis} patients in two different cohorts of gene expression data from melanoma and ovarian cancer. The network-based methods were assessed on two different protein-protein interaction networks and the three selected methods are:

\begin{itemize}
\item \textit{\textbf{NetRank}} \cite{winter2012google}
\item \textit{\textbf{Taylor's Method}} \cite{taylor2009dynamic}, which is based on identifying differential network behaviour between sub-networks
\item \textit{\textbf{BSS/WSS approach}}, which is presented for the first time in this paper and it is inspired by Taylor's differential correlation measure.  
\end{itemize} 

All these approaches can be categorised as feature selection methods and they were used to identify the most significant features to classify the patients. Then, each of these signatures was used to train three different classifiers (Random Forest, Support Vector Machines and Diagonal Linear Discriminant Analysis). The results achieved from each combination of feature selection method and classifier were compared using the classification error rate. The error rates were estimated under 100 rounds of 5-fold cross-validation approach. Moreover, the authors assessed also the stability of the feature selection methods, accuracy at a patient-level and class-specific classification error rate (see subsection \ref{setup_mel_ov} for further details on the experimental workflow). NetRank, coupled with SVM or DLDA classifiers, is the method with the best classification error rate on the melanoma dataset but it is not the best method on the ovarian cancer dataset, where it is outperformed by moderated t-statistic coupled with SVM classifier. The class-specific analysis shows that the patients with a good prognosis are easier to classify in the melanoma dataset but, in the ovarian cancer dataset, it is easier to classify the patients with poor prognosis. NetRank is the method with the best features stability with an average of 63\% of features in common considering only the first 50 features. Another interesting result comes from the analysis at a patient-level. Indeed, it proofs that there are a group of patients that is always classified correctly by every method (patients ``easy to classify'') and another group which is never classified correctly (patients ``hard to classify''). These two groups of patients are present in both the datasets. Some of the remaining patients are better classified by NetRank and moderated t-statistic method and other remaining patients are better classified by the other methods. Basically, NetRank and moderated t-statistic are able to capture a different subset of the sample space in the melanoma dataset. However, in the ovarian cancer dataset, the network-based methods and the gene-set method capture a different subset of patients. All these observations are confirmed using two different protein-protein networks: MetaCore\textsuperscript{TM} PPI network and iRefWeb PPI network. 

\section{Analysis on Pancreatic ductal adenocarcinoma dataset}

As previously stated, we want to follow the experimental set-up of the reference paper \cite{winter2012google} on the same publicly available pancreatic cancer dataset.\\
The dataset is free to download in the \textit{ArrayExpress Archive} \cite{parkinson2007arrayexpress} under the accession number E-MEXP-2780. It consists of gene expression data from 30 patients that suffer from pancreatic ductal adenocarcinoma. More precisely, the gene expression levels were measured by microarray technology (platform: Affymetrix GeneChip
Human Genome U133 Plus 2.0) and the raw probe level intensity files (CEL files)  were processed using RMA (robust multi-array average expression measure). The processing steps involve background correction, quantile normalization, and summarization based on multi-array model fit using the median polish algorithm. The above pre-processing steps were  accomplished using the \textit{rma function} with default settings from the Bioconductor \textit{affy} package \cite{gautier2004affy}. The expression values were $log_{2}$ transformed.
The final gene expression matrix \textbf{\textit{M}} is stored in the ArrayExpress database with the \textbf{\textit{pdata}} dataframe, which contains all the phenotypic variables related to the different 30 samples available. The \textbf{\textit{M}} matrix contains the expression level of 54,675 probe sets for each patient and the \textbf{\textit{pdata}} dataframe contains 31 phenotypic variables (e.g. time of survival, sex, disease state, etc.) for each patient.

\subsection{Experimental set-up for the pancreatic cancer dataset} \label{setup_ggc}

In this paragraph it is schematically presented the workflow to run the experiments on the pancreatic cancer dataset. The experimental set-up is designed to be as much as possible similar to the one described in the reference paper. 

\subsubsection{Pre-filtering steps on the initial dataset}

The first part of the experimental workflow consists in the application of some pre-filtering methods on the initial expression matrix \textbf{\textit{M}}:

\begin{enumerate}
\item The probe sets with a mean expression level below 6 on the $log_{2}$ scale are filtered out from the matrix \textbf{\textit{M}}. The purpose is to remove some noise due to genes having a low expression and 29695 probe sets remain in the matrix.
\item Then, we remove the probe sets that show a little variation among the patients because they are not so informative. So, probe sets with a standard deviation below 0.5 on the $log_{2}$ scale are left out and only 15251 probe sets remain in the matrix. 
\item In the last filtering step, only the probe set with the highest mean expression for each gene is retained and the final number of genes is 9175.
\end{enumerate}

The annotation of the genes from the probe sets is required to carry out the last pre-filtering step. It is accomplished using the annotation package \textit{hgu133plus2.db}. We have to report that the authors obtain 7871 genes at the end of the pre-filtering procedure, despite we tried to follow strictly the workflow described in the reference paper and we used the same release of the annotation package.\\ 
Finally, we obtain the \textit{filtered matrix \textbf{fM}} which is used to carry out all the tests on the pancreatic cancer dataset.

\subsubsection{Creation of the vector of labels}

It is also necessary to create a vector that contains the label of each patient. Indeed, this vector represents the \textit{true outcome} of the patients to compare with respect to the \textit{predicted outcome} obtained from the chosen classifier. The comparison between true and predicted outcome is necessary to evaluate the performances of the classifiers.\\ 
We can split the patients into two classes (\textit{good prognosis - GP} and \textit{poor prognosis - PP}) that are based on the \textit{time of survival} of the patients contained in the \textit{\textbf{pdata}} dataframe. Patients are labelled as PP if their time of survival is below 17.5 months, otherwise the label is GP. 17.5 months is the median survival time. We are usually more interested in the accurate prediction of the patients with a poor prognosis, so from this point forward the \textit{positive patients}, in other words patients belonging to the group $V_{C} \in V$ under study, are the patients with the label PP (which is the phenotype \textit{C} under study).

\subsubsection{Monte Carlo cross-validation technique}

Finally, we can proceed with the application of the classifier on the filtered dataset. We use the \textbf{Monte Carlo cross-validation (MCCV) technique}, as described in the reference paper \cite{winter2012google}, to evaluate the performances of P-Net.

\begin{figure}[H]
\centering
\includegraphics[scale=0.7]{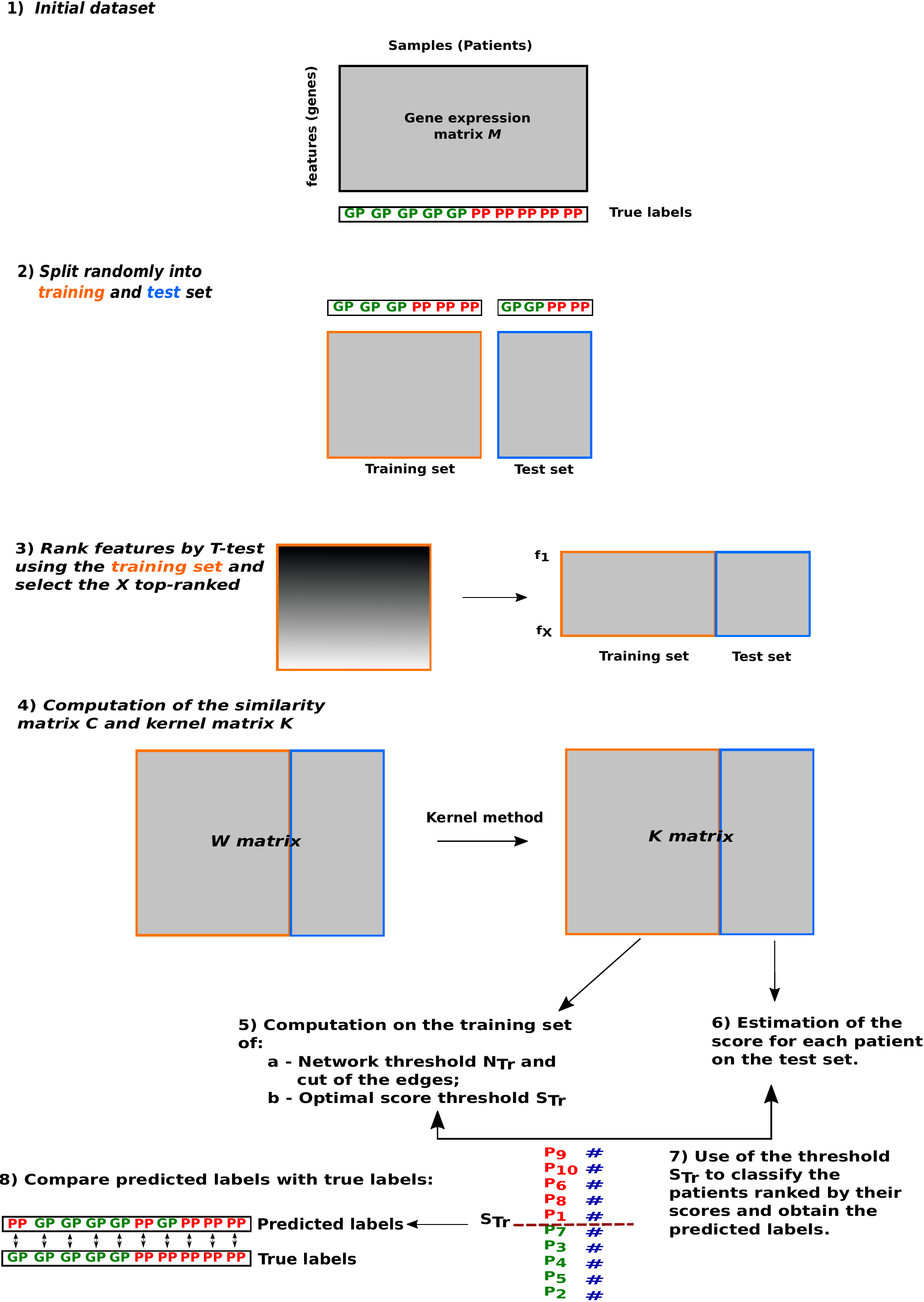}
\caption [MCCV workflow using P-Net for the classification]{\textbf{Monte Carlo cross-validation workflow using P-Net for the classification.}} 
\label{fig_setup_ggc}
\end{figure}

The following steps implement one round of the MCCV and they are repeated 1000 times to get a robust evaluation of the accuracy (see figure \ref{fig_setup_ggc}):

\begin{enumerate}[a)]
\item The data are randomly split in \textit{training set} and \textit{test set}. The split is balanced; this means that the number of samples with good prognosis and poor prognosis in the test set are either equal or differ by at most one. In this way, we can avoid the over-representation of one group in the training set.
 
\item The next step is the application of the \textbf{T-test} as feature selection method (see section \ref{feat_sel} for further details) on the training set. The purpose of the feature selection is to pick the \textit{x} most significant features out in order to obtain a better accuracy from the predictor and to reduce the computation time. If we choose $x=1000$, after the application of the T-test on the training set, we obtain a \textit{p-value} for each feature and then we can select the first 1000 features with the smallest \textit{p-value}.
We obtain the filtered matrix \textbf{\textit{tM}} with 1000 rows and 30 columns, which are respectively genes and patients. The T-test is computed only on the training set data to avoid the problem of the selection bias \cite{ambroise2002selection}.
 
\item Different features can be selected at each iteration of the MCCV and we can obtain a different matrix \textbf{\textit{tM}}. The matrix is used as input of \textbf{P-Net}, where the optimisation of the edge threshold is carried out by internal leave-one-out on the training set data and the estimation of the scores is performed on the test set data.
 
\item The performances of the algorithm can be evaluated exploiting different kinds of measures (e.g. accuracy, error rate, precision, recall, AUC, etc.). Their computation starts from the patients' score and it requires a \textit{score threshold}, which is necessary to split the patients into two classes. Basically, P-Net gives back as output a score for each patient at the end of each MCCV iteration and we can rank them with respect to the phenotype under study. After the ranking step, we need a score threshold to split the patients into the classes. If the score of the patient is below the threshold we assign the patient to the group GP, if the score is above the threshold we assign the patient to the group PP. To find the \textit{optimal score threshold} we compute the scores on the training set and we test an arbitrary vector of quantiles. Each quantile is used as threshold to classify the patients (the classification is based on the scores computed on the training set) and to compute the corresponding accuracy. At the end we can select the \textit{optimal score threshold}, which is equal to the quantile that returns the best accuracy. 

\item The score threshold is used to classify each patient according to its score computed on the test set and to obtain the \textit{predicted label}. 

\item We compare the \textit{predicted labels} and the \textit{true labels}, so we can compute the above mentioned performance measures. 
\end{enumerate}

In the reference paper \cite{winter2012google}, Support Vector Machines (SVM) are used as classifier and SVMs are combined with different kinds of feature selection methods. At the end of the 1000 rounds of MCCV the accuracy can be computed for each feature selection method tested and we can compare our approach, based on T-test as feature selection method followed by P-Net, with respect to the results in the reference paper. 

\subsection{Results of the procedure following strictly the experimental set-up} \label{exp_ggc_extensive}

A series of experiments is showed in this section, where we follow the experimental set-up explained in subsection \ref{setup_ggc}.\\
More precisely, in these experiments we selected the p-step Random Walk Kernel among all the kernel methods available to build the kernelized matrix \textbf{\textit{K}}. This choice is not random but it is based on some preliminary experiments that we ran on a different dataset called TRANSBIG, which is used in the paper \cite{haibe2012three}. TRANSBIG contains microarray data from patients afflicted by breast cancer. The results of these experiments are not showed in this thesis but they proved that the \textit{p-step Random Walk Kernel} is a good choice to capture the topological characteristics of a patient network. In the following experiments we test different number of steps (\textit{p}=1, 2, 3, 4, 8, 10, 15, 20, 50) for the Random Walk and this is fundamental to tune.\\
For each number of steps \textit{p} we want to test the performances of the P-Net algorithm that are also related to the \textit{training ratio}, which is the ratio of patients belonging to the training set. It is well-known that higher is the dimension of the training set and better are the results because the algorithm has more information to exploit to build a good predictor \cite{beleites2013sample}.\\  
The type of \textit{kernelized score function} used to assign a score to each patient is another parameter to tune. So, we make the experiments with all the possible score functions described in subsection \ref{step4}.

\begin{figure}[H]
\centering
\includegraphics[scale=0.7]{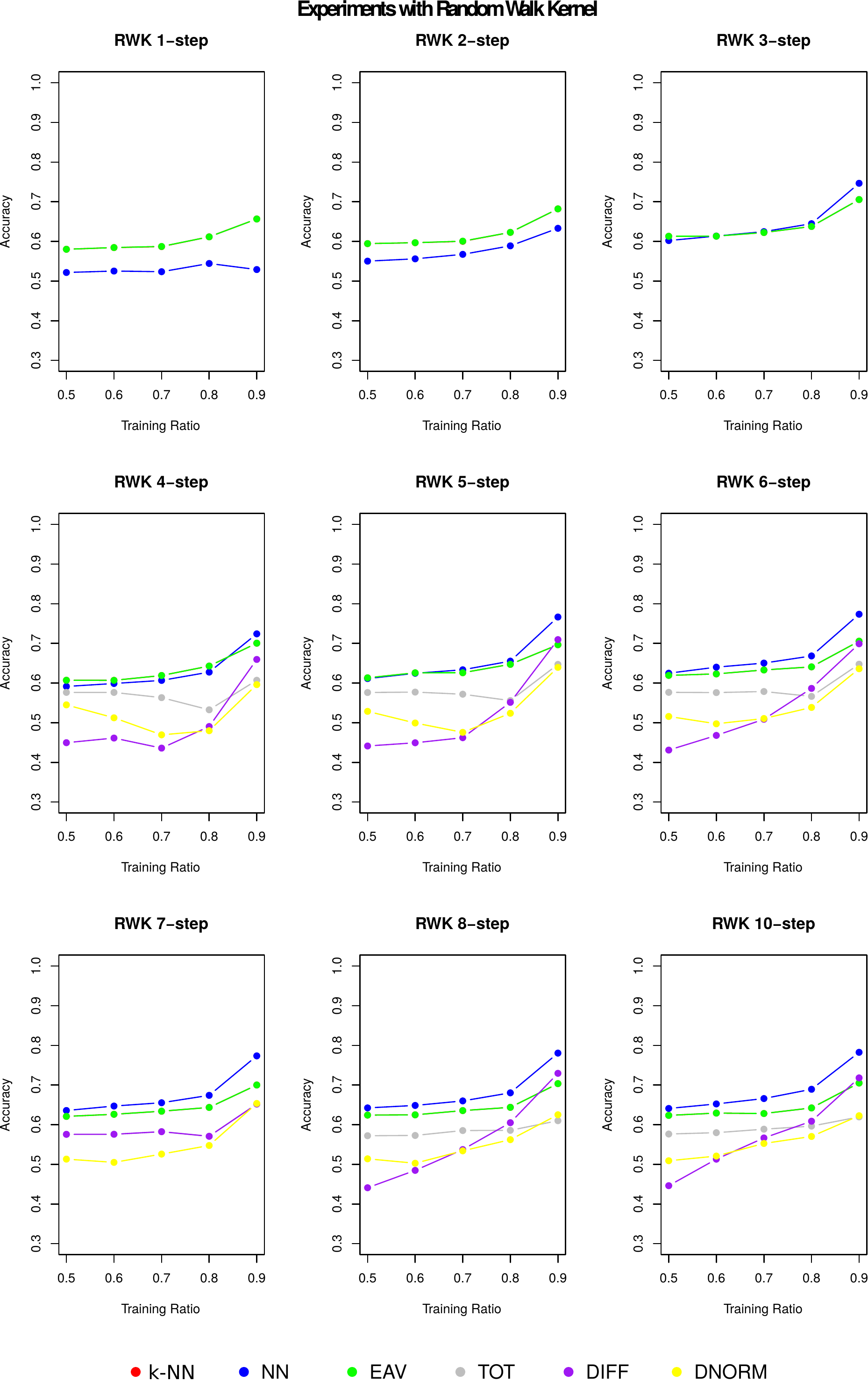}
\caption [Experiments with p-step Random Walk Kernel, where p=1,2,3,4,5,6,7,8,10]{\textbf{Experiments with p-step Random Walk Kernel, where p=1, 2, 3, 4, 5, 6, 7, 8, 10.} For each value of \textit{p} we plot the training ratio in the x-axis and the accuracy in the y-axis. Each line in the plot corresponds to a different kernelized score function.} \label{fig_rwk1-10}
\end{figure}

In \textbf{summary}, we make a series of experiments where we try to tune the above-mentioned parameters (number of steps in the Random Walk and score function) for different training set dimensions.\\
In the figure \ref{fig_rwk1-10} we can see the accuracy achieved by P-Net with respect to the different training ratio and steps \textit{p} of the Random Walk Kernel. Every line is related to the use of a different \textit{kernelized score function}. As we can see, the accuracy improves with the increasing of the training ratio and P-Net achieves the best results with a training ratio of 0.9, which corresponds to 28 samples in the training set. However, this is a predictable result because with a bigger training set the algorithm has more information to build an efficient predictor. \\
We can also observe that k-NN score, NN score and Average score seem to be the best kernelized score functions. When we use the Random walk kernel with \textit{p}=1,2,3, the k-NN and average score achieve the best results (almost the same results between them), but when we increase the value of \textit{p} the NN score is the one with the best results. In this context the kernelized score functions \textit{Total score}, \textit{Differential score} and \textit{Differential normalized score} achieve worst results with respect to the group of score functions (\textit{Average score}, \textit{Nearest Neighbour score}, \textit{k-Nearest Neighbour score}), which use only the positive patients $V_{C}$ to compute the scores. In the microarray data we usually have the gene expression level from few patients, so we can build little networks. When we have few samples it is important to exploit all of them and the use of both \textit{positive} and \textit{negative} patients is designed to reach this purpose. However, in our kernel matrix \textit{\textbf{K}} we have elements $k_{ij}$ really similar each other and the use of both the groups of patients can introduce some ``noise'' in the computation of the scores, leading to the wrong classification of the samples.\\
We can also see how the \textit{Average score} and the \textit{k-Nearest Neighbour score} lead to equal or similar results. This is related to the definition of these two functions. In the Average score we assign a score to the patient \textit{i} considering only the positive neighbours of the patient and then we average their weights. In the k-NN score  we sum the weights of the \textit{k} nearest positive neighbours of the considered patient. The positive patients considered in the Average score and the subset of \textit{k} positive examples considered in the k-NN score can be the same  and, in this case, we obtain the same score from both the functions. The fact that the results of these score functions are equal or similar depends on the topology of the graph and, more precisely, on the value of \textit{k} we choose for the k-NN function. For example, if the node \textit{A} has 3 positive neighbours and we choose $k=3$ it is possible that the k-NN score function selects the same neighbours of the Average function and we obtain the same scores for the node A. 

\begin{figure}[H]
\centering
\includegraphics[scale=0.6]{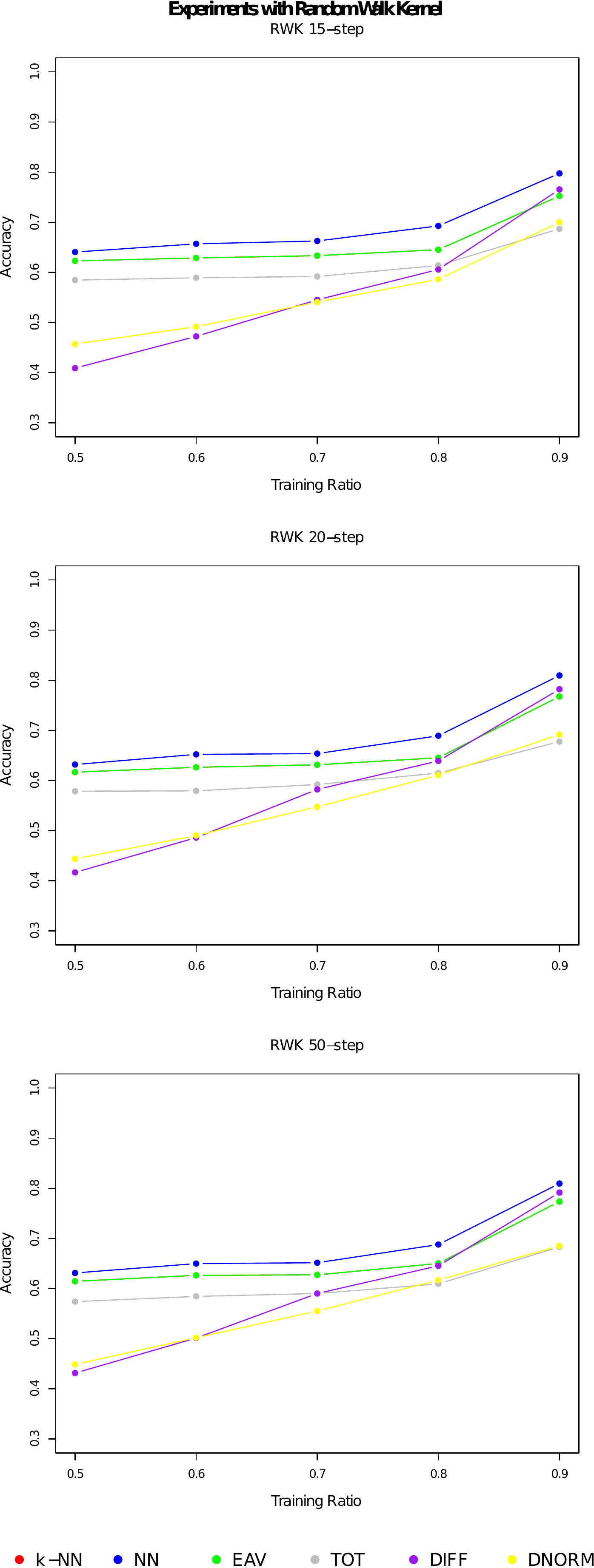}
\caption[Experiments with p-step Random Walk Kernel, where p=15,20,50]{\textbf{Experiments with p-step Random Walk Kernel, where p=15, 20, 50.} For each value of \textit{p} we plot the training ratio in the x-axis and the accuracy in the y-axis. Each line in the plot corresponds to a different kernelized score function.} \label{fig_rwk15-50}
\end{figure}

In the precedent figure, there is a slight but significant increase in the accuracy value when we increase the number of steps. In any case this situation cannot continue endlessly and the algorithm has to reach a convergence point sooner or later. Indeed, the Random Walk tends to a stationary distribution when $p \rightarrow \infty $ , if the graph \textit{G} is not bipartite \cite{asz80random}. To proof the convergence of the \textit{p-step Random Walk Kernel} we try to further increase the value of \textit{p} and the results are showed in figure \ref{fig_rwk15-50}. As we can see, the accuracy does not improve with the increase of the number of steps to 15, 20, and 50. The above observations about the behaviour of the \textit{kernelized score functions} remain unchanged. 

\begin{figure}[H]
\centering
\includegraphics[scale=0.6]{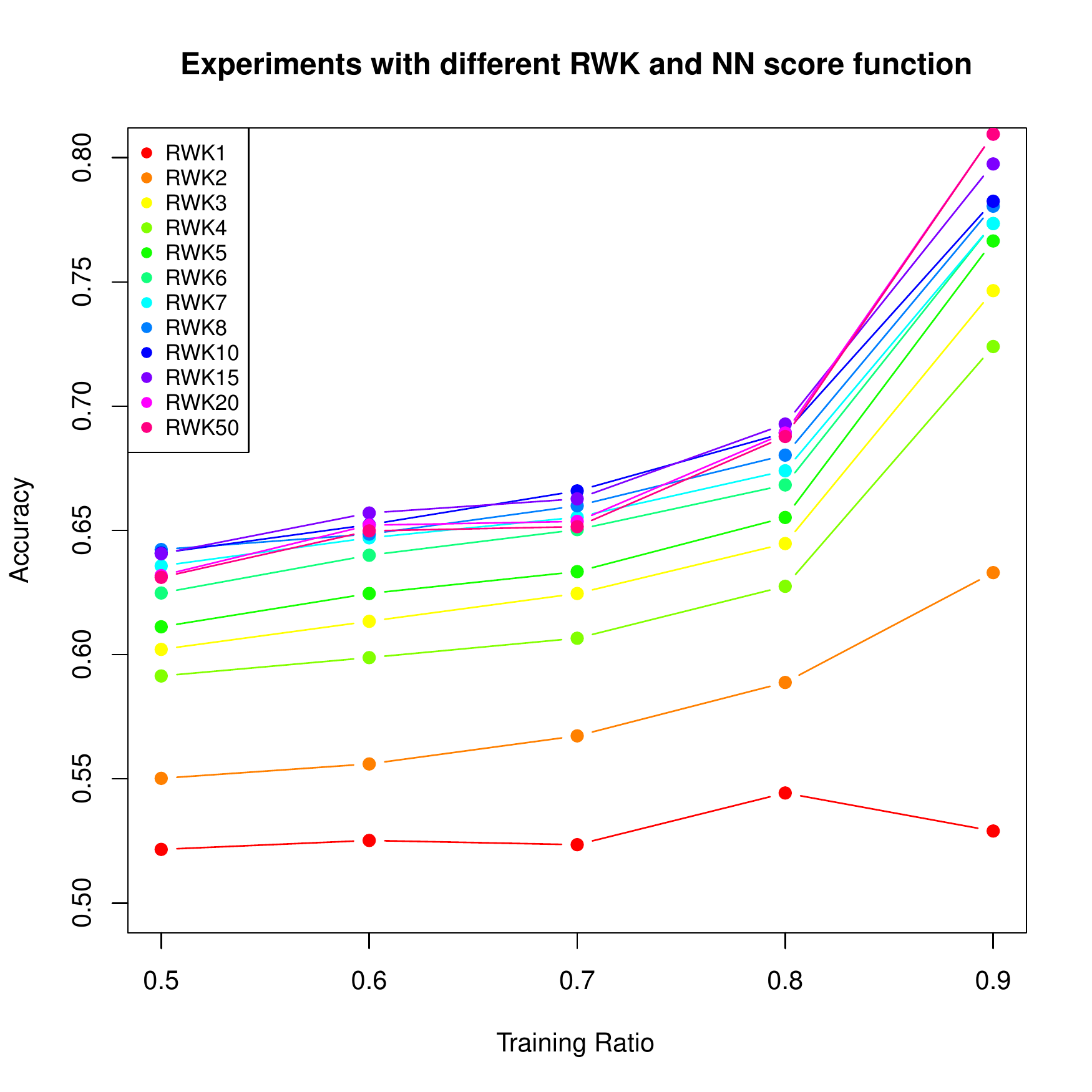}
\caption[Experiments with different steps of the Random Walk Kernel and Nearest Neighbour score function]{\textbf{Experiments with different steps of the Random Walk Kernel and Nearest Neighbour score function.} We plot the training ratio in the x-axis and the accuracy in the y-axis.} \label{fig_comparison_Pnet}
\end{figure}

If we consider only the experiments where we use the \textit{Nearest Neighbour score function}, which is the function with the best results, we can plot all the tests in the same plot (figure \ref{fig_comparison_Pnet}). 

In the precedent figure we can clearly see how the accuracy achieved from P-Net increases until $p=8$ and then the performances stabilize and we do not see any improvement in the accuracy. 

\subsection{Convergence of the p-step Random Walk Kernel}

\begin{figure}[H]
\centering
\includegraphics[scale=0.5]{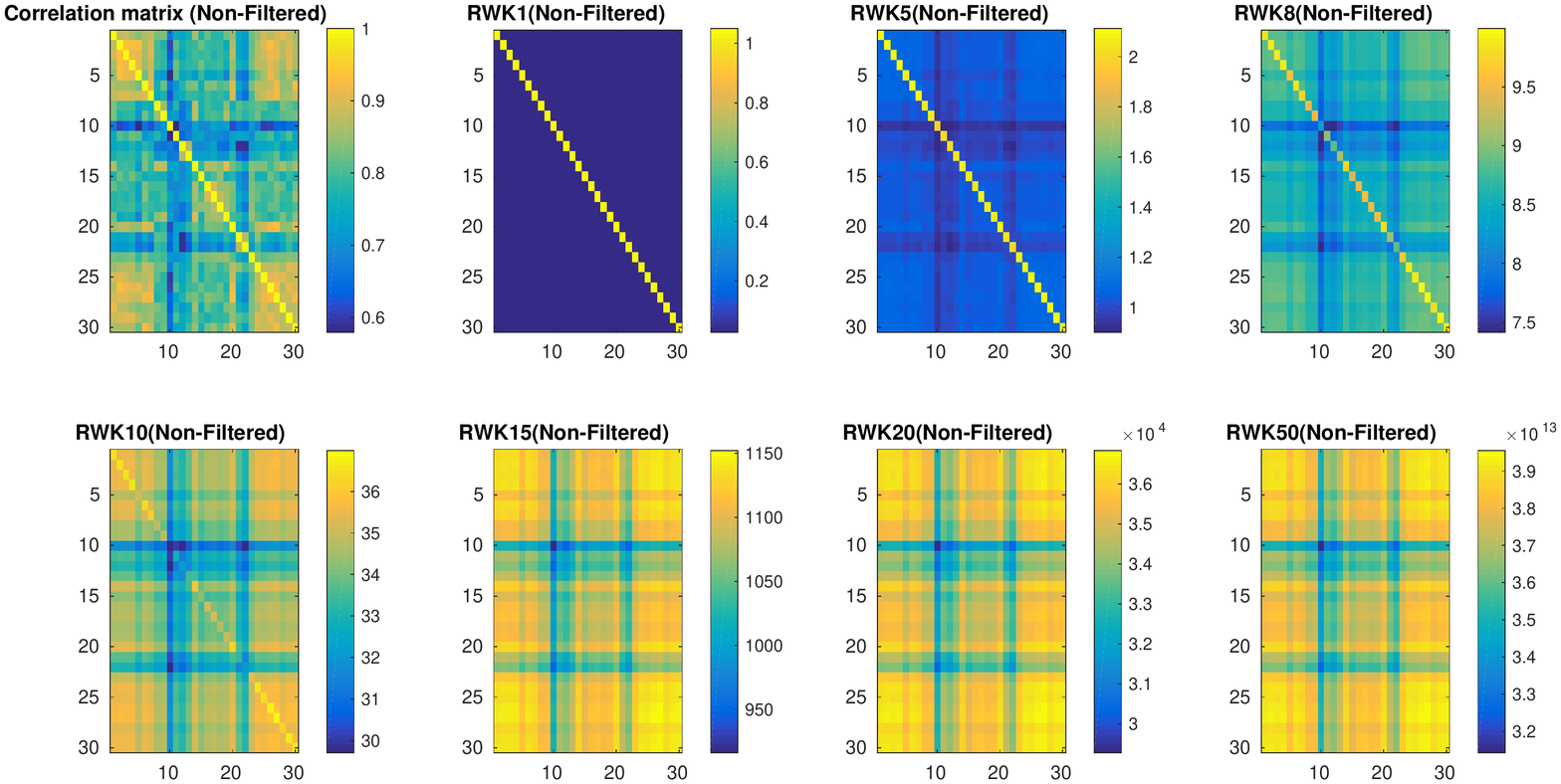}
\caption[Plots of the matrices \textit{\textbf{W}} and \textbf{\textit{K}} with Imagesc and without the feature selection step]{\textbf{Plots of the matrices \textit{\textbf{W}} and \textbf{\textit{K}} with Imagesc and without the feature selection step.} For each matrix \textbf{\textit{K}} a different number of steps for the RWK is used.} 
\label{fig_conv_allfeat}
\end{figure}

\begin{figure}[H]
\centering
\includegraphics[scale=0.5]{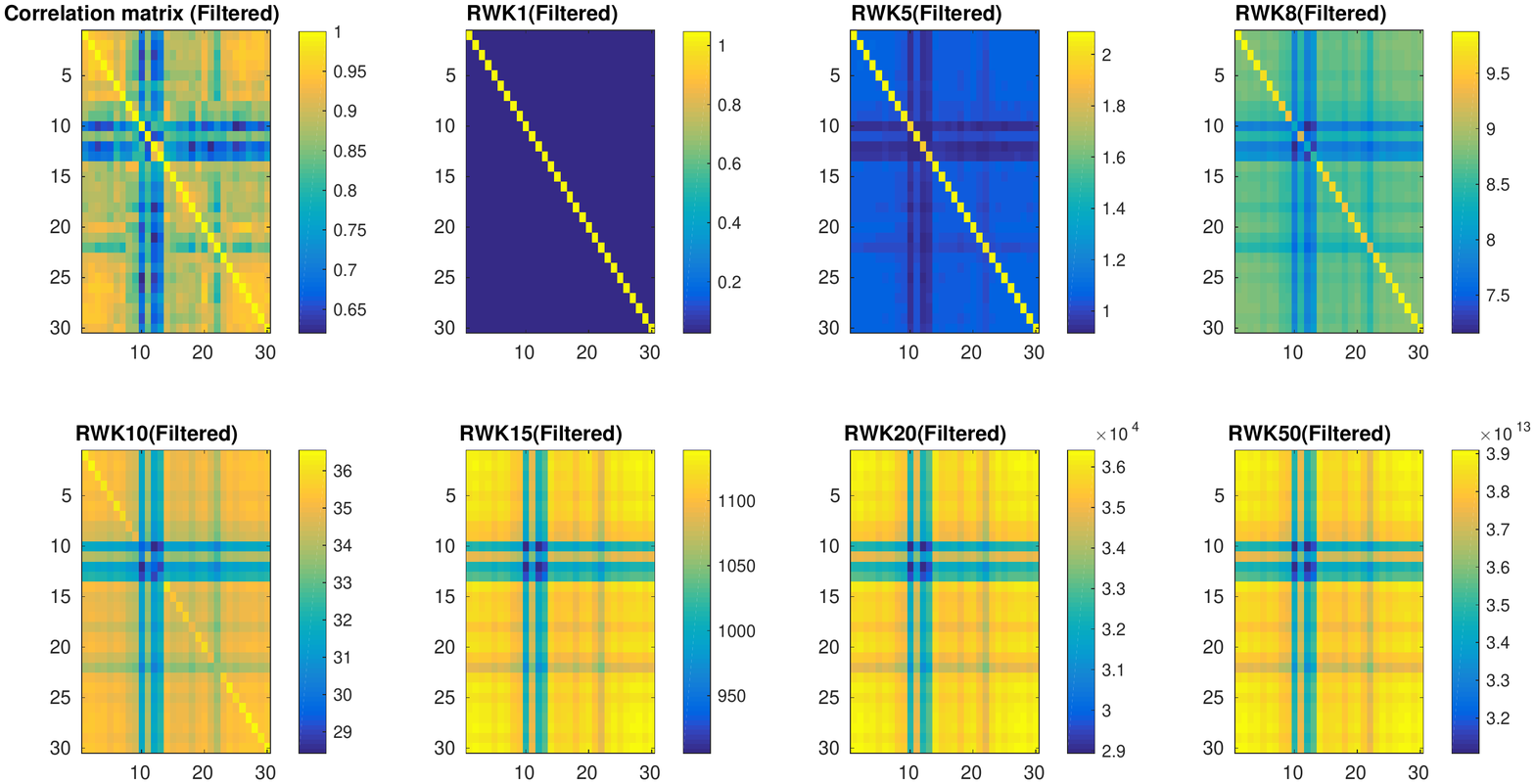}
\caption[Plots of the matrices \textit{\textbf{W}} and \textbf{\textit{K}} with Imagesc and the feature selection step]{\textbf{Plots of the matrices \textit{\textbf{W}} and \textbf{\textit{K}} with Imagesc and the feature selection step.} For each matrix \textbf{\textit{K}} a different number of steps for the RWK is used.} 
\label{fig_conv_1000feat}
\end{figure}

From the results showed in the subsection \ref{exp_ggc_extensive}, we can realise that the p-step Random Walk Kernel tends to reach a convergence point as we expect from the literature \cite{asz80random}. We realize a series of plots to understand better this phenomenon where we can see how the information in our matrices change with the increasing of the parameter \textit{p}. In substance, we want to show how the weights in our matrices change and which is roughly the point where the information do not change and the algorithm converge. To achieve this goal, we used the \textit{imagesc} function from MATLAB and Statistics Toolbox Release R2016a, The MathWorks, Inc., Natick, Massachusetts, United States. The imagesc function takes a matrix as input and it displays the matrix as an image that uses the range of colours between blue and yellow. In our case, the output image is an \textit{m}$\times$\textit{m} (\textit{m}=30, which is the number of patients) grid where each element of the matrix decides the colour of 1 pixel in the image.\\
We plot the images of the correlation matrix \textbf{\textit{W}} and the Random Walk Kernel matrices \textbf{\textit{K}} with different values of the parameter \textit{p}. In the figure \ref{fig_conv_allfeat} we used all the available features to obtain the matrices, in other words we skipped the feature selection step by T-test. In the figure \ref{fig_conv_1000feat} instead we apply the T-test filter. 

In both the precedent pictures we can see how the weights into the matrices (which are related to the colour of the pixels) change until we use the 10-steps Random Walk Kernel. After this point, the images are really close each other and the information into the matrices is more or less the same. It is a sign that the algorithm reached the convergence point and it is useless to continue to increase the value of \textit{p} because we cannot obtain different results.\\
The precedent hypothesis is confirmed by the table \ref{table_conv_res}, which contains the accuracy values under the Nearest Neighbour kernelized score function (it is the score function with the overall best results) for different training set sizes and steps of RWK. 

\begin{table}[H]
\centering
\resizebox{\textwidth}{!}{%
\begin{tabular}{|c|c|c|c|c|c|}
\hline
\textit{\textbf{Steps of RWK}} & \textit{\textbf{16 samples}} & \textit{\textbf{18 samples}} & \textit{\textbf{20 samples}} & \textit{\textbf{24 samples}} & \textit{\textbf{28 samples}} \\ \hline
\textit{RWK 8-steps} & \cellcolor[HTML]{FCFF2F}64.23\% & 64.85\% & 65.99\% & 68.03\% & 78.05\% \\ \hline
\textit{RWK 10-steps} & 64.09\% & 65.23\% & \cellcolor[HTML]{FCFF2F}66.59\% & 68.93\% & 78.25\% \\ \hline
\textit{RWK 15-steps} & 64.06\% & \cellcolor[HTML]{FCFF2F}65.70\% & 66.27\% & \cellcolor[HTML]{FCFF2F}69.28\% & 79.75\% \\ \hline
\textit{RWK 20-steps} & 63.18\% & 65.21\% & 65.36\% & 68.92\% & \cellcolor[HTML]{FCFF2F}80.95\% \\ \hline
\textit{RWK 50-steps} & 63.11\% & 64.98\% & 65.15\% & 68.78\% & \cellcolor[HTML]{FCFF2F}80.95\% \\ \hline
\end{tabular}%
}
\caption[Accuracies of P-Net using the Nearest-Neighbour score function and different values of \textit{p}]{\textbf{Accuracies of P-Net using the Nearest-Neighbour score function and different values of \textit{p}.} The cells highlighted in yellow contain the best accuracy for each training set size.}
\label{table_conv_res}
\end{table}

The increment in the performances of P-Net is not so related to the number of steps when we consider values of \textit{p} higher than 8. In the table \ref{table_conv_res} the best results for each training set size are highlighted in yellow and we can see how the highest accuracy is not always achieved by the highest number of steps. For example, if we compare the accuracies achieved with \textit{RWK 8-steps} and \textit{RWK 50-steps} we can notice that when we use 50 steps we obtain better results only for 18, 24 and 28 samples and in the first two cases the increase is not so significant ($0.13 \%$ and $0.75 \%$, respectively).

\begin{figure}[H]
\centering
\includegraphics[scale=0.40]{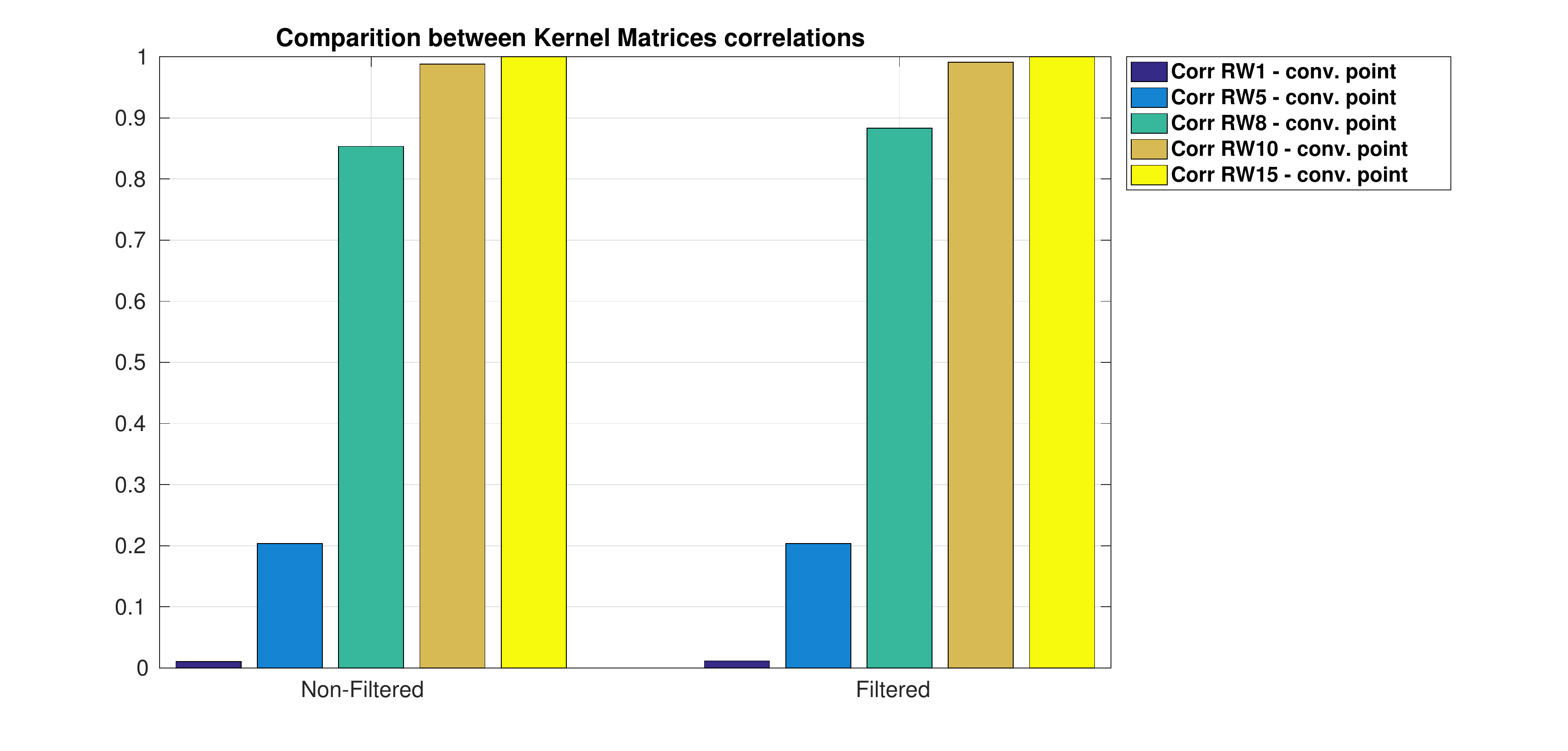}
\caption[Correlation between Kernel matrices and convergence point]{\textbf{Correlation between Kernel matrices and convergence point.} Each rectangle corresponds to the Pearson correlation between one Kernel matrix and the convergence point, where the convergence point is the correlation between the 20-steps RWK and 50-steps RWK. The group of columns on the left considers the non-filtered kernel matrices and the other one considers the filtered kernel matrices.} 
\label{fig_conv_point}
\end{figure}

As further proof of the convergence we create the plot \ref{fig_conv_point} with MATLAB. In this picture we consider as \emph{convergence point} the  Pearson correlation between the Kernel matrices with \textit{p=20} and \textit{p=50}. Then, we compute the correlation between the \textit{convergence point} and all the other matrices and finally we plot the results in a histogram. 

We can see that with the increase of the value of \textit{p} the columns become higher. 10-steps Random Walk Kernel is really close to the convergence point and when $p=15$ the correlation is 1. In this way we can show how the information contained in the kernel matrices changes for small values of \textit{p} and how it remains almost essentially the same for higher values of \textit{p}. 

\subsection{Application of t-SNE to the pancreatic cancer dataset}

The pancreatic cancer dataset contains 30 patients and 9175 genes, after the pre-filtering steps discussed in subsection \ref{setup_ggc}. So, we have to deal with an high-dimensional dataset and, even if we use a graph to structure the data, it is difficult to imagine how the data are located in a two-dimensional or three-dimensional space. The visualisation of the data in a single map is important to understand if there are clusters of patients in our data (for example, if patients with a poor prognosis are closer than patients with a good prognosis). If there are such clusters, the classification problem should be easier to handle.\\
Different methods are available to visualise high-dimensional data points and one possibility is to use a \textit{dimensionality reduction method} \cite{fodor2002survey, van2009dimensionality}. In this kind of methods, the purpose is to reduce the number of features to consider and to convert an high-dimensional dataset to a dataset with two or three dimensions to visualise in a single 2D or 3D map. During this reduction step, we have the necessity to keep the global and local structure of our data points. This is feasible because we start from the assumption that not all the features have the same importance for the description of the data structure.

The chosen dimensionality reduction method is \textbf{\textit{t-SNE}}, which stands for \textit{t-Distributed Sto-\\chastic Neighbour Embedding} \cite{maaten2008visualizing}. In the crowd of available techniques we choose t-SNE because of the following documented advantages over the most known methods:

\begin{enumerate}
\item t-SNE is a non-linear technique able to preserve the local structure of the data (in other words, it tends to keep closer similar data points in the low-dimensional representation).
\item It shows good results in the representation of real high-dimensional datasets, instead many methods show good results on artificial data and they fail on real datasets. 
\item It is able to capture the global and local structures of the data at the same time. 
\end{enumerate}

\begin{figure} [H]
\centering
  \begin{subfigure}[H]{1\textwidth}
    \includegraphics[width=\textwidth]{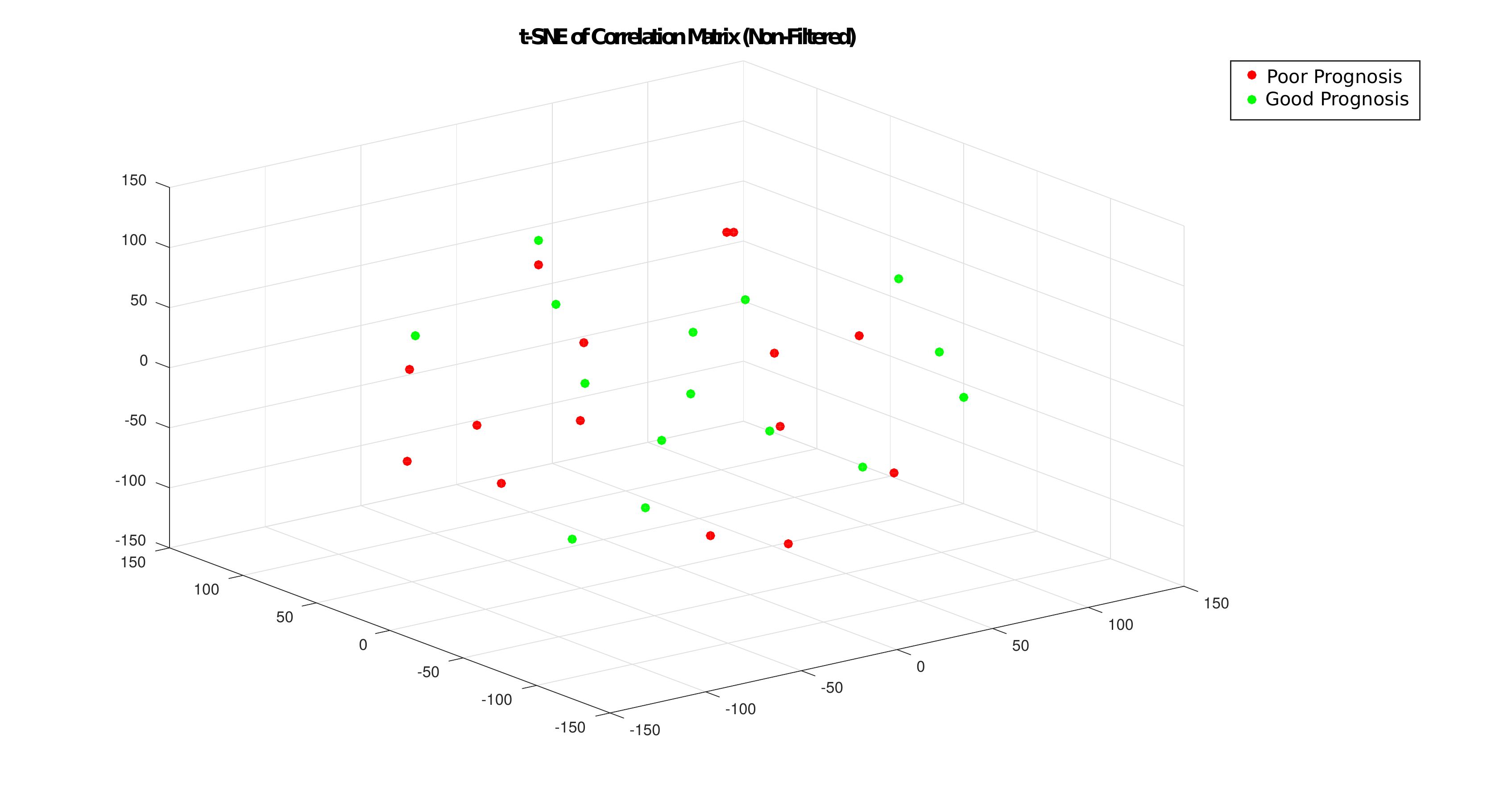}
  \end{subfigure}
  \begin{subfigure}[H]{1\textwidth}
    \includegraphics[width=\textwidth]{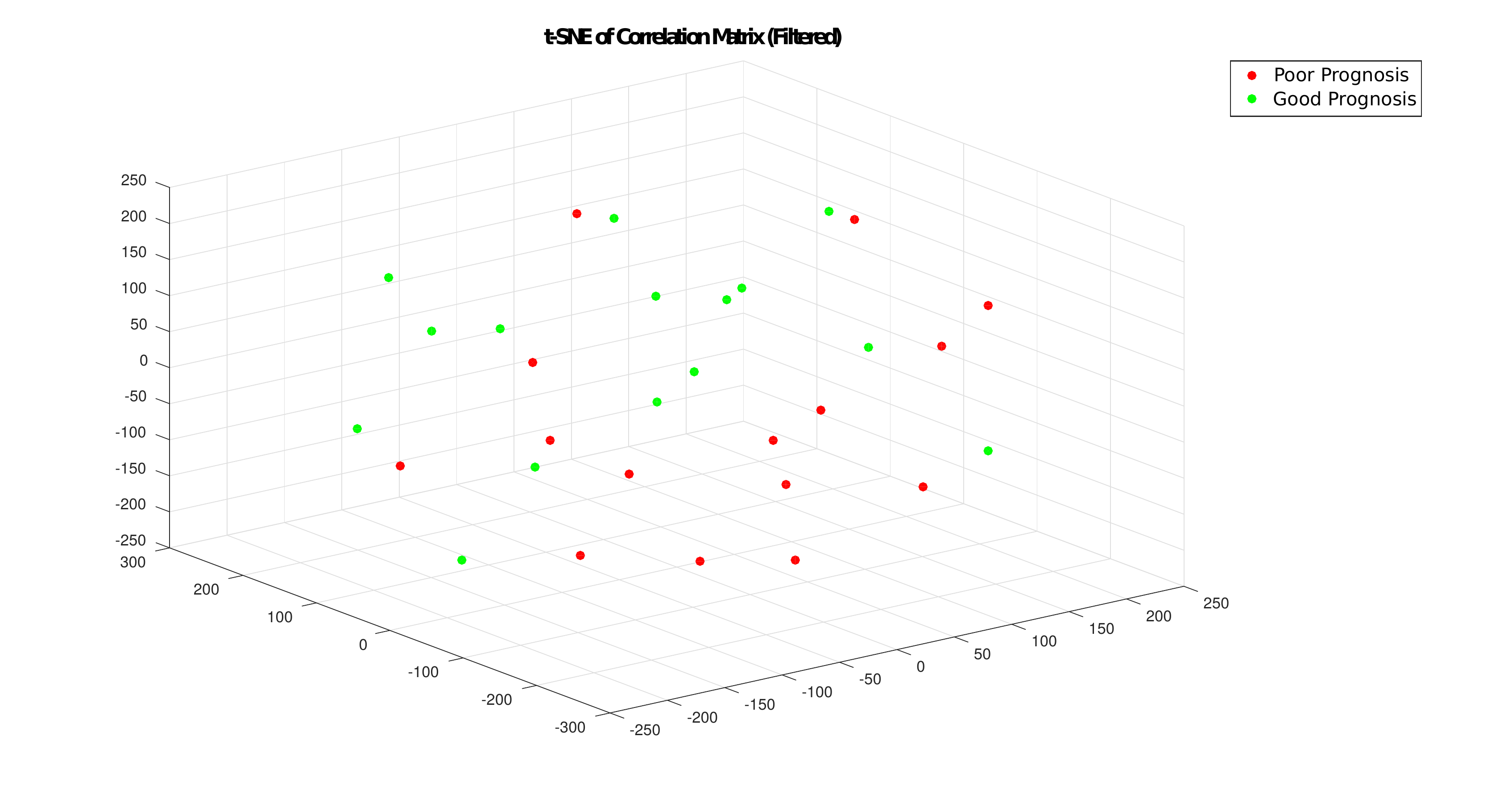}
  \end{subfigure}
  \caption[t-SNE applied on the correlation matrices \textit{\textbf{W}} with all the available features and after filtering by T-test]{\textbf{t-SNE applied on the correlation matrices \textit{\textbf{W}} with all the available features (Top figure) and after filtering by T-test (Bottom figure).}}
  \label{fig_tSNE_Wmatrix}
\end{figure}

In the figure \ref{fig_tSNE_Wmatrix} we can see the spatial distribution of the data points, where each point corresponds to one patient, from the correlation matrix \textit{\textbf{W}} in two conditions: using all the available 9175 features and after the filtering step by T-test, which selects the first 1000 features with the lowest p-value. 

We can see how the spatial distribution of the data changes with the application of the feature selection method on the correlation matrix. The data points from patients with a poor prognosis seem to be closer each other after the application of the feature selection step. This can be an advantage in the context of classification tasks and prioritisation of patients with respect to a given phenotype.  

\begin{figure}[H]
\includegraphics[scale=0.45]{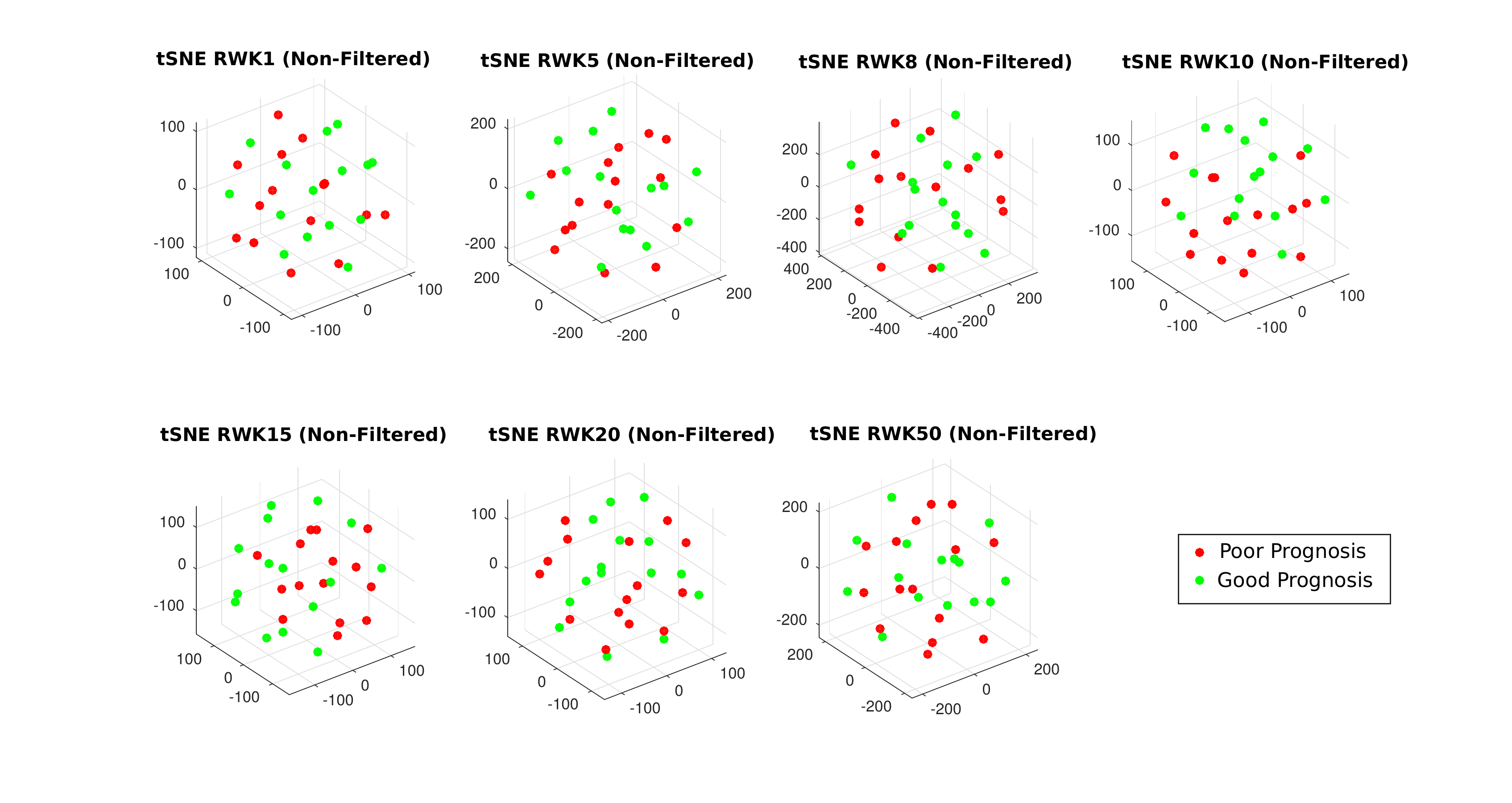}
\caption [t-SNE applied on the Random Walk Kernel matrices with different values of \textit{p} without filtering step]{\textbf{t-SNE applied on the Random Walk Kernel matrices with different values of \textit{p} without filtering step.}} 
\label{fig_tSNE_Kmatrices_allfeat}
\end{figure}

\begin{figure}[H]
\includegraphics[scale=0.45]{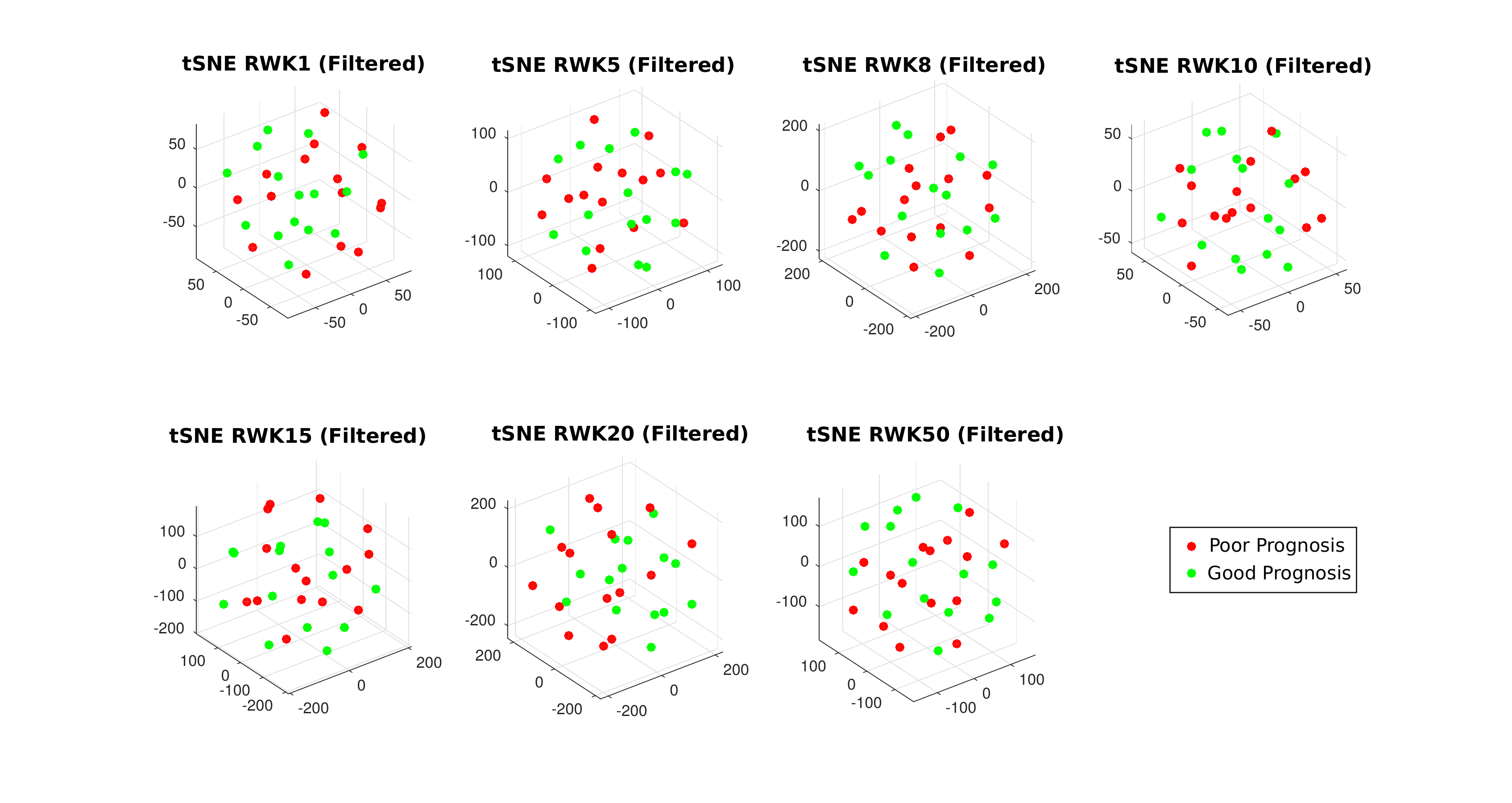}
\caption[t-SNE applied on the Random Walk Kernel matrices with different values of \textit{p} with filtering step]{\textbf{t-SNE applied on the Random Walk Kernel matrices with different values of \textit{p} with filtering step.}} 
\label{fig_tSNE_Kmatrices_1000feat}
\end{figure}

However, for the pancreatic cancer dataset we exploited the Kernel matrices so it is interesting to visualise the spatial distribution of the data points from this kind of matrices. 
In the figures \ref{fig_tSNE_Kmatrices_allfeat} and \ref{fig_tSNE_Kmatrices_1000feat} there is the output of t-SNE for the kernel matrices \textit{\textbf{K}} applying different steps of the p-step Random Walk Kernel; also in this case we consider the two conditions ``non-filtered'' and ``filtered''. We can see again how the distribution of the labelled patients changes but it is more difficult to spot some clear group in the data.

The fact that we cannot spot defined groups with the same label, even after the application of the Random Walk Kernel, is an index of the intrinsic difficulty of the classification task on this specific dataset for P-Net. If we consider all the available kernelized score functions (see \ref{step4}), the computation of the score for the patient \textit{i} is always based on the weight of the edges of its neighbours $k_{ij}$. So, if the patients belonging to the same class are close, it is easier that the algorithm assigns the correct label to the patient \textit{i} .  

\subsection{Comparison with the performances of the methods assessed in the reference paper}

In the reference paper \cite{winter2012google} five different methods, for ranking genes to find a signature able to discriminate between the two classes of interest (good prognosis and poor prognosis), are tested. Then, for each method an SVM predictor is trained using only the 5-10 top ranked genes as features and the predictor is used to classify the patients.\\
In the table \ref{table_comparison} the accuracies and corresponding standard errors of the methods are showed for different training set sizes. The assessed methods are: Pearson correlation, Spearman correlation, Fold change, t-test, Random (selection of random genes as a control method) and NetRank. There are also the results obtained with two variants of NetRank: Direct Neighbour and Constant c.

\begin{table}[H]
\centering
\resizebox{12cm}{!}{
\begin{tabular}{cccccc}
\hline
\multicolumn{1}{|c|}{\textbf{Training set size}}    & \multicolumn{1}{c|}{16}      & \multicolumn{1}{c|}{18}      & \multicolumn{1}{c|}{20}      & \multicolumn{1}{c|}{24}      & \multicolumn{1}{c|}{28}      \\ \hline
\multicolumn{1}{l}{}                                & \multicolumn{1}{l}{}         & \multicolumn{1}{l}{}         & \multicolumn{1}{l}{}         & \multicolumn{1}{l}{}         & \multicolumn{1}{l}{}         \\ \hline
\multicolumn{1}{|c|}{\textbf{Ranking method}}       & \multicolumn{5}{c|}{\textbf{Accuracies}}                                                                                                                 \\ \hline
\multicolumn{1}{|c|}{\textit{Pearson correlation}}  & \multicolumn{1}{c|}{56.47\%} & \multicolumn{1}{c|}{58.63\%} & \multicolumn{1}{c|}{60.82\%} & \multicolumn{1}{c|}{65.87\%} & \multicolumn{1}{c|}{70.55\%} \\ \hline
\multicolumn{1}{|c|}{\textit{Spearman correlation}} & \multicolumn{1}{c|}{54.57\%} & \multicolumn{1}{c|}{55.35\%} & \multicolumn{1}{c|}{57.41\%} & \multicolumn{1}{c|}{61.20\%} & \multicolumn{1}{c|}{68.95\%} \\ \hline
\multicolumn{1}{|c|}{\textit{Fold change}}          & \multicolumn{1}{c|}{47.21\%} & \multicolumn{1}{c|}{46.33\%} & \multicolumn{1}{c|}{46.27\%} & \multicolumn{1}{c|}{44.38\%} & \multicolumn{1}{c|}{39.45\%} \\ \hline
\multicolumn{1}{|c|}{\textit{t-test}}               & \multicolumn{1}{c|}{54.74\%} & \multicolumn{1}{c|}{54.88\%} & \multicolumn{1}{c|}{56.00\%} & \multicolumn{1}{c|}{58.43\%} & \multicolumn{1}{c|}{65.00\%} \\ \hline
\multicolumn{1}{|c|}{\textit{Random}}               & \multicolumn{1}{c|}{54.51\%} & \multicolumn{1}{c|}{55.38\%} & \multicolumn{1}{c|}{56.06\%} & \multicolumn{1}{c|}{55.33\%} & \multicolumn{1}{c|}{57.25\%} \\ \hline
\multicolumn{1}{|c|}{\textit{NetRank}}              & \multicolumn{1}{c|}{62.63\%} & \multicolumn{1}{c|}{63.53\%} & \multicolumn{1}{c|}{64.60\%} & \multicolumn{1}{c|}{68.63\%} & \multicolumn{1}{c|}{72.30\%} \\ \hline
\multicolumn{1}{|c|}{\textit{Direct neighbour}}      & \multicolumn{1}{c|}{56.70\%} & \multicolumn{1}{c|}{58.73\%} & \multicolumn{1}{c|}{61.02\%} & \multicolumn{1}{c|}{66.10\%} & \multicolumn{1}{c|}{70.55\%} \\ \hline
\multicolumn{1}{|c|}{\textit{Constant c}}           & \multicolumn{1}{c|}{63.54\%} & \multicolumn{1}{c|}{63.82\%} & \multicolumn{1}{c|}{64.34\%} & \multicolumn{1}{c|}{65.15\%} & \multicolumn{1}{c|}{63.05\%} \\ \hline
\multicolumn{1}{|c|}{\textit{P-Net (RWK8, NN)}}     & \multicolumn{1}{c|}{64.23\%} & \multicolumn{1}{c|}{64.85\%} & \multicolumn{1}{c|}{65.99\%} & \multicolumn{1}{c|}{68.03\%} & \multicolumn{1}{c|}{78.05\%} \\ \hline
\multicolumn{1}{l}{}                                & \multicolumn{1}{l}{}         & \multicolumn{1}{l}{}         & \multicolumn{1}{l}{}         & \multicolumn{1}{l}{}         & \multicolumn{1}{l}{}         \\
\multicolumn{1}{l}{}                                & \multicolumn{1}{l}{}         & \multicolumn{1}{l}{}         & \multicolumn{1}{l}{}         & \multicolumn{1}{l}{}         & \multicolumn{1}{l}{}         \\ \hline
\multicolumn{1}{|c|}{\textbf{Ranking method}}       & \multicolumn{5}{c|}{\textbf{Standard errors}}                                                                                                            \\ \hline
\multicolumn{1}{|c|}{\textit{Pearson correlation}}  & \multicolumn{1}{c|}{0.39\%}  & \multicolumn{1}{c|}{0.41\%}  & \multicolumn{1}{c|}{0.45\%}  & \multicolumn{1}{c|}{0.60\%}  & \multicolumn{1}{c|}{0.97\%}  \\ \hline
\multicolumn{1}{|c|}{\textit{Spearman correlation}} & \multicolumn{1}{c|}{0.38\%}  & \multicolumn{1}{c|}{0.41\%}  & \multicolumn{1}{c|}{0.47\%}  & \multicolumn{1}{c|}{0.58\%}  & \multicolumn{1}{c|}{1.02\%}  \\ \hline
\multicolumn{1}{|c|}{\textit{Fold change}}          & \multicolumn{1}{c|}{0.34\%}  & \multicolumn{1}{c|}{0.37\%}  & \multicolumn{1}{c|}{0.42\%}  & \multicolumn{1}{c|}{0.53\%}  & \multicolumn{1}{c|}{1.04\%}  \\ \hline
\multicolumn{1}{|c|}{\textit{t-test}}               & \multicolumn{1}{c|}{0.39\%}  & \multicolumn{1}{c|}{0.42\%}  & \multicolumn{1}{c|}{0.47\%}  & \multicolumn{1}{c|}{0.61\%}  & \multicolumn{1}{c|}{1.10\%}  \\ \hline
\multicolumn{1}{|c|}{\textit{Random}}               & \multicolumn{1}{c|}{0.40\%}  & \multicolumn{1}{c|}{0.40\%}  & \multicolumn{1}{c|}{0.46\%}  & \multicolumn{1}{c|}{0.58\%}  & \multicolumn{1}{c|}{1.08\%}  \\ \hline
\multicolumn{1}{|c|}{\textit{NetRank}}              & \multicolumn{1}{c|}{0.34\%}  & \multicolumn{1}{c|}{0.38\%}  & \multicolumn{1}{c|}{0.43\%}  & \multicolumn{1}{c|}{0.56\%}  & \multicolumn{1}{c|}{0.99\%}  \\ \hline
\multicolumn{1}{|c|}{\textit{Direct neighbour}}      & \multicolumn{1}{c|}{0.39\%}  & \multicolumn{1}{c|}{0.41\%}  & \multicolumn{1}{c|}{0.46\%}  & \multicolumn{1}{c|}{0.60\%}  & \multicolumn{1}{c|}{0.97\%}  \\ \hline
\multicolumn{1}{|c|}{\textit{Constant c}}           & \multicolumn{1}{c|}{0.32\%}  & \multicolumn{1}{c|}{0.33\%}  & \multicolumn{1}{c|}{0.38\%}  & \multicolumn{1}{c|}{0.56\%}  & \multicolumn{1}{c|}{1.04\%}  \\ \hline
\multicolumn{1}{|c|}{\textit{P-Net (RWK8, NN)}}     & \multicolumn{1}{c|}{2.02\%} & \multicolumn{1}{c|}{2.23\%} & \multicolumn{1}{c|}{2.38\%} & \multicolumn{1}{c|}{3.38\%} & \multicolumn{1}{c|}{7.38\%} \\ \hline
\end{tabular}
}
\caption[Comparison table for the methods applied on the pancreatic cancer dataset]{\textbf{Comparison table for the methods applied on the pancreatic cancer dataset.}}
\label{table_comparison}
\end{table}

\begin{figure}[H]
\centering
\includegraphics[scale=0.7]{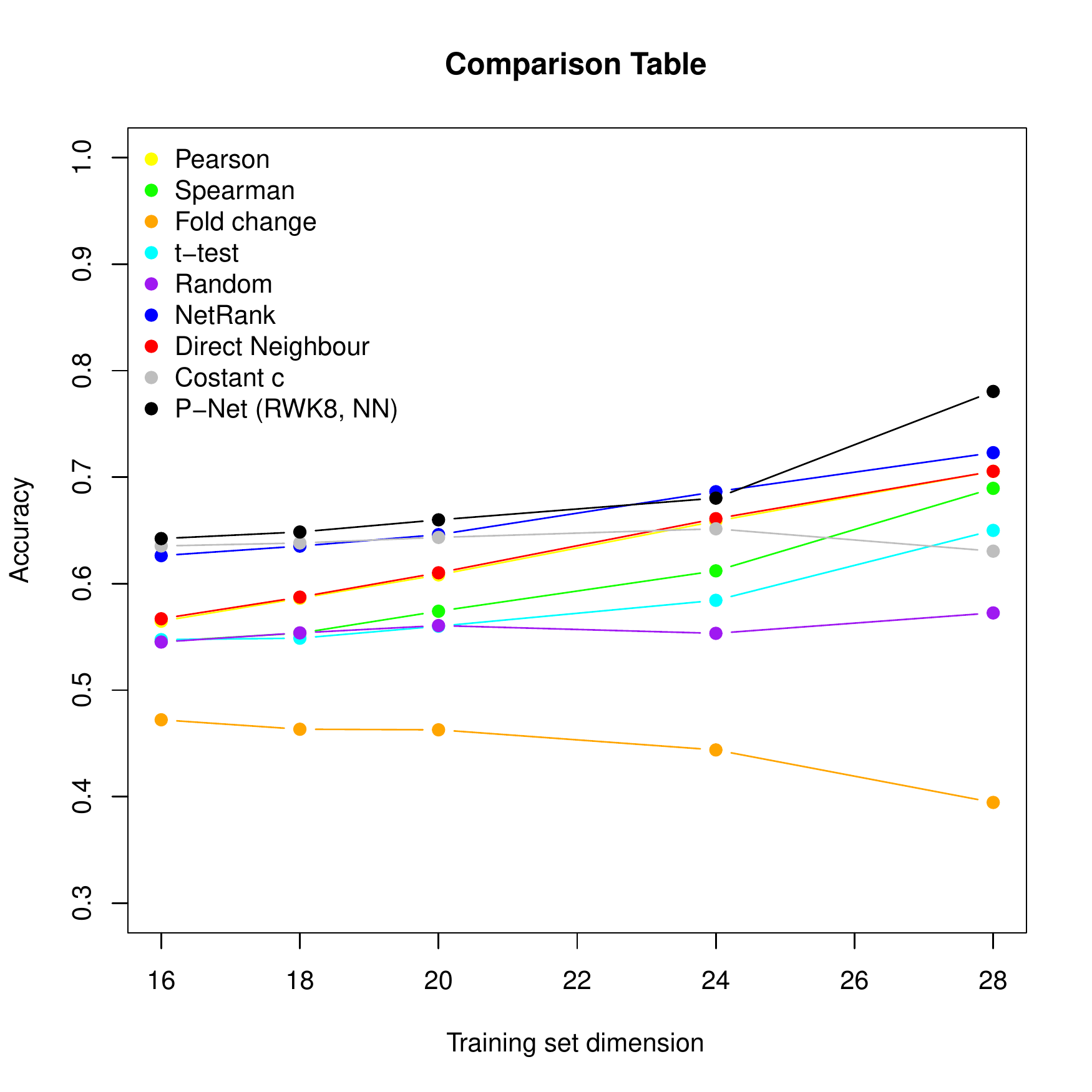}
\caption[Plot to compare the accuracies of different methods for the pancreatic cancer dataset]{\textbf{Plot to compare the accuracies of different methods for the pancreatic cancer dataset.}} \label{fig_comparison_plot}
\end{figure}

The accuracies showed in table \ref{table_comparison} are plotted in the figure \ref{fig_comparison_plot} to have a clearer overview of the results. The P-Net algorithm with 8-step Random Walk Kernel and Nearest Neighbour score function achieves better results than the other methods for each training set dimension if we consider only the accuracy values. The only exception is when the training set consists of 24 patients.
In this case, P-Net shows a slightly worst result with respect to the leader method NetRank and the difference between them is $0.6 \%$. P-Net achieves slightly better results with respect to Net-Rank for the other training set dimensions and the accuracy of P-Net is $78.05 \%$ ($5.75 \%$ higher than NetRank) with a training set of 28 patients.\\
Even if our accuracies are higher than the other methods, we can see from the second table in \ref{table_comparison} that our standard errors (also called Standard Error of the Mean - SEM) are higher than the NetRank ones. So, we can conclude that P-Net is less precise than NetRank.\\
Finally, if we consider both accuracy and standard error of P-Net we can say that P-Net shows comparable results with respect to NetRank (the best method used in the reference paper).  

\subsection{P-Net without the T-test as feature selection method}

\begin{figure}[H]
\centering
\includegraphics[scale=0.4]{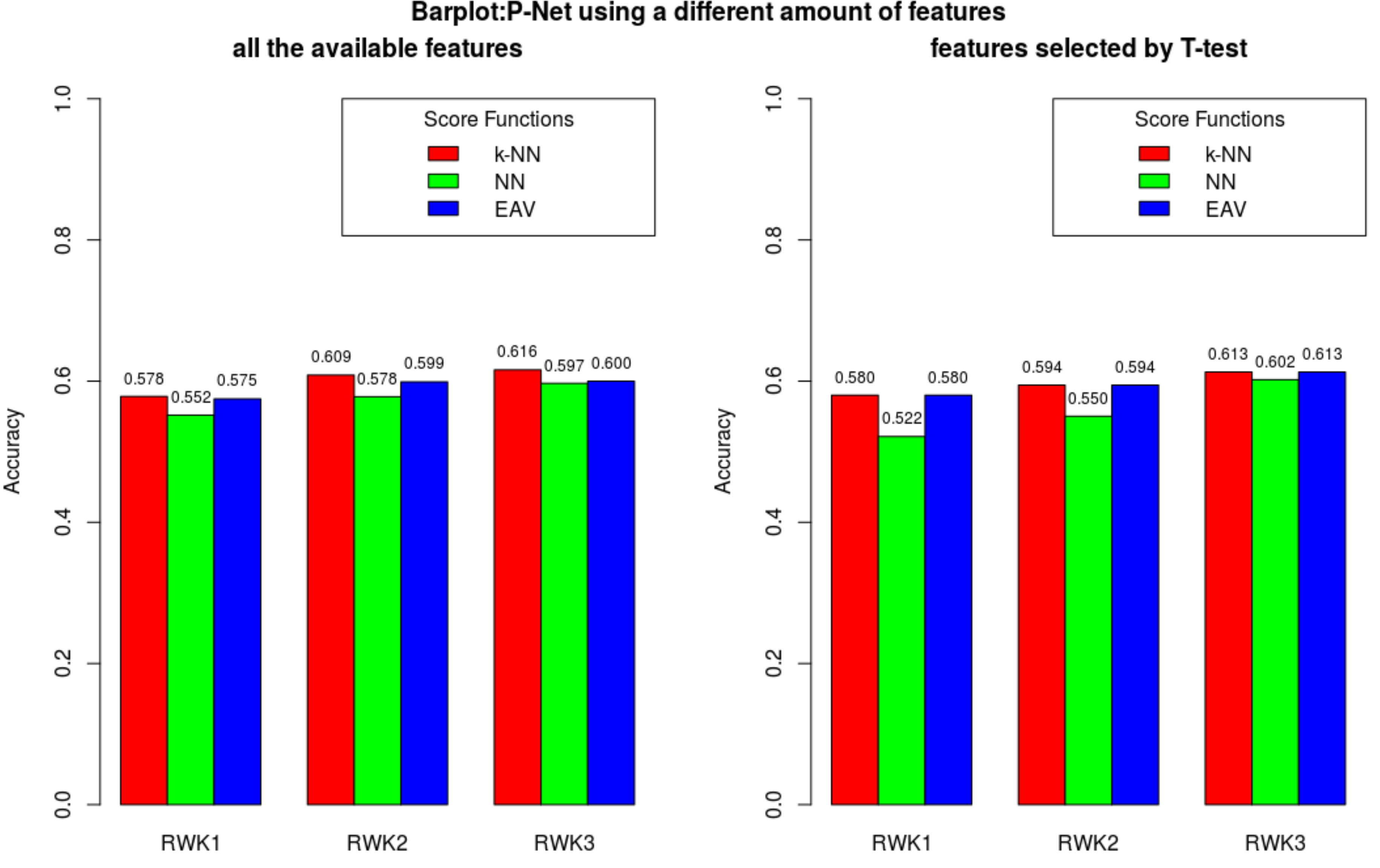}
\caption[Barplot to compare the performances of P-Net for a different amount of features]{\textbf{Barplot to compare the performances of P-Net using different amount of features and different steps for the Random Walk Kernel.}} \label{fig_barplot_pnetallfeat_rwk}
\end{figure}

In the experimental set-up explained in  \ref{setup_ggc} we used the T-test as univariate filter to select the most important features/genes to discriminate between the two considered classes \textit{good prognosis} and \textit{poor prognosis}. In this subsection, we want to assess the impact of this feature selection step on the performances of P-Net. Our purpose is to understand if the selection of a subset of genes, which are the most discriminative in theory, is useful to boost the performances of P-Net. We repeated the same experimental set-up without the initial use of the T-test method in order to achieve this goal. In this way P-Net employs as input all the available 9175 genes extracted from the \textit{pre-filtering steps} (described in subsection \ref{setup_ggc}) to compute the correlation matrix \textbf{\textit{W}}. The steps after the feature selection through T-test remain the same.\\
The first experiment is made using all the available features as input of P-Net with a different number of steps of the Random Walk Kernel ($p=1,2,3$) and different \textit{kernelized score functions}. We tested only the k-Nearest Neighbour score, the Nearest Neighbour score and the Average score because these are the functions that show the best results in the subsection \ref{exp_ggc_extensive}. The accuracies achieved from P-Net using all the available features and using only a subset of features, selected by T-test, are showed in the table \ref{table_amoutoffeat} and in the corresponding bar chart \ref{fig_barplot_pnetallfeat_rwk}. 

\begin{table}[H]
\centering
\resizebox{14cm}{3cm}{%
\begin{tabular}{ccccccc}
\cline{2-7}
\multicolumn{1}{l|}{} & \multicolumn{6}{c|}{\textit{\textbf{Accuracy}}} \\ \hline
\multicolumn{1}{|c|}{\textit{Score functions}} & \multicolumn{1}{c|}{\textit{\begin{tabular}[c]{@{}c@{}}RWK 1-step \\ (all features)\end{tabular}}} & \multicolumn{1}{c|}{\textit{\begin{tabular}[c]{@{}c@{}}RWK 1-step \\ (T-test features)\end{tabular}}} & \multicolumn{1}{c|}{\textit{\begin{tabular}[c]{@{}c@{}}RWK 2-step \\ (all features)\end{tabular}}} & \multicolumn{1}{c|}{\textit{\begin{tabular}[c]{@{}c@{}}RWK 2-step \\ (T-test features)\end{tabular}}} & \multicolumn{1}{c|}{\textit{\begin{tabular}[c]{@{}c@{}}RWK 3-step \\ (all features)\end{tabular}}} & \multicolumn{1}{c|}{\textit{\begin{tabular}[c]{@{}c@{}}RWK 3-step \\ (T-test features)\end{tabular}}} \\ \hline
\multicolumn{1}{|c|}{\textit{k-NN}} & \multicolumn{1}{c|}{57.83\%} & \multicolumn{1}{c|}{58.01\%} & \multicolumn{1}{c|}{60.86\%} & \multicolumn{1}{c|}{59.44\%} & \multicolumn{1}{c|}{61.60\%} & \multicolumn{1}{c|}{61.29\%} \\ \hline
\multicolumn{1}{|c|}{\textit{NN}} & \multicolumn{1}{c|}{55.19\%} & \multicolumn{1}{c|}{52.16\%} & \multicolumn{1}{c|}{57.80\%} & \multicolumn{1}{c|}{55.02\%} & \multicolumn{1}{c|}{59.69\%} & \multicolumn{1}{c|}{60.21\%} \\ \hline
\multicolumn{1}{|c|}{\textit{EAV}} & \multicolumn{1}{c|}{57.48\%} & \multicolumn{1}{c|}{58.01\%} & \multicolumn{1}{c|}{59.90\%} & \multicolumn{1}{c|}{59.44\%} & \multicolumn{1}{c|}{59.99\%} & \multicolumn{1}{c|}{61.29\%} \\ \hline
\multicolumn{1}{l}{} & \multicolumn{1}{l}{} & \multicolumn{1}{l}{} & \multicolumn{1}{l}{} & \multicolumn{1}{l}{} & \multicolumn{1}{l}{} & \multicolumn{1}{l}{} \\ \cline{2-7} 
\multicolumn{1}{l|}{} & \multicolumn{6}{c|}{\textit{\textbf{Standard error of the mean (SEM)}}} \\ \hline
\multicolumn{1}{|c|}{\textit{Score functions}} & \multicolumn{1}{c|}{\textit{\begin{tabular}[c]{@{}c@{}}RWK 1-step\\ (all features)\end{tabular}}} & \multicolumn{1}{c|}{\textit{\begin{tabular}[c]{@{}c@{}}RWK 1-step\\ (T-test features)\end{tabular}}} & \multicolumn{1}{c|}{\textit{\begin{tabular}[c]{@{}c@{}}RWK 2-step \\ (all features)\end{tabular}}} & \multicolumn{1}{c|}{\textit{\begin{tabular}[c]{@{}c@{}}RWK 2-step \\ (T-test features)\end{tabular}}} & \multicolumn{1}{c|}{\textit{\begin{tabular}[c]{@{}c@{}}RWK 3-step \\ (all features)\end{tabular}}} & \multicolumn{1}{c|}{\textit{\begin{tabular}[c]{@{}c@{}}RWK 3-step \\ (T-test features)\end{tabular}}} \\ \hline
\multicolumn{1}{|c|}{\textit{k-NN}} & \multicolumn{1}{c|}{2.01\%} & \multicolumn{1}{c|}{2.20\%} & \multicolumn{1}{c|}{1.97\%} & \multicolumn{1}{c|}{2.12\%} & \multicolumn{1}{c|}{1.83\%} & \multicolumn{1}{c|}{2.09\%} \\ \hline
\multicolumn{1}{|c|}{\textit{NN}} & \multicolumn{1}{c|}{2.27\%} & \multicolumn{1}{c|}{1.97\%} & \multicolumn{1}{c|}{2.26\%} & \multicolumn{1}{c|}{2.00\%} & \multicolumn{1}{c|}{1.96\%} & \multicolumn{1}{c|}{2.00\%} \\ \hline
\multicolumn{1}{|c|}{\textit{EAV}} & \multicolumn{1}{c|}{1.95\%} & \multicolumn{1}{c|}{2.20\%} & \multicolumn{1}{c|}{1.94\%} & \multicolumn{1}{c|}{2.12\%} & \multicolumn{1}{c|}{1.96\%} & \multicolumn{1}{c|}{2.09\%} \\ \hline
\end{tabular}%
}
\caption[Tables to show the performances of P-Net with different amount of features]{\textbf{Tables to show the performances of P-Net with different amount of features.} In the top table there are the accuracies and in the bottom table the corresponding SEM.}
\label{table_amoutoffeat}
\end{table}

As we can see, from the table and bar chart above, the accuracies achieved by P-Net are really similar in the two considered conditions. A possible explanation is that the univariate filter with T-test is not able to select the most important features to prioritise patients with respect to the outcome \textit{C} under study. However, another possible interpretation is that we need to select a number of features lower than 1000 to improve the performances of the algorithm.

\begin{figure}[H]
\centering
\includegraphics[scale=0.6]{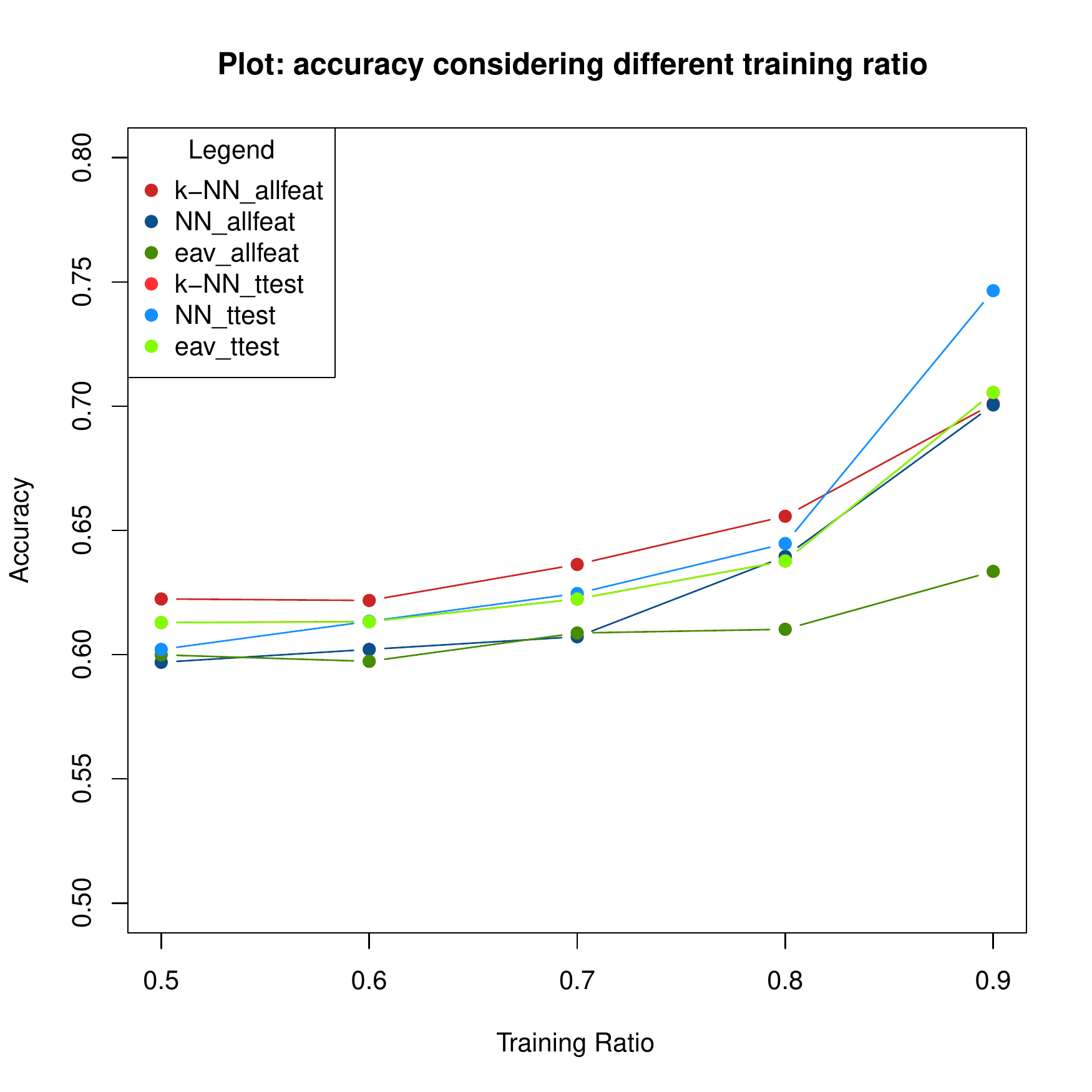}
\caption[P-Net perfomances for different amount of features and training set sizes]{\textbf{Plot to compare the performances of P-Net using different dimensions of the training set and different amount of features.} In the legend ``ttest'' specify the use of the feature selection step instead ``allfeat'' means that we used all the available features.} \label{fig_plot_pnetall_trainratio}
\end{figure}

In the precedent experiment, all the tests are made with a training set size of 16 samples (training ratio = 0.5). To understand if the effect of the feature selection has a greater impact when the training set size is different, we make other experiments considering only the 3-step Random Walk Kernel under different sizes of the training set (the accuracies are showed in table \ref{table_amountoffeat_tr} and plot \ref{fig_plot_pnetall_trainratio}). 

\begin{table}[H]
\centering
\resizebox{\textwidth}{3cm}{%
\begin{tabular}{lllllllllllllll}
 &  &  &  &  &  &  &  &  &  &  &  &  &  &  \\ \cline{2-6} \cline{9-13}
\multicolumn{1}{l|}{} & \multicolumn{5}{c|}{\textit{\textbf{Accuracy (all features)}}} &  & \multicolumn{1}{l|}{} & \multicolumn{5}{c|}{\textit{\textbf{Accuracy (T-test features)}}} &  &  \\ \cline{1-6} \cline{8-13}
\multicolumn{1}{|c|}{\textit{Score functions}} & \multicolumn{1}{c|}{\textit{tr05}} & \multicolumn{1}{c|}{\textit{tr06}} & \multicolumn{1}{c|}{\textit{tr07}} & \multicolumn{1}{c|}{\textit{tr08}} & \multicolumn{1}{c|}{\textit{tr09}} & \multicolumn{1}{c|}{\textit{}} & \multicolumn{1}{c|}{\textit{Score functions}} & \multicolumn{1}{c|}{\textit{tr05}} & \multicolumn{1}{c|}{\textit{tr06}} & \multicolumn{1}{c|}{\textit{tr07}} & \multicolumn{1}{c|}{\textit{tr08}} & \multicolumn{1}{c|}{\textit{tr09}} &  &  \\ \cline{1-6} \cline{8-13}
\multicolumn{1}{|c|}{\textit{k-NN}} & \multicolumn{1}{c|}{62.24\%} & \multicolumn{1}{c|}{62.18\%} & \multicolumn{1}{c|}{63.63\%} & \multicolumn{1}{c|}{65.57\%} & \multicolumn{1}{c|}{70.10\%} & \multicolumn{1}{c|}{} & \multicolumn{1}{c|}{\textit{k-NN}} & \multicolumn{1}{c|}{61.29\%} & \multicolumn{1}{c|}{61.33\%} & \multicolumn{1}{c|}{62.24\%} & \multicolumn{1}{c|}{63.77\%} & \multicolumn{1}{c|}{70.55\%} &  &  \\ \cline{1-6} \cline{8-13}
\multicolumn{1}{|c|}{\textit{NN}} & \multicolumn{1}{c|}{59.69\%} & \multicolumn{1}{c|}{60.21\%} & \multicolumn{1}{c|}{60.72\%} & \multicolumn{1}{c|}{63.95\%} & \multicolumn{1}{c|}{70.05\%} & \multicolumn{1}{c|}{} & \multicolumn{1}{c|}{\textit{NN}} & \multicolumn{1}{c|}{60.21\%} & \multicolumn{1}{c|}{61.34\%} & \multicolumn{1}{c|}{62.46\%} & \multicolumn{1}{c|}{64.47\%} & \multicolumn{1}{c|}{74.65\%} &  &  \\ \cline{1-6} \cline{8-13}
\multicolumn{1}{|c|}{\textit{EAV}} & \multicolumn{1}{c|}{59.99\%} & \multicolumn{1}{c|}{59.73\%} & \multicolumn{1}{c|}{60.87\%} & \multicolumn{1}{c|}{61.02\%} & \multicolumn{1}{c|}{63.35\%} & \multicolumn{1}{c|}{} & \multicolumn{1}{c|}{\textit{EAV}} & \multicolumn{1}{c|}{61.29\%} & \multicolumn{1}{c|}{61.32\%} & \multicolumn{1}{c|}{62.24\%} & \multicolumn{1}{c|}{63.77\%} & \multicolumn{1}{c|}{70.55\%} &  &  \\ \cline{1-6} \cline{8-13}
 &  &  &  &  &  &  &  &  &  &  &  &  &  &  \\ \cline{2-6} \cline{9-13}
\multicolumn{1}{c|}{} & \multicolumn{5}{c|}{\textit{\textbf{Standard error of the mean (all features)}}} &  & \multicolumn{1}{c|}{} & \multicolumn{5}{c|}{\textit{\textbf{Standard error of the mean (T-test features)}}} &  &  \\ \cline{1-6} \cline{8-13}
\multicolumn{1}{|c|}{\textit{Score functions}} & \multicolumn{1}{c|}{\textit{tr05}} & \multicolumn{1}{c|}{\textit{tr06}} & \multicolumn{1}{c|}{\textit{tr07}} & \multicolumn{1}{c|}{\textit{tr08}} & \multicolumn{1}{c|}{\textit{tr09}} & \multicolumn{1}{c|}{} & \multicolumn{1}{c|}{\textit{Score functions}} & \multicolumn{1}{c|}{\textit{tr05}} & \multicolumn{1}{c|}{\textit{tr06}} & \multicolumn{1}{c|}{\textit{tr07}} & \multicolumn{1}{c|}{\textit{tr08}} & \multicolumn{1}{c|}{\textit{tr09}} &  &  \\ \cline{1-6} \cline{8-13}
\multicolumn{1}{|c|}{\textit{k-NN}} & \multicolumn{1}{c|}{1.97\%} & \multicolumn{1}{c|}{2.27\%} & \multicolumn{1}{c|}{2.65\%} & \multicolumn{1}{c|}{3.46\%} & \multicolumn{1}{c|}{8.18\%} & \multicolumn{1}{c|}{} & \multicolumn{1}{c|}{\textit{k-NN}} & \multicolumn{1}{c|}{2.09\%} & \multicolumn{1}{c|}{2.27\%} & \multicolumn{1}{c|}{2.52\%} & \multicolumn{1}{c|}{3.51\%} & \multicolumn{1}{c|}{7.88\%} &  &  \\ \cline{1-6} \cline{8-13}
\multicolumn{1}{|c|}{\textit{NN}} & \multicolumn{1}{c|}{1.96\%} & \multicolumn{1}{c|}{2.07\%} & \multicolumn{1}{c|}{2.50\%} & \multicolumn{1}{c|}{3.42\%} & \multicolumn{1}{c|}{8.25\%} & \multicolumn{1}{c|}{} & \multicolumn{1}{c|}{\textit{NN}} & \multicolumn{1}{c|}{2.00\%} & \multicolumn{1}{c|}{2.19\%} & \multicolumn{1}{c|}{2.44\%} & \multicolumn{1}{c|}{3.40\%} & \multicolumn{1}{c|}{7.79\%} &  &  \\ \cline{1-6} \cline{8-13}
\multicolumn{1}{|c|}{\textit{EAV}} & \multicolumn{1}{c|}{1.96\%} & \multicolumn{1}{c|}{2.06\%} & \multicolumn{1}{c|}{2.40\%} & \multicolumn{1}{c|}{3.39\%} & \multicolumn{1}{c|}{8.36\%} & \multicolumn{1}{c|}{} & \multicolumn{1}{c|}{\textit{EAV}} & \multicolumn{1}{c|}{2.09\%} & \multicolumn{1}{c|}{2.27\%} & \multicolumn{1}{c|}{2.52\%} & \multicolumn{1}{c|}{3.51\%} & \multicolumn{1}{c|}{7.88\%} &  &  \\ \cline{1-6} \cline{8-13}
 &  &  &  &  &  &  &  &  &  &  &  &  &  &   
\end{tabular}%
}
\caption[Accuracies achieved from P-Net using different training set sizes]{\textbf{Accuracies achieved from P-Net using different training set sizes.}}
\label{table_amountoffeat_tr}
\end{table}

From the table and the plot above, we can see the impact of the feature selection step when the training set size increases. Even if the results are similar there is a slight improvement with the use of the T-test, at least when we consider the Nearest Neighbour score function and the Average score function. The improvement is surprising when the training ratio is 0.9. Indeed, in this case, we have an improvement of $4.06 \%$ and $7.20 \%$ (with NN score and EAV score, respectively) and in both the situations there is a decrease in the standard error of the mean.
It seems that the feature selection step gives an advantage to P-Net especially when we have a big training set size. 

\subsection{P-Net applied to a reduced number of features selected by T-test}

Until this moment, we used the T-test to select the most important 1000 features with the smallest p-value. The decision to pick the first 1000 features is absolutely arbitrary and another number of features could be selected. So, we designed two tests where we select a different reduced number of features. More precisely, we decided to use the 8-step Random Walk Kernel and to test all the score functions and all the training set sizes in both the tests. But, in the first test we select only 200 features by T-test and in the second one we reduce the number of features to 100.\\
The accuracies from these two tests are showed in the tables \ref{table_200feat} and \ref{table_100feat}.

\begin{table}[H]
\centering
\resizebox{\textwidth}{1.7cm}{%
\begin{tabular}{c|c|c|c|c|c|lc|c|c|c|c|c|l}
\cline{2-6} \cline{9-13}
\multicolumn{1}{l|}{} & \multicolumn{5}{c|}{\textit{\textbf{Accuracy (200 features selected)}}} &  & \multicolumn{1}{l|}{} & \multicolumn{5}{c|}{\textit{\textbf{SEM (200 features selected)}}} &  \\ \cline{1-6} \cline{8-13}
\multicolumn{1}{|l|}{\textit{Score functions}} & \textit{tr05} & \textit{tr06} & \textit{tr07} & \textit{tr08} & \textit{tr09} & \multicolumn{1}{l|}{} & \textit{Score functions} & \textit{tr05} & \textit{tr06} & \textit{tr07} & \textit{tr08} & \textit{tr09} &  \\ \cline{1-6} \cline{8-13}
\multicolumn{1}{|c|}{\textit{k-NN}} & 62.50\% & 63.38\% & 64.02\% & 65.65\% & 71.50\% & \multicolumn{1}{l|}{} & \textit{k-NN} & 2.25\% & 2.42\% & 2.64\% & 3.55\% & 7.60\% &  \\ \cline{1-6} \cline{8-13}
\multicolumn{1}{|c|}{\textit{NN}} & 62.72\% & 64.12\% & 65.80\% & 68.87\% & 77.75\% & \multicolumn{1}{l|}{} & \textit{NN} & 2.13\% & 2.35\% & 2.61\% & 3.60\% & 7.46\% &  \\ \cline{1-6} \cline{8-13}
\multicolumn{1}{|c|}{\textit{EAV}} & 62.50\% & 63.38\% & 64.02\% & 65.65\% & 71.50\% & \multicolumn{1}{l|}{} & \textit{EAV} & 2.25\% & 2.42\% & 2.64\% & 3.55\% & 7.60\% &  \\ \cline{1-6} \cline{8-13}
\multicolumn{1}{|c|}{\textit{TOT}} & 57.11\% & 57.68\% & 57.30\% & 57.32\% & 64.90\% & \multicolumn{1}{l|}{} & \textit{TOT} & 2.34\% & 2.55\% & 2.92\% & 3.94\% & 6.01\% &  \\ \cline{1-6} \cline{8-13}
\multicolumn{1}{|c|}{\textit{DIFF}} & 42.76\% & 48.92\% & 54.57\% & 60.70\% & 72.10\% & \multicolumn{1}{l|}{} & \textit{DIFF} & 2.45\% & 2.72\% & 3.04\% & 3.86\% & 7.94\% &  \\ \cline{1-6} \cline{8-13}
\multicolumn{1}{|c|}{\textit{DNORM}} & 51.93\% & 51.80\% & 53.50\% & 54.88\% & 64.30\% & \multicolumn{1}{l|}{} & \textit{DNORM} & 2.50\% & 2.76\% & 3.02\% & 3.99\% & 6.16\% &  \\ \cline{1-6} \cline{8-13}
\end{tabular}%
}
\caption[Accuracies achieved from P-Net with 200 features selected by T-test]{\textbf{Accuracies achieved from P-Net with 200 features selected by T-test.}}
\label{table_200feat}
\end{table}

\begin{table}[H]
\centering
\resizebox{\textwidth}{1.7cm}{%
\begin{tabular}{c|c|c|c|c|c|lc|c|c|c|c|c|l}
\cline{2-6} \cline{9-13}
\multicolumn{1}{l|}{} & \multicolumn{5}{c|}{\textit{\textbf{Accuracy (100 features selected)}}} &  & \multicolumn{1}{l|}{} & \multicolumn{5}{c|}{\textit{\textbf{SEM (100 features selected)}}} &  \\ \cline{1-6} \cline{8-13}
\multicolumn{1}{|l|}{\textit{Score functions}} & \textit{tr05} & \textit{tr06} & \textit{tr07} & \textit{tr08} & \textit{tr09} & \multicolumn{1}{l|}{} & \textit{Score functions} & \textit{tr05} & \textit{tr06} & \textit{tr07} & \textit{tr08} & \textit{tr09} &  \\ \cline{1-6} \cline{8-13}
\multicolumn{1}{|c|}{\textit{k-NN}} & 62.09\% & 62.88\% & 64.04\% & 66.60\% & 75.90\% & \multicolumn{1}{l|}{} & \textit{k-NN} & 2.29\% & 2.42\% & 2.70\% & 3.64\% & 7.61\% &  \\ \cline{1-6} \cline{8-13}
\multicolumn{1}{|c|}{\textit{NN}} & 62.04\% & 63.64\% & 65.22\% & 68.07\% & 78.45\% & \multicolumn{1}{l|}{} & \textit{NN} & 2.25\% & 2.41\% & 2.68\% & 3.59\% & 7.22\% &  \\ \cline{1-6} \cline{8-13}
\multicolumn{1}{|c|}{\textit{EAV}} & 62.09\% & 62.89\% & 64.04\% & 66.60\% & 75.90\% & \multicolumn{1}{l|}{} & \textit{EAV} & 2.29\% & 2.42\% & 2.70\% & 3.64\% & 7.61\% &  \\ \cline{1-6} \cline{8-13}
\multicolumn{1}{|c|}{\textit{TOT}} & 57.90\% & 58.08\% & 57.69\% & 56.52\% & 66.75\% & \multicolumn{1}{l|}{} & \textit{TOT} & 2.38\% & 2.60\% & 2.97\% & 3.97\% & 5.52\% &  \\ \cline{1-6} \cline{8-13}
\multicolumn{1}{|c|}{\textit{DIFF}} & 43.26\% & 48.71\% & 54.64\% & 60.03\% & 74.05\% & \multicolumn{1}{l|}{} & \textit{DIFF} & 2.59\% & 2.67\% & 3.12\% & 4.07\% & 7.88\% &  \\ \cline{1-6} \cline{8-13}
\multicolumn{1}{|c|}{\textit{DNORM}} & 52.31\% & 51.61\% & 53.94\% & 54.83\% & 66.25\% & \multicolumn{1}{l|}{} & \textit{DNORM} & 2.58\% & 2.76\% & 3.03\% & 4.06\% & 5.58\% &  \\ \cline{1-6} \cline{8-13}
\end{tabular}%
}
\caption[Accuracies achieved from P-Net with 100 features selected by T-test]{\textbf{Accuracies achieved from P-Net with 100 features selected by T-test.}}
\label{table_100feat}
\end{table}

\begin{table}[H]
\centering
\resizebox{\textwidth}{1.7cm}{%
\begin{tabular}{c|c|c|c|c|c|lc|c|c|c|c|c|l}
\cline{2-6} \cline{9-13}
\multicolumn{1}{l|}{} & \multicolumn{5}{c|}{\textit{\textbf{Accuracy (1000 features selected)}}} &  & \multicolumn{1}{l|}{} & \multicolumn{5}{c|}{\textit{\textbf{SEM (1000 features selected)}}} &  \\ \cline{1-6} \cline{8-13}
\multicolumn{1}{|l|}{\textit{Score functions}} & \textit{tr05} & \textit{tr06} & \textit{tr07} & \textit{tr08} & \textit{tr09} & \multicolumn{1}{l|}{} & \textit{Score functions} & \textit{tr05} & \textit{tr06} & \textit{tr07} & \textit{tr08} & \textit{tr09} &  \\ \cline{1-6} \cline{8-13}
\multicolumn{1}{|c|}{\textit{k-NN}} & 62.41\% & 62.48\% & 63.55\% & 64.37\% & 70.35\% & \multicolumn{1}{l|}{} & \textit{k-NN} & 2.08\% & 2.26\% & 2.43\% & 3.23\% & 7.51\% &  \\ \cline{1-6} \cline{8-13}
\multicolumn{1}{|c|}{\textit{NN}} & 64.23\% & 64.85\% & 65.99\% & 68.03\% & 78.05\% & \multicolumn{1}{l|}{} & \textit{NN} & 2.02\% & 2.23\% & 2.38\% & 3.38\% & 7.38\% &  \\ \cline{1-6} \cline{8-13}
\multicolumn{1}{|c|}{\textit{EAV}} & 62.41\% & 62.48\% & 63.55\% & 64.37\% & 70.35\% & \multicolumn{1}{l|}{} & \textit{EAV} & 2.08\% & 2.26\% & 2.43\% & 3.23\% & 7.51\% &  \\ \cline{1-6} \cline{8-13}
\multicolumn{1}{|c|}{\textit{TOT}} & 57.21\% & 57.28\% & 58.52\% & 58.57\% & 61.00\% & \multicolumn{1}{l|}{} & \textit{TOT} & 2.15\% & 2.44\% & 2.54\% & 3.58\% & 6.21\% &  \\ \cline{1-6} \cline{8-13}
\multicolumn{1}{|c|}{\textit{DIFF}} & 44.09\% & 48.47\% & 53.72\% & 60.50\% & 72.95\% & \multicolumn{1}{l|}{} & \textit{DIFF} & 2.19\% & 2.56\% & 3.01\% & 3.69\% & 7.50\% &  \\ \cline{1-6} \cline{8-13}
\multicolumn{1}{|c|}{\textit{DNORM}} & 51.39\% & 50.27\% & 53.40\% & 56.22\% & 62.50\% & \multicolumn{1}{l|}{} & \textit{DNORM} & 2.36\% & 2.64\% & 2.89\% & 3.69\% & 6.16\% &  \\ \cline{1-6} \cline{8-13}
\end{tabular}%
}
\caption[Accuracies achieved from P-Net with 1000 features selected by T-test]{\textbf{Accuracies achieved from P-Net with 1000 features selected by T-test.}}
\label{table_1000feat}
\end{table}

\begin{figure}[H]
\centering
\includegraphics[scale=0.7]{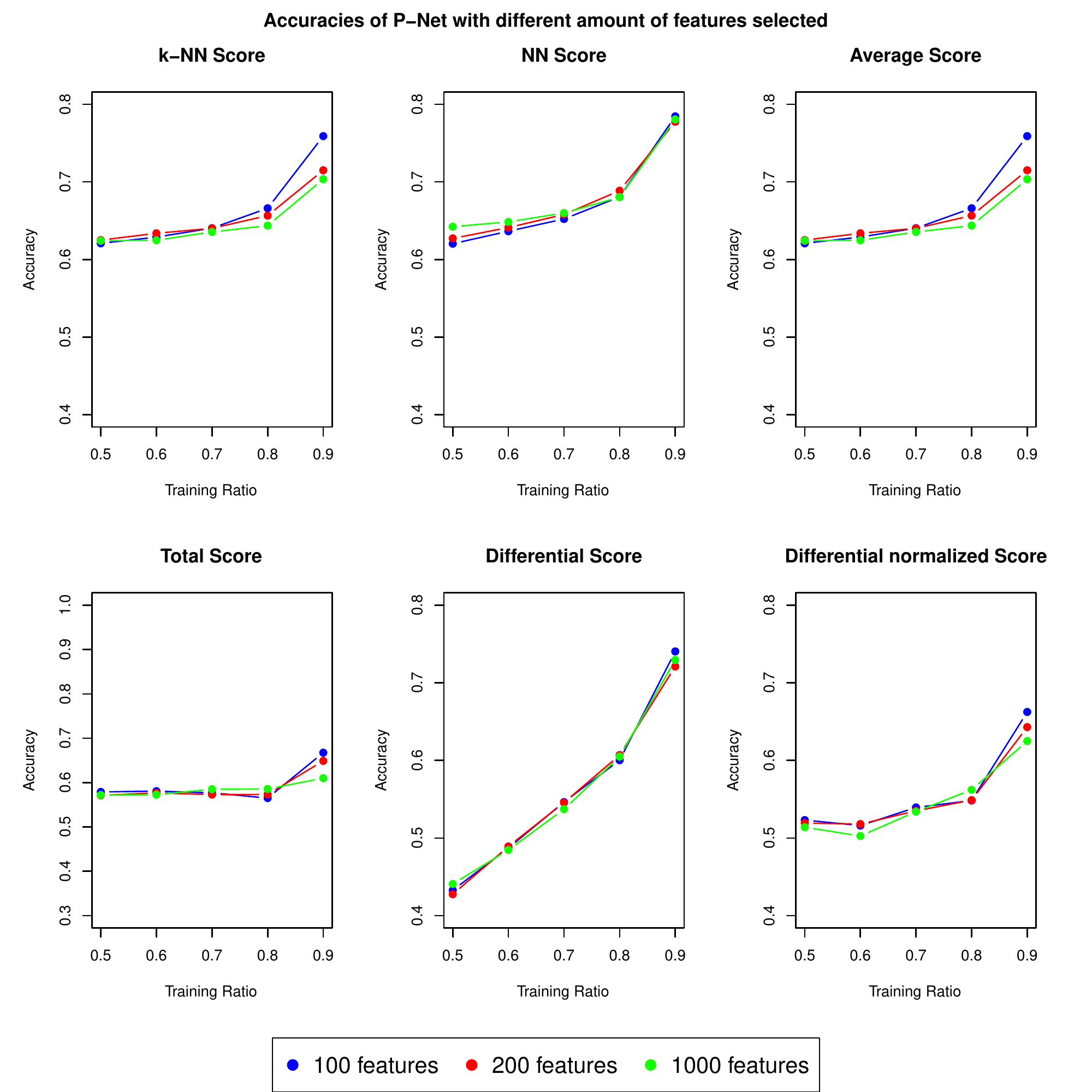}
\caption[Accuracies of P-Net with different amount of features selected]{\textbf{Plot to compare the performances of P-Net using different amounts of selected features by T-test.} Each plot concern a different kernelized score function.} \label{fig_plot_reducedfeat}
\end{figure}

The same results showed in the above tables are plotted in the figure \ref{fig_plot_reducedfeat} to have a better general overview of the results. As we can see from the plot, the reduction of the number of selected features does not improve in general the accuracy. The accuracies are almost the same for the k-NN, Average and Total score when we consider small training ratio.  When the training ratio is 0.8-0.9, there is an improvement in the accuracy related to the reduction of the selected features. We obtain an improvement of $5.55\%$ for k-NN and Average scores and of $5.75\%$ for the Total score. The corresponding Standard error of the Mean (SEM) has a slight increase with the features reduction ($0.1 \%$ for the k-NN and Average score and $0.69\%$ for the Total score). When we test the Differential Score function, the results are always almost the same and for the Differential Normalized score, it is not clear the impact of the number of features on the P-Net performances. The score function that always achieves the best results is the Nearest Neighbour function. For the training ratio 0.5, 0.6 and 0.7, in this case, we can see a slight reduction of the accuracy with a smaller number of features and an higher SEM. For the training ratio 0.8 and 0.9 there isn't a specific trend in the data.\\
Finally, we can conclude that the reduction of the selected features does not generally improve the performances of P-Net, even if we increase the number of samples in the training set. Probably there are some important features to prioritize the samples also after the $ 200^{th}$ ranked feature and if we not consider these features the algorithm has less information to compute the correlation matrix \textbf{\textit{W}} and to assign the right weights to the edges of the graph \textit{G}. The reduction of the accuracy reflects the previously mentioned situation. 

\subsection{Experiment with the use of Support Vector Machine as classifier}

\begin{figure}[H]
\centering
\includegraphics[scale=0.5]{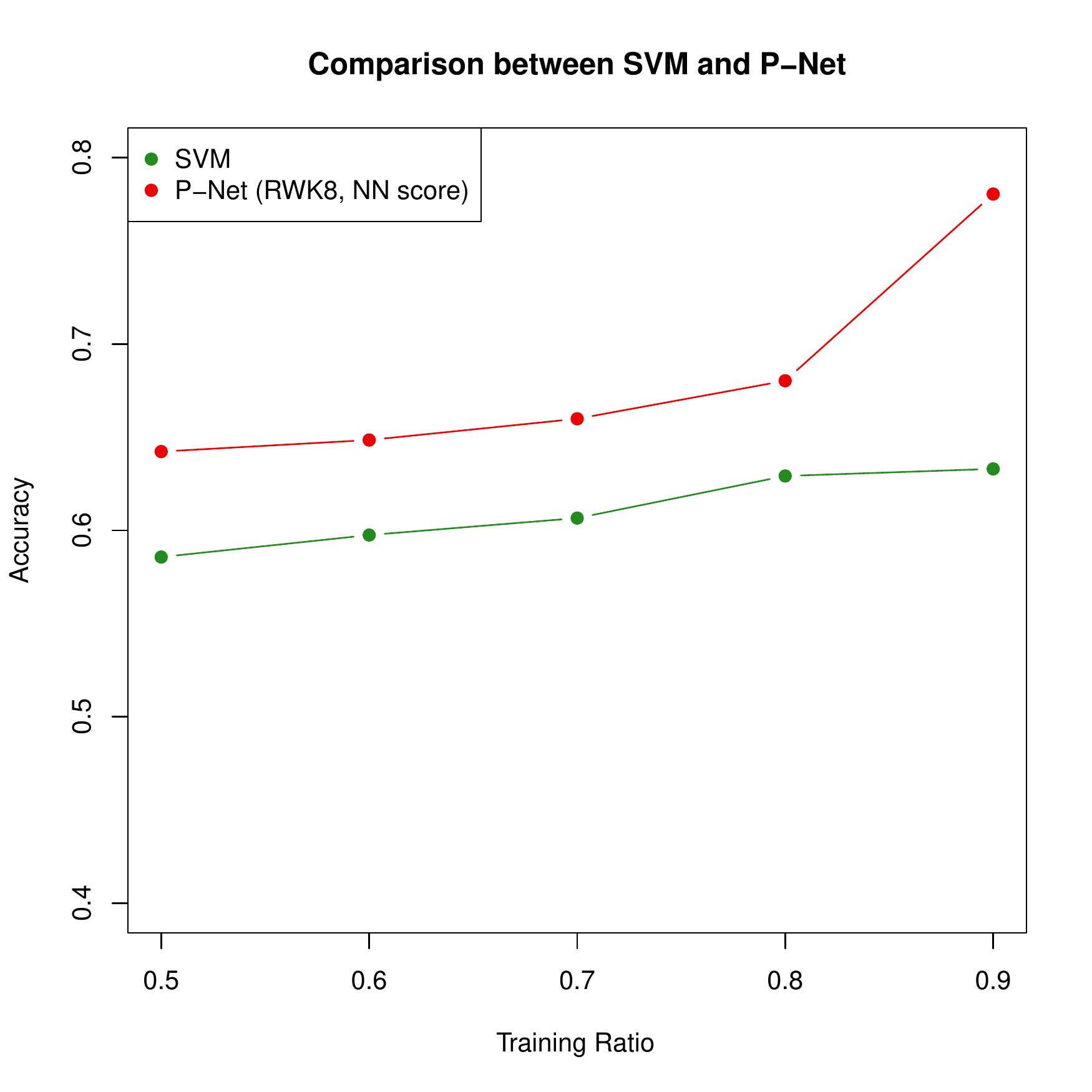}
\caption [Plot to compare the performances of SVM and P-Net after the selection of the first 1000 features by T-test]{\textbf{Plot to compare the performances of SVM and P-Net after the selection of the first 1000 features by T-test.} In the P-Net algorithm is used the 8-steps Random Walk Kernel and the Nearest Neighbour score function.} 
\label{fig_comparison_plot_svm}
\end{figure}

In the reference paper \cite{winter2012google}, one of the feature selection methods tested is the T-test, which is the same method we use in our workflow. However, in the paper, they use the T-test to select the first 5-10 features and then they use only these features to train an SVM classifier. The choice to use only 5-10 features is related to the purpose to obtain a signature with a size suitable to be applicable in a clinical setting using immunohistochemistry staining of the signature proteins. Our goal is just to predict the class of each sample with a good accuracy, so our signature is bigger (1000 genes).\\
In this subsection, we run another experiment where we use the usual workflow (selecting 1000 features by T-test) but we replace P-Net with a Support Vector Machine classifier. In this way, we can compare the results achieved by P-Net with respect to the results achieved by the SVM. The results are showed in the table \ref{table_svm} and the comparison with P-Net is showed in the plot \ref{fig_comparison_plot_svm}.

\begin{table}[H]
\centering
\resizebox{\textwidth}{1.2cm}{%
\begin{tabular}{c|c|c|c|c|c|llc|c|c|c|c|c|ll}
\cline{2-6} \cline{10-14}
\multicolumn{1}{l|}{} & \multicolumn{5}{c|}{\textit{\textbf{Accuracy}}} &  &  & \multicolumn{1}{l|}{} & \multicolumn{5}{c|}{\textit{\textbf{Standard error of the mean}}} &  &  \\ \cline{1-6} \cline{9-14}
\multicolumn{1}{|c|}{\textit{Method}} & \textit{tr05} & \textit{tr06} & \textit{tr07} & \textit{tr08} & \textit{tr09} &  & \multicolumn{1}{l|}{} & \textit{Method} & \textit{tr05} & \textit{tr06} & \textit{tr07} & \textit{tr08} & \textit{tr09} &  &  \\ \cline{1-6} \cline{9-14}
\multicolumn{1}{|c|}{\textit{SVM}} & 58.57\% & 59.75\% & 60.66\% & 62.92\% & 63.30\% &  & \multicolumn{1}{l|}{} & \textit{SVM} & \textit{2.11\%} & \textit{2.18\%} & \textit{2.54\%} & \textit{3.39\%} & \textit{5.98\%} &  &  \\ \cline{1-6} \cline{9-14}
\multicolumn{1}{|c|}{\textit{\begin{tabular}[c]{@{}c@{}}P-Net  \\ (RWK-8, NN score)\end{tabular}}} & 64.23\% & 64.85\% & 65.99\% & 68.03\% & 78.05\% &  & \multicolumn{1}{l|}{} & \textit{\begin{tabular}[c]{@{}c@{}}P-Net \\ (RWK-8, NN score)\end{tabular}} & \textit{2.02\%} & \textit{2.23\%} & \textit{2.38\%} & \textit{3.38\%} & \textit{7.38\%} &  &  \\ \cline{1-6} \cline{9-14}
\end{tabular}%
}
\caption[Comparison of the accuracies of P-Net and SVM]{\textbf{Accuracies of SVM and P-Net with the use of T-Test to select the first 1000 features.}}
\label{table_svm}
\end{table}

P-Net achieves better results with respect to the linear Support Vector Machine, especially when we consider the training ratio 0.9. Indeed, for this size of the training set, we have an increase of the accuracy of $14.75\%$ and an increase of the standard error of the mean of $1.4 \%$.

\section{Analysis on Melanoma and Ovarian cancer datasets}

In this section, we follow the experimental set-up of the reference paper \cite{barter2014network}. Two datasets are used in the paper:
\begin{itemize}
\item \textbf{Melanoma dataset}: contains expression data from 79 patients with metastatic (stage III) melanoma.
\item \textbf{Ovarian cancer dataset}: contains expression data from 195 patients. 185 patients are afflicted with stage III high-grade primary papillary serous tumours  of the ovary and 10 samples are collected from normal ovarian surface epithelium. 
\end{itemize}

Both the datasets can be downloaded from Gene Expression Omnibus - GEO \cite{barrett200619, edgar2002gene}, which is a public functional genomics data repository, with the accession numbers 53118 and 26712, respectively. 

\subsubsection{Melanoma dataset}

From the Gene Expression Omnibus web page of this dataset, we know that the platform used to measure the gene expression levels is Illumina HumanWG-6 v3.0 expression beadchip. The array annotation was performed on R-2.11.0 with the annotation package \textit{illuminaHumanv3.db}; quality control was performed on all chips using R/Bioconductor and the \textit{lumi} package. Data normalization was performed using a variance-stabilizing transform (VST) and quantile normalization as implemented in the \textit{lumi} package.\\
In this dataset, the initial gene expression matrix \textbf{\textit{M}} contains 79 patients and 26085 probe sets. The corresponding 39 phenotypic variables for each patient are stored in the matrix \textit{\textbf{pdata}}. Then, we select from the matrix \textbf{\textit{M}} only the patients used in the reference paper exploiting the tumourIDs, which are available in the electronic supplementary material of the paper. However, in the reference paper they select 47 patients but we select only 44 patients because in the Melanoma dataset the patients with the IDs 205, 249 and 385 are missing.\\
Another important step is the \textbf{creation of the vector of labels} that contains the corresponding label for each patient. Following the paper, we define the two classes of patients: \textit{poor prognosis} and \textit{good prognosis}. These groups are defined as having time from surgery to death from melanoma greater than 4 years with no sign of relapse or less than one year. So, the dataset contains 23 patients with good prognosis and 21 patients with poor prognosis. 

\subsubsection{Ovarian cancer dataset}

The platform used to measure the gene expression levels is Affymetrix Human Genome U133A Array and Robust Multi-Array Analysis (RMA) was performed.\\
In this dataset the initial expression matrix \textbf{\textit{M}} contains 195 patients and 22283 probe sets. The matrix \textbf{\textit{pdata}} contains 37 phenotypic variables for each patient. Also in this dataset, we select from the matrix \textbf{\textit{M}} only the patients used in the reference paper exploiting the tumourIDs from the electronic supplementary material. We extracted 72 patients according with the reference paper.\\ 
The next step is the \textbf{creation of the vector of labels}. Following the paper, we defined the \textit{poor prognosis} class as patients who died within 2 years after surgery and the \textit{good prognosis} class as patients alive more than 3 years after surgery. We obtained 33 patients with good prognosis and 39 patients with poor prognosis. 

At this point the Melanoma dataset contains 44 samples and 26085 probe sets, instead the Ovarian cancer dataset contains 72 samples and 22283 probe sets.

\subsection{Experimental Set-up for the Melanoma and Ovarian cancer datasets} \label{setup_mel_ov}

In the experimental workflow the first step is the filtering of the expression matrix \textbf{\textit{M}} exploiting the protein-protein interaction (PPI) network iRefWeb \cite{turner2010irefweb} (iRefIndex version=8.0) and the package \textit{iRefR} \cite{morairefr, mora2011irefr}. We decided to select from the PPI network the interactions with the following characteristics:

\begin{itemize}
\item only human-human interactions
\item direct interactions and physical interactions
\end{itemize} 

The choice of these features is based on the reference paper but the explanations provided by the authors are not complete and this is the reason why we obtained a slightly different number of nodes and edges in the network (7541 nodes and 21977 edges, instead of 7256 nodes and 21049 edges as in the paper).\\
Once we obtain the PPI network with the desired characteristics, exploiting the \textit{iRefR} package, we use the genes contained in the network to filter the genes contained in the matrices \textit{\textbf{M}} of the two datasets. To achieve this goal, the probeIDs (the univocal identifiers assigned to each probe in a microarray) are extracted from the initial matrix \textbf{\textit{M}}. Then we use a suitable annotation data package to retrieve the mapping between the probeIDs and the corresponding Entrez IDs. We exploit the function \textit{create\_id\_conversion\_table} from the \textit{iRefR} package to obtain the Entrez IDs from the genes of the PPI network. Finally, from the Entrez IDs of the matrix \textbf{\textit{M}} we select only the IDs in common with the network and we can use the common IDs to filter the matrix \textbf{\textit{M}}. 
The annotation step is necessary to filter both the datasets with the genes of the PPI network. In the Melanoma dataset we used the annotation data contained in the Bioconductor package \textit{illuminaHumanv3.db} and in the Ovarian cancer dataset we used the Bioconductor package \textit{hgu133a.db}. The annotation of the probe sets is required to retrieve the corresponding genes and we did it on R-3.1.1. Finally, we obtained two filtered matrices \textbf{\textit{fM}}:
\begin{itemize}
\item The Melanoma filtered matrix contains 9433 genes and 44 samples
\item The Ovarian cancer filtered matrix contains 10771 genes and 72 samples
\end{itemize}

\subsubsection{Core of the experimental set-up}

\begin{figure}[tbp]
\centering
\includegraphics[scale=0.8]{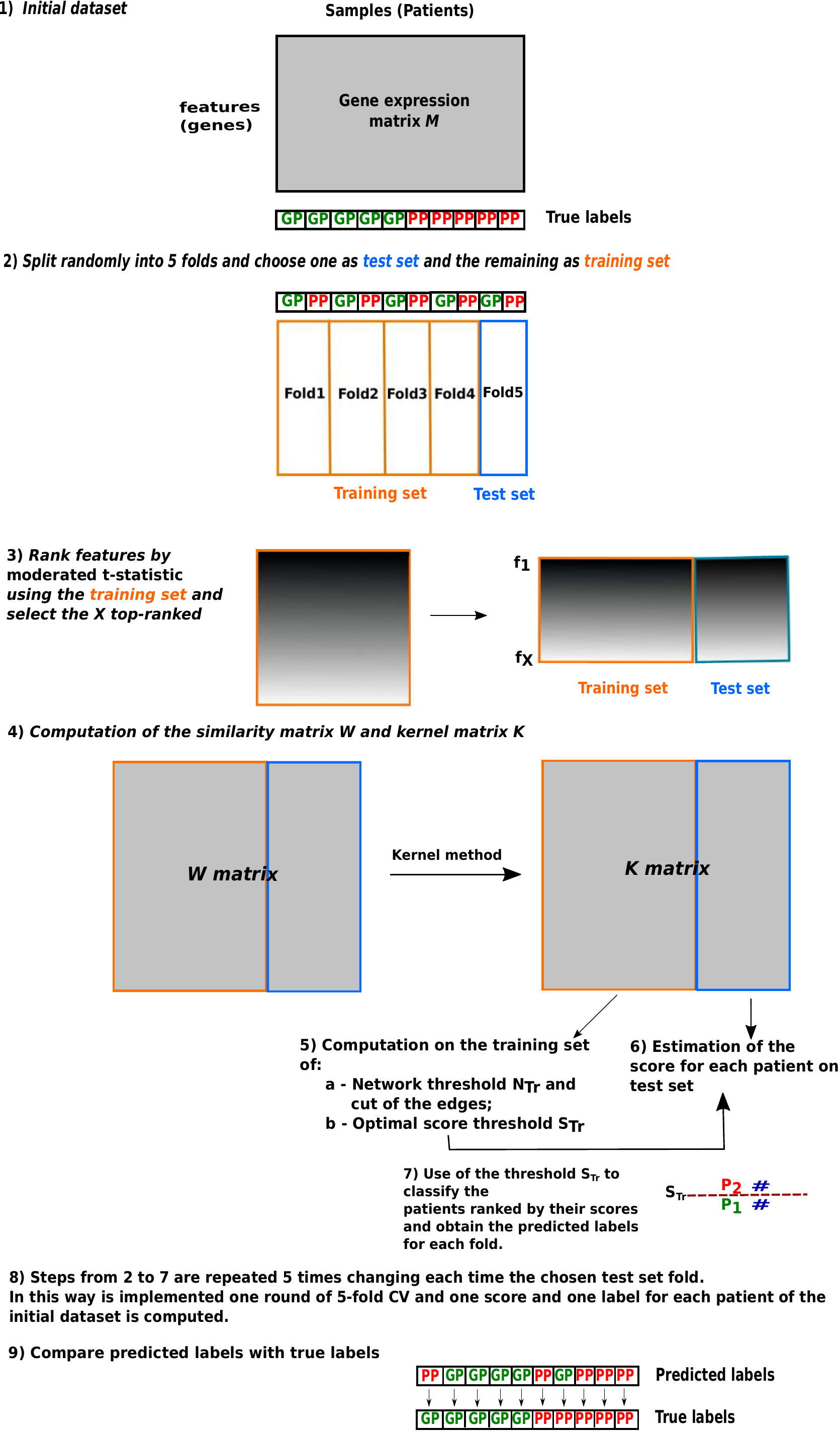}
\caption[Workflow for the Melanoma and Ovarian cancer datasets using P-Net as classifier]{\textbf{Workflow for the Melanoma and Ovarian cancer datasets using P-Net as classifier.}} 
\label{fig_setup_melov}
\end{figure}

The core of the experimental set-up applied in the reference paper consists in the application of a feature selection method and 100 rounds of \textit{5-fold cross validation} to assess the performances of some classifiers.
In our case we apply the same overall procedure: our feature selection method is \textbf{the moderated t-statistic} and our classifier is \textbf{P-Net}. 
We proceed describing the steps that implement one round of the 5-fold cross validation procedure (see figure \ref{fig_setup_melov}), which are repeated 100 times on both the filtered matrices \textbf{\textit{fM}} computed from the Melanoma and Ovarian cancer datasets: 

\begin{enumerate}
\item The \textit{n} samples in the dataset \textbf{\textit{fM}} are randomly split  in 5 subsets with the same size or as similar as possible.
 
\item The first subset is selected as test set and the remaining four subsets make up the training set. 

\item At this point, we apply a feature selection method to select a subset of informative features from the dataset. We decided to use the \textbf{moderated t-statistic method} to extract the subset of features and, also this time, the feature selection step has to be performed only on the training set data to avoid the selection bias problem \cite{ambroise2002selection}. The moderated t-statistic is implemented using the \textit{lmFit} and \textit{eBayes} functions from the package \textit{limma} \cite{smyth2005limma} in R \cite{venables2016introduction} and for each iteration of the 5-fold CV we select the first 1000 features with the smallest p-value. The selected features could be different at each iteration and we could obtain five different \textbf{\textbf{tM}} matrices. For more details about the moderated t-statistic see \fullref{feat_sel}.
 
\item Each matrix \textbf{\textit{tM}} is used as input of the \textbf{P-Net} algorithm to compute five distinct kernel matrices \textbf{\textit{K}}. Then the optimisation of the edge threshold is carried out by internal leave-one-out only on the training set and the estimation of the scores is performed only on the test set data.
The evaluation of the performances of the algorithm requires the use of some measures (e.g. accuracy, AUC, recall). The computation of these measures starting from the patients' scores needs a \textit{score threshold}, which is necessary to split the patients into the two classes (good and poor prognosis). Basically, at the end of one round of the 5-fold cross validation, the P-Net algorithm returns as output one score for each patient and we can rank them with respect to the phenotype under study. To compute the performance measures we need a threshold to assign to each patient a label derived from the score. If the score of the patient is below the threshold we assign the patient to the group GP, if the score is above the threshold we assign the patient to the group PP. To find the optimal score threshold we compute the scores on the training set and then we test an arbitrary vector of quantiles. Each quantile is used as threshold to classify the patients (the classification is based on the scores which are computed on the training set) and to compute the corresponding accuracy. At the end, we select the \textit{optimal score threshold}, which is the quantile that returns the best accuracy. 

\item The score threshold is used to assign to each patient a label and to obtain the \textit{vector of predicted labels}.
 
\item We compare the \textit{predicted labels} with respect to the \textit{true labels} and we can compute the above mentioned performance measures. 

\item The precedent steps (2-6) are repeated 5 times and each time the fold select as test set changes.
\end{enumerate} 

The steps from 1 to 7 implement 1 round of the 5-fold cross validation and we repeat them 100 times.

\subsubsection{Further analysis}

The performances of the classification can be evaluated in different ways and, in the precedent paragraph, we mentioned a series of possible measures. In the reference paper, the authors propose a series of further criteria to evaluate the classification:

\begin{itemize}
\item \textbf{Stability:} We evaluate the stability of the feature selection method by considering the average number of selected features that are common to each pair of 500 cross-validation folds (this is the global number of folds obtained from 100 rounds of 5-fold cross-validation). In the paper, they assess the average stability when the top 20, 30, 40 and 50 features are selected in each cross-validation fold. However, the evaluation of the feature selection stability is not the goal of our work, so we limited our evaluation to the top 20 features. 

\item \textbf{Error estimation under 5-fold cross validation:} This is the main system to assess the classification performances. As we mentioned previously, the dataset is split into 5 subsets and each subset is used as test set for each round of the 5 fold-CV. At the end, we can compute 5 fold error rates, where one fold error-rate is defined as following:

$$ Ferror_{i}= \dfrac{number\:of\:test\:set\:misclassified\:patients}{number\:of\:total\:test\:set\:patients} $$

We repeat the precedent computation for each combination of training-test set and we obtain the following \textit{cross-validation error}:

$$ CVerror=\dfrac{\sum_{i=1}^{5}Ferror_{i}}{5} $$ 

The final error is defined as the average of the CVerrors over the 100 rounds of 5-fold CV. 

\item \textbf{Accuracy analysis at a patient-level:} in this case, we want to assess the classification accuracy related to each patient. For each round of the 5-fold CV, each individual patient is included only in one fold. This means that, at the end of one round, we obtain a single predicted class for each patient. At this point, to compute the \textit{patient error rate} we use the formula: 

$$ Patient\:error\:rate=\dfrac{number\:of\:misclassified\:patients}{100\:rounds} $$

\item \textbf{Class-specific error rate:} This error rate is computed considering only the classification of the patients inside the two separated classes. The goal is to understand if there is a class that is easier to classify than the other one. Also in this case, the error rates derive from the average class-specific error rates over the 100 rounds of 5-fold CV.  
\end{itemize}

\subsection{Results with fixed number of selected features} \label{res_melov}

\begin{figure}[H]
\centering
\includegraphics[scale=0.6]{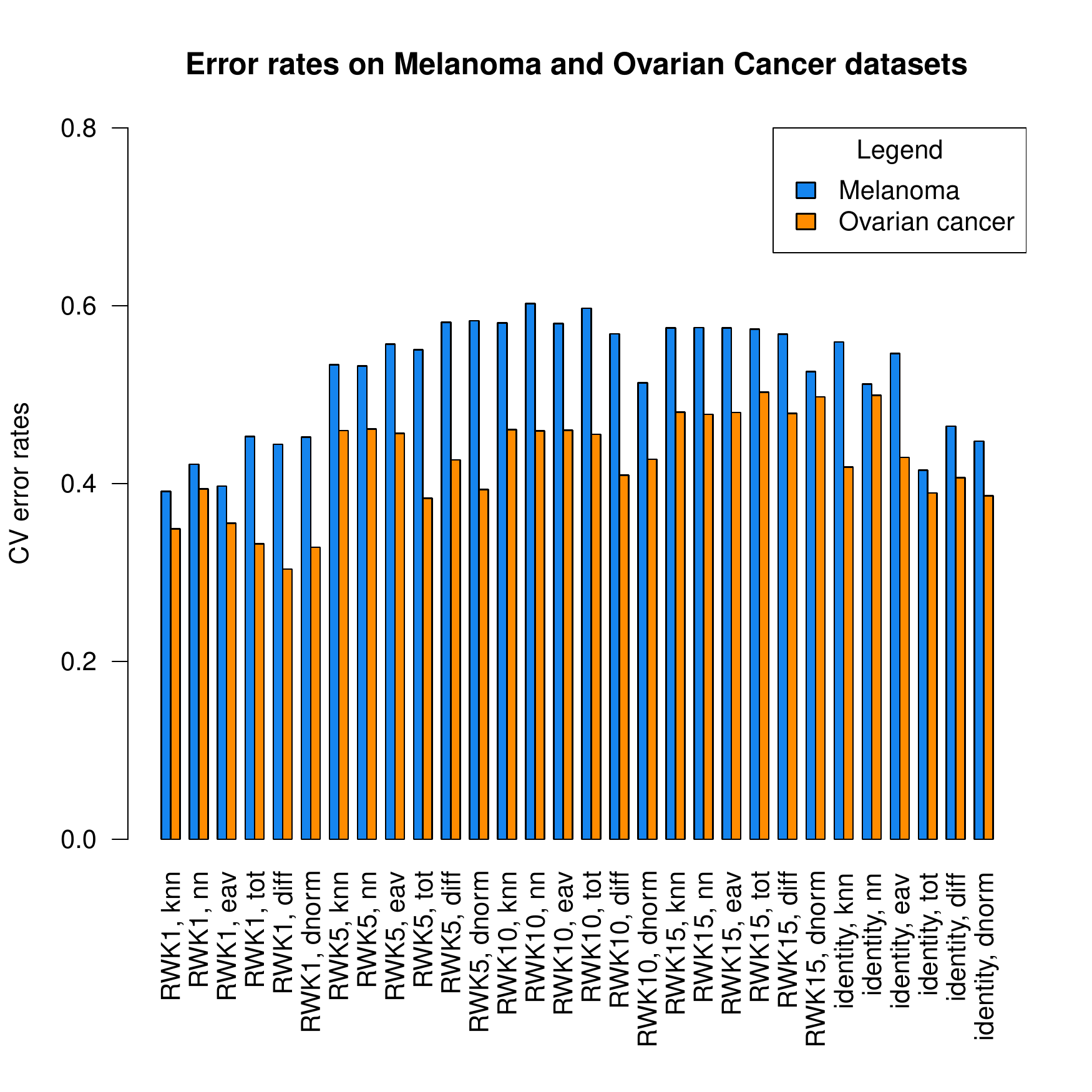}
\caption[Plot of the Cross-validation error rates on Melanoma and Ovarian cancer datasets]{\textbf{Plot of the Cross-validation error rates on the Melanoma and Ovarian cancer datasets.} The error rates of the Melanoma dataset are showed in blue, the Ovarian cancer ones in orange. In the plot there are the error rates for each kernel method and kernelized score function assessed.} \label{fig_plot_CVerror}
\end{figure}

The results showed in this section are obtained following the experimental set-up described in the subsection \ref{setup_mel_ov}.\\
The first part of the experimental procedure consists in the application of the moderated t-statistic as features selection method followed by P-Net, as classifier, under 100 rounds of 5-fold CV. In the moderated t-statistic we select as gene signature the first 1000 features with the smallest p-value and we use only these features to built the correlation matrix \textbf{\textit{W}}. At the end of the entire procedure, we compute the \textit{cross-validation error} ($CVerror$) as described previously.\\
We decided to try in the P-Net algorithm different types of \textit{kernels} to understand which is the method that achieves the best performances. We decided to test the \textit{p-step Random Walk Kernel} with different values of \textit{p}(1,5,10,15). Since the best results occur with the 1-step Random Walk Kernel, we decided to try also the \textit{Identity kernel}, which leaves the initial matrix \textbf{\textit{W}} unchanged in P-Net (see \fullref{type_kernel} for more details). Moreover, for each kind of kernel method we try all the available \textit{kernelized score functions} described in \ref{step4}.\\
The cross-validation errors obtained from this first series of experiments on the Melanoma dataset and Ovarian cancer dataset are illustrated in the table \ref{table_CVerror_melov} and in the plot \ref{fig_plot_CVerror}. As can be seen, the best cross-validation error rate is achieved by the 1-step Random Walk Kernel, in both the datasets. However, the kernelized score function with the best result is the \textit{k-Nearest Neighbour score function} ($39.11\%$) in the Melanoma dataset but it is the \textit{Differential score} ($30.38\%$) in the Ovarian cancer dataset. From the plot \ref{fig_plot_CVerror} we can notice that the error rates for the Melanoma dataset are always higher than the error rates for the Ovarian cancer dataset. The patients in the Melanoma dataset seem harder to classify for P-Net. This observation is in contrast with the results achieved from the other methods evaluated in the reference paper.\\
Another interesting point is that in both the datasets the performances tend to decrease with the increasing of the number of steps in the Random Walk Kernel. This is an opposite behaviour with respect to the trend observed in the pancreatic cancer dataset \ref{exp_ggc_extensive} and it is probably related to the differences in the topological characteristics of the patient networks. 

\begin{table}[H]
\centering
\begin{tabular}{|c|c|c|}
\hline
\textit{\textbf{Method}} & \textit{\textbf{\begin{tabular}[c]{@{}c@{}}CV error rate\\ (Melanoma)\end{tabular}}} & \textit{\textbf{\begin{tabular}[c]{@{}c@{}}CV error rate\\ (Ovarian cancer)\end{tabular}}} \\ \hline
\textit{RWK1, k-NN} & 39.11\% & 34.90\% \\ \hline
\textit{RWK1, NN} & 42.18\% & 39.42\% \\ \hline
\textit{RWK1, EAV} & 39.72\% & 35.55\% \\ \hline
\textit{RWK1, TOT} & 45.29\% & 33.22\% \\ \hline
\textit{RWK1, DIFF} & 44.42\% & 30.38\% \\ \hline
\textit{RWK1, DNORM} & 45.21\% & 32.85\% \\ \hline
\textit{RWK5, k-NN} & 53.37\% & 45.97\% \\ \hline
\textit{RWK5, NN} & 53.21\% & 46.12\% \\ \hline
\textit{RWK5, EAV} & 55.69\% & 45.67\% \\ \hline
\textit{RWK5, TOT} & 55.06\% & 38.34\% \\ \hline
\textit{RWK5, DIFF} & 58.15\% & 42.66\% \\ \hline
\textit{RWK5, DNORM} & 58.31\% & 39.33\% \\ \hline
\textit{RWK10, k-NN} & 58.09\% & 46.07\% \\ \hline
\textit{RWK10, NN} & 60.23\% & 45.92\% \\ \hline
\textit{RWK10, EAV} & 58.00\% & 45.98\% \\ \hline
\textit{RWK10, TOT} & 59.72\% & 45.53\% \\ \hline
\textit{RWK10, DIFF} & 56.83\% & 40.93\% \\ \hline
\textit{RWK10, DNORM} & 51.34\% & 42.73\% \\ \hline
\textit{RWK15, k-NN} & 57.49\% & 48.05\% \\ \hline
\textit{RWK15, NN} & 57.54\% & 47.80\% \\ \hline
\textit{RWK15, EAV} & 57.49\% & 47.99\% \\ \hline
\textit{RWK15, TOT} & 57.37\% & 50.27\% \\ \hline
\textit{RWK15, DIFF} & 56.82\% & 47.89\% \\ \hline
\textit{RWK15, DNORM} & 52.60\% & 49.76\% \\ \hline
\textit{identity, k-NN} & 55.92\% & 41.84\% \\ \hline
\textit{identity, NN} & 51.19\% & 49.94\% \\ \hline
\textit{identity, EAV} & 54.64\% & 42.94\% \\ \hline
\textit{identity, TOT} & 41.49\% & 38.96\% \\ \hline
\textit{identity, DIFF} & 46.47\% & 40.67\% \\ \hline
\textit{identity, DNORM} & 44.78\% & 38.62\% \\ \hline
\end{tabular}%
\caption[Cross-validation error rates on the Melanoma and Ovarian cancer datasets]{\textbf{Cross-validation error rates on the Melanoma and Ovarian cancer datasets using different kernel methods and kernelized score functions.}}
\label{table_CVerror_melov}
\end{table}

\subsubsection{Comparison with the results in the reference paper}

\begin{figure}[H]
\centering
\includegraphics[scale=0.6]{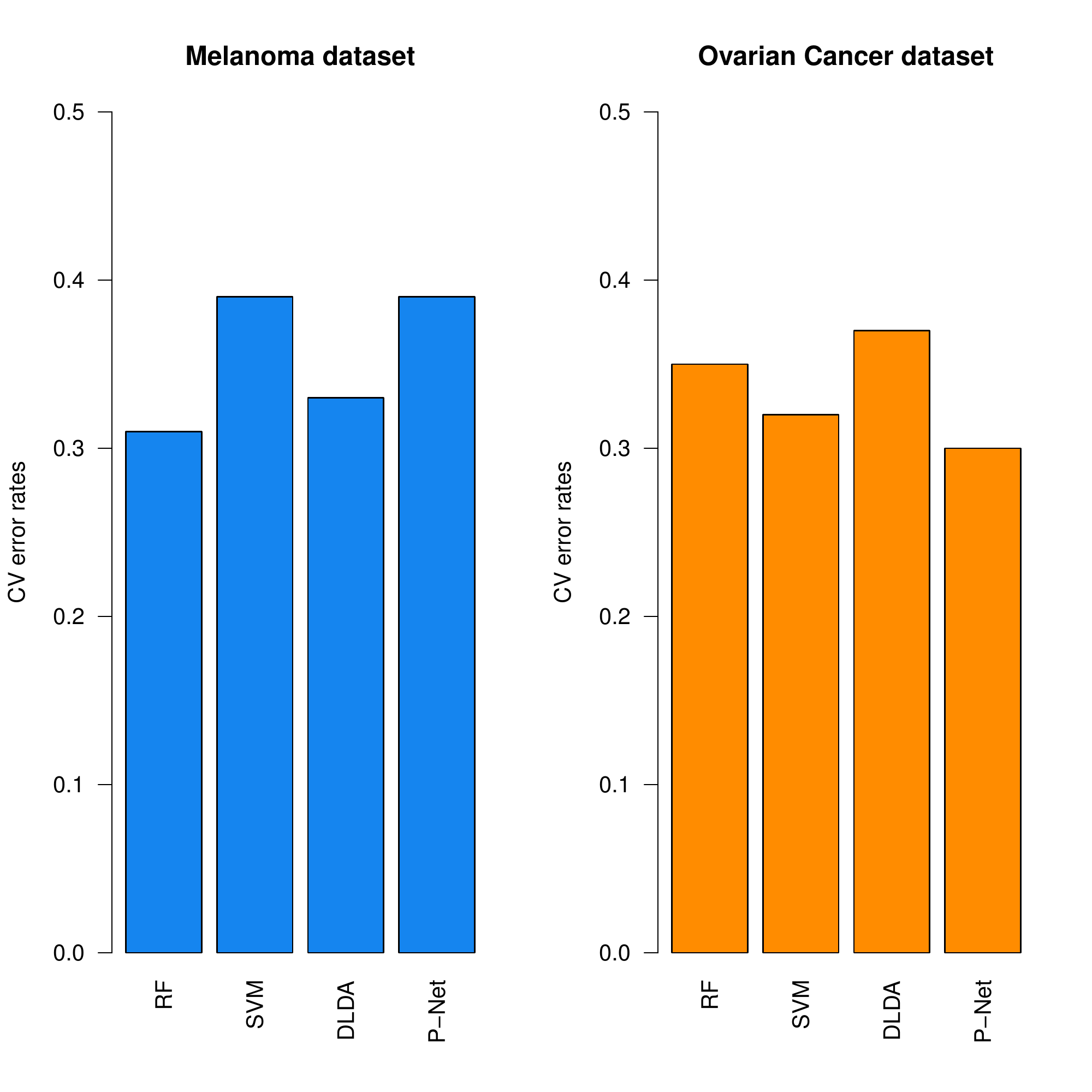}
\caption[Plots to compare the CV error rates of the different classifiers on Melanoma and Ovarian cancer datasets]{\textbf{Plots to compare the CV error rates of the different classifiers on Melanoma and Ovarian cancer datasets.}} 
\label{fig_barplot_comparison_melov}
\end{figure}

In the reference paper the authors evaluated five different feature selection methods: \textit{moderated t-statistic}, \textit{Median expression}, \textit{NetRank}, \textit{Taylor's Method} and \textit{BSS/WSS approach}. For each of these methods, the error rates were estimated using 100 rounds of 5-fold cross-validation and the following classifiers: Random Forest (RF), Support Vector Machine (SVM) and diagonal linear discriminant analysis (DLDA). In our experiments, we use the \textit{moderated t-statistic} as feature selection method and P-Net as classifier. So, to compare the P-Net results with the cross-validation error rates achieved in the reference paper, we consider only the results achieved with the use of the \textit{moderated t-statistic} to select the most significant features in the reference paper. The CV error rates are showed in the table \ref{table_CVerror_paper} and in the plot \ref{fig_barplot_comparison_melov}. 

\begin{table}[H]
\centering
\resizebox{9cm}{!}{%
\begin{tabular}{c|c|c|}
\cline{2-3}
\multicolumn{1}{l|}{} & \multicolumn{2}{c|}{\textit{\textbf{CV error rates}}} \\ \cline{2-3} 
\multicolumn{1}{l|}{} & \textit{\textbf{Melanoma dataset}} & \textit{\textbf{Ovarian Cancer dataset}} \\ \hline
\multicolumn{1}{|c|}{\textit{\textbf{RF}}} & \cellcolor[HTML]{FCFF2F}31\% & 35\% \\ \hline
\multicolumn{1}{|c|}{\textit{\textbf{SVM}}} & 39\% & 32\% \\ \hline
\multicolumn{1}{|c|}{\textit{\textbf{DLDA}}} & 33\% & 37\% \\ \hline
\multicolumn{1}{|c|}{\textit{\textbf{P-Net}}} & 39\% & \cellcolor[HTML]{FCFF2F}30\% \\ \hline
\end{tabular}%
}
\caption[Cross-Validation error rates on Melanoma and Ovarian cancer datasets for different classifiers]{\textbf{Cross-validation error rates achieved from the reference paper and P-Net on the Melanoma and Ovarian Cancer datasets.} All the classifiers use the \textit{moderated t-statistic} as feature selection method. The yellow background of some cells indicates the best results obtained in the specific dataset.}
\label{table_CVerror_paper}
\end{table}

The best CV error rates obtained from the baseline methods are $31\%$ on the Melanoma dataset and $32\%$ on the Ovarian cancer dataset. The best results achieved by our method P-Net are $39\%$ and $30\%$, respectively. P-Net achieves the lowest error rate on the Ovarian Cancer dataset and it obtains the same error rate of the SVM classifier on the Melanoma dataset. 

The cross-validation error rate is just one of the statistical analysis we carry out on these two datasets. We also evaluate the stability of the feature selection method with respect to the selection of the top 20 features, the accuracy analysis at a patient-level and the class-specific error rate. 

\begin{figure}[H]
\centering
\includegraphics[scale=1]{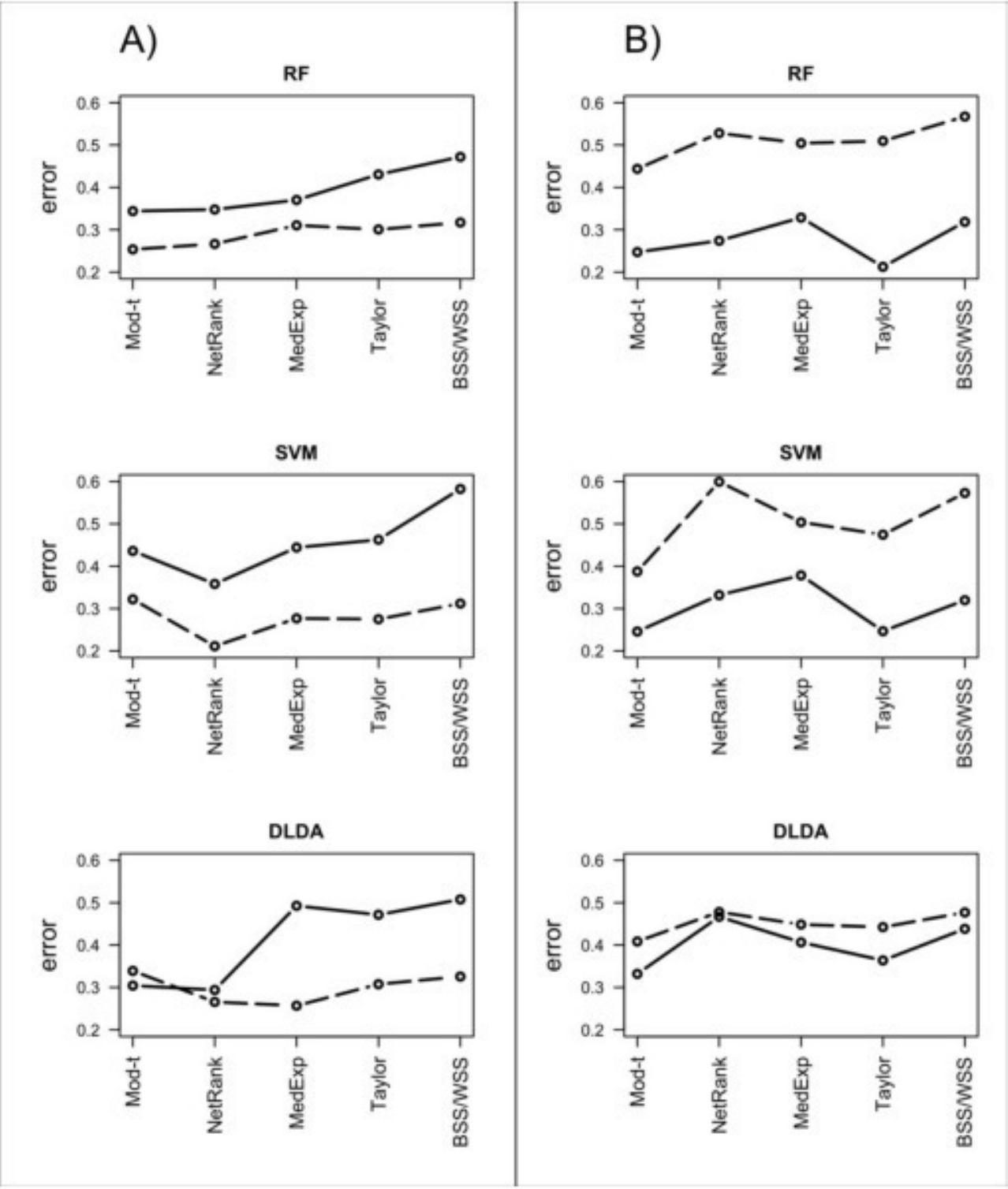}
\caption[Class-specific classification error rates from the reference paper \cite{barter2014network}]{\textbf{Class-specific classification error rates for the classifiers and feature selection methods assessed in the paper \cite{barter2014network}.} The GP (dotted line) and PP (solid line) error rates for each feature selection method are presented for the RF classifier, SVM classifier and DLDA classifier using A) the melanoma dataset and B) the ovarian cancer dataset. Image from \cite{barter2014network}.} 
\label{fig_class_specific_er_paper}
\end{figure}

\begin{figure}[H]
\centering
\includegraphics[scale=0.8]{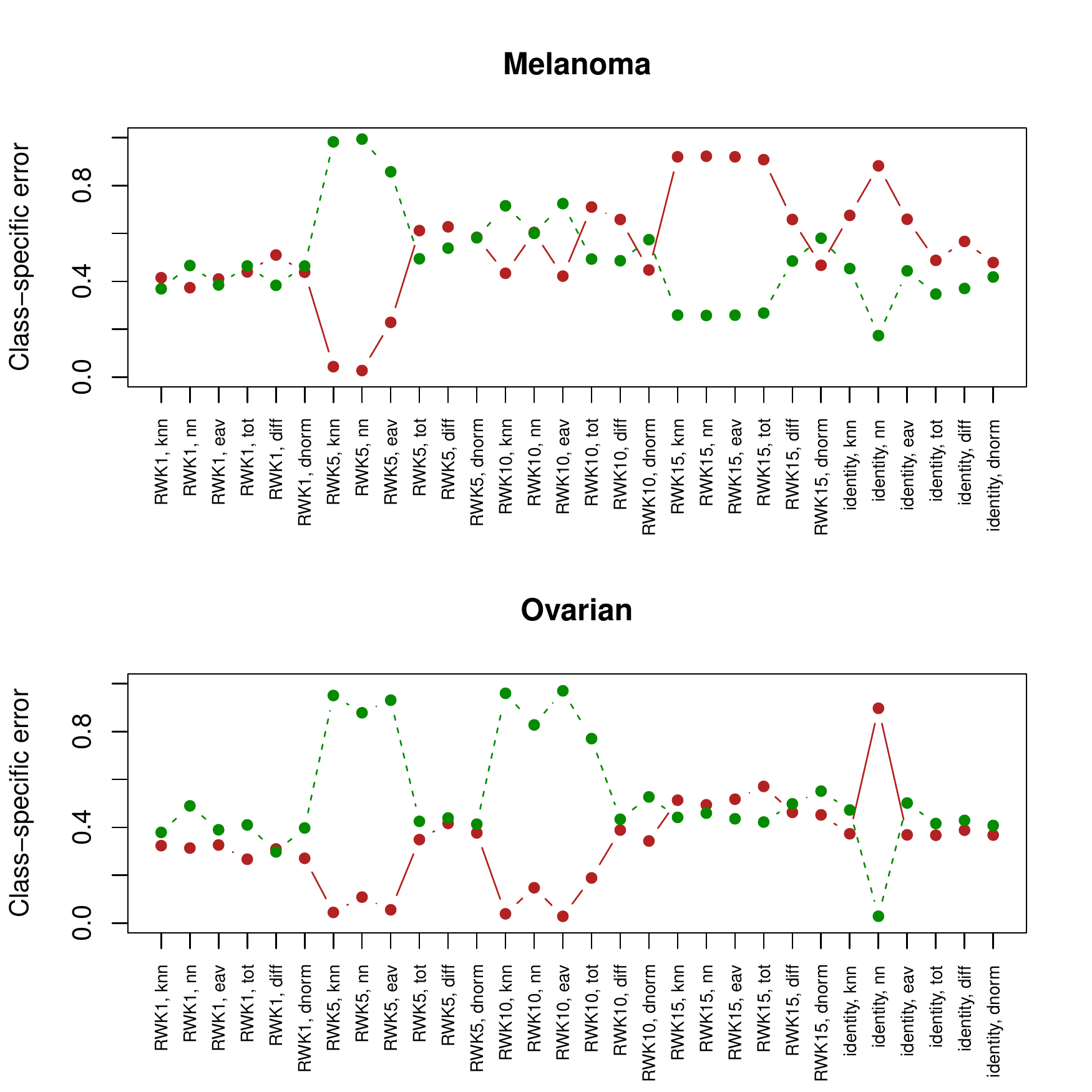}
\caption[Plot of the class-specific error rates on Melanoma and Ovarian cancer datasets]{\textbf{Plot of the class-specific error rates on Melanoma and Ovarian cancer datasets.} In the plot there are the error rates for each kernel method and kernelized score function assessed. The error rates related to the \textit{poor prognosis} and \textit{good prognosis} class are showed in red and green, respectively.} 
\label{fig_plot_classerror_allkernel}
\end{figure}

\begin{figure}[H]
\centering
\includegraphics[scale=0.50]{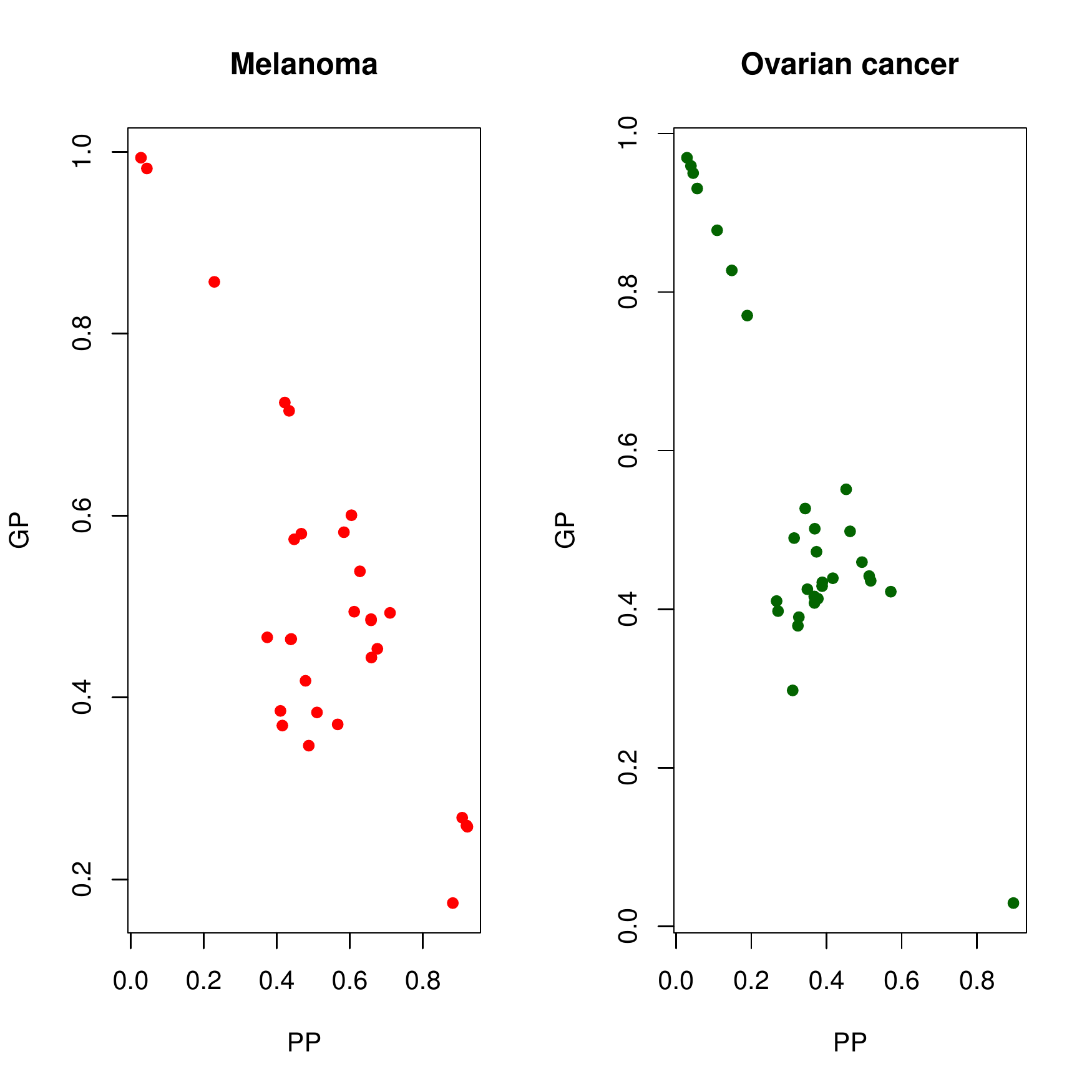}
\caption[Plot to assess the proportionality between GP and PP error rates]{\textbf{Plot to assess the proportionality between GP and PP class-specific error rates.} The x-axis is the PP error rate and the y-axis is the GP error rate.} 
\label{fig_plot_PPvsGP}
\end{figure}

\paragraph{Class-specific error rates} 
We evaluate the class-specific error rates for each combination of \textit{kernel methods} and \textit{kernelized score functions} assessed on the two datasets (see figure \ref{fig_plot_classerror_allkernel}). In the Melanoma dataset in most of the cases the patients belonging to the good prognosis class are easier to classify and this observation confirms the results in the reference paper. Instead, in the Ovarian cancer dataset in most of the cases the patients with poor prognosis are easier to classify and the results of the paper are confirmed again. It is worth noticing that the class-specific error rates seem inversely proportional. In other words, higher is the error rate in one class, lower is the error rate in the other one. We tried to plot the class specific error rates for the two classes to see if there is this correlation between GP and PP error rates (see figure \ref{fig_plot_PPvsGP}). As we can see, the proportionality between the class-specific error rates for the two classes is not true for all the combinations of kernel methods and score functions.\\   
Now we continue comparing the class-specific error rates achieved from P-Net with respect to the baseline methods in the paper. Unfortunately, in the reference paper, the class-specific error rates obtained from the three assessed classifiers (RF, SVM and DLDA) using the \textit{moderated t-statistic} as feature selection method are showed only in the below figure \ref{fig_class_specific_er_paper} and the specific values are not available. Considering that our purpose is to compare the results of P-Net with the results of the three classifiers using the same feature selection method (Mod-t) for all the predictors, we have to refer to the following plots.  

In the Melanoma dataset, P-Net obtains $41.52\%$ of error rate for the PP class that is slightly better than the result achieved by SVM but worst than the results achieved from RF and DLDA classifiers. Instead, P-Net obtains an error rate of $36.91\%$ for the GP class, which is worst respect to the other methods (even if it is close to the results achieved from SVM and DLDA classifiers).\\
In the Ovarian cancer dataset, P-Net achieves $31\%$ of error rate in the PP class and $29.76\%$ of error rate in the GP class. The former result is worst than the RF classifier and SVM classifier but it performs slightly better than the DLDA classifier. The latter result outperforms all the baseline methods (with a rough difference of 12\% with respect to the RF classifier and 10\% for the SVM and DLDA classifiers).
 
\paragraph{Evaluation of the stability}

In our experimental set-up, we evaluate the stability related to the top 20 features. Basically, we consider the number of top 20 selected features pair-wise in common over the 100 rounds of 5-fold cross-validation. We use the \emph{moderated t-statistic} as feature selection method, which is employed also in the reference paper. So, we expect that our stability has to be similar to the stability achieved in the reference paper. P-Net obtains a stability of $34\%$ in the Melanoma dataset and $29\%$ in the Ovarian dataset. The stability achieved from the paper is $38\%$ in the Melanoma dataset and a comparable value in the Ovarian cancer dataset. The difference in the features stability between our work and the reference paper is probably related to the different gene annotation and different number of selected genes through the PPI network (as explained in subsection \ref{setup_mel_ov}). 

\paragraph{Accuracy analysis at a patient-level}

\begin{figure}[H]
\centering
\includegraphics[scale=0.8]{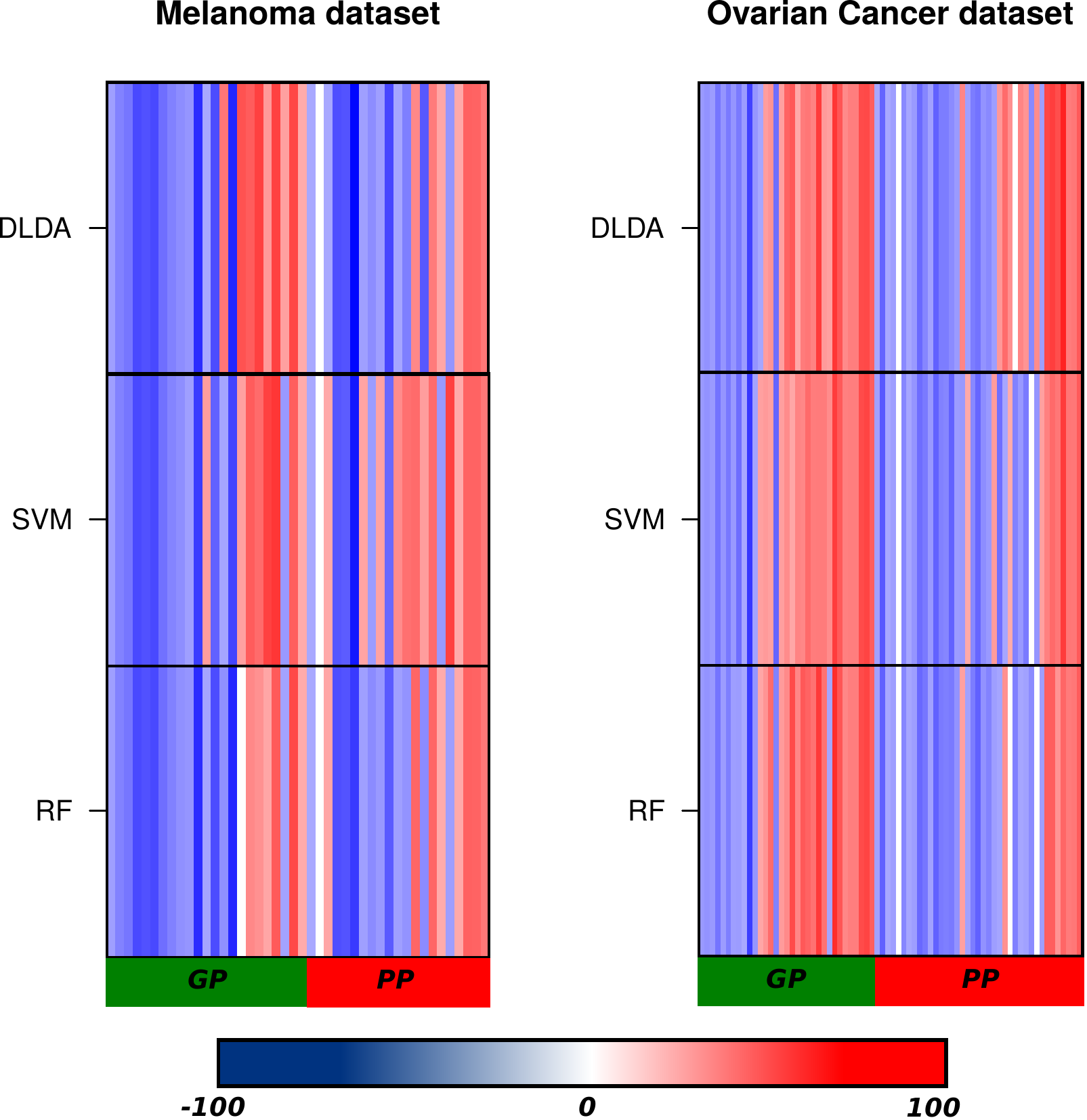}
\caption[Heatmaps of the relative classification accuracies at the patient level]{\textbf{Heatmaps of the relative classification accuracies at the patient level.} The analysis in the left column refers to the Melanoma dataset and the analysis in the right column refers to the Ovarian cancer dataset. Each cell represents the difference between the accuracy achieved by P-Net and the accuracy obtained by another classifier (RF, SVM or DLDA) for a specific patient. The value of a cell is 100 if P-Net classifies correctly the evaluated patient in all the 100 rounds of the 5-fold cross-validation and the other classifier achieves 0 of accuracy. The opposite situation leads to a value of -100. If P-Net and the evaluated classifier obtain the same result, the cell has value zero and the corresponding colour is white. Positive accuracy values show red shades and negative values show blue shades. In the y-axis, there are the employed classification methods, whose accuracy values are subtracted from the accuracy of P-Net. In the x-axis, there is the label of each patient, where the green colour refers to patients with a \textit{good prognosis - GP} and the red colour refers to patients with a \textit{poor prognosis - PP}.} 
\label{fig_heatmap_differences_melov}
\end{figure}

We compared the patient-specific accuracies obtained from P-Net with respect to the accuracies obtained by the other methods (see figure \ref{fig_heatmap_differences_melov}). 
As we can see, the accuracies achieved from P-Net are significantly different compared to the accuracies achieved by the other classifiers, which use the \textit{moderated t-statistic} as feature selection method.
In the melanoma dataset, there are 15 patients that are almost always classified correctly by P-Net ($accuracy \geq 80$) and 5 patients never classified correctly ($accuracy \leq 20$).
In the ovarian cancer dataset, there are 28 patients that are almost always classified correctly by P-Net ($accuracy \geq 80$) and 0 patients never classified correctly ($accuracy \leq 20$).
In both the datasets, we can see that P-Net is able to improve the accuracy of some patients that are generally difficult to classify by the other methods. This is true especially if we look at the patients belonging to the good prognosis class in both the datasets. We can identify these patients from the red colour of the cells in all the rows of the heatmaps in the figure \ref{fig_heatmap_differences_melov}.\\
The fact that P-Net has a typical patient-specific behaviour with respect to the other methods is an advantage. Indeed, we can think to use P-Net in combination with other methods to improve the accuracy of the classification.  

\subsection{Results of P-Net with different numbers of features selected on the Melanoma dataset}

\begin{figure}[H]
\centering
\includegraphics[scale=0.7]{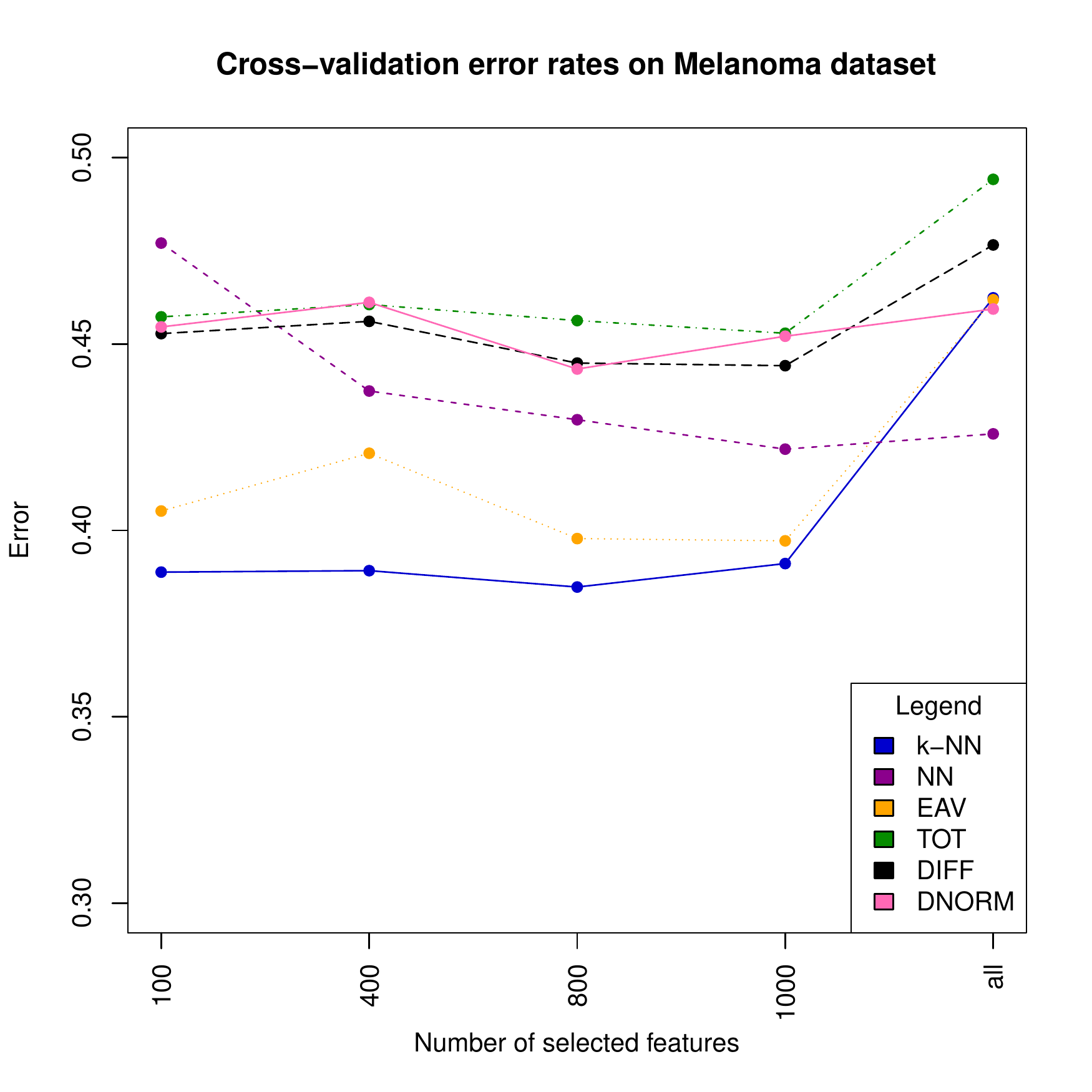}
\caption[Plot of the Cross-validation error rates for different numbers of features selected on the Melanoma dataset]{\textbf{Plot of the  Cross-validation error rates for different numbers of features selected on the Melanoma dataset.} In all the experiments we use the 1-step Random Walk kernel and we test all the available \textit{kernelized score functions}.} 
\label{fig_plot_melanoma_nfeat}
\end{figure}

In the precedent subsection \ref{res_melov}, we evaluate the performances of our method, which consists in the use of moderated t-static followed by P-Net as classifier, and in both the datasets we select the first \textit{n=1000} features with the smaller p-value. The choice to select 1000 features is completely arbitrary and it does not exclude the possibility that the selection of a different number \textit{n} of features can lead to lowest cross-validation error rates and better performances in general. Indeed, it is possible that the most informative features are less than the first 1000 and we considered some additional features, which are not so important with respect to the classification task. The introduction of these additional features in our analysis could introduce problems of statistical noise. However, it is also possible that the most significant features are more than 1000 and we avoided to use part of them in our analysis.
In this subsection, we want to assess the impact of the number of selected features on the performances of P-Net to see if we can improve the results. We repeated the entire experimental set-up changing the value of \textit{n} to 100, 400, 800 and 9433 (all the available features) on the Melanoma dataset. In these tests, we used the \textit{1-step Random Walk kernel}, which is the kernel method that shows the best performance in the subsection \ref{res_melov}, and all the available \textit{kernelized score functions}. As we can see in figure \ref{fig_plot_melanoma_nfeat}, the \textit{k-Nearest Neighbour score function} remains the score function that achieves the lowest error rates for all the tested numbers of selected features \textit{n}. The only exception concerns the use of all the available features where the best score function is the Nearest-neighbour score. \\
Moreover, we can see how the reduction of the number of selected features does not significantly improve the classification performances and the selection of 1000 features seems to be a good choice on this dataset.  

\section{Visualization of the network of patients} \label{visualization}
P-Net is a network-based method and this means we can look at our data in two different ways:
\begin{itemize}
\item Matrix of patients;
\item Weighted graph $G=<V,E,w>$ of patients. 
\end{itemize}

So, P-Net is able to rank and classify patients with respect to a specific phenotype \textit{C} under study but also to give us a different way to look at our data through a graph structure. This can be really useful because we can visualise which patients are close and how much they are close simply looking at the graph.\\
For example, we have the possibility to build a graph from the correlation matrix \textbf{\textit{W}} and the kernel matrix \textbf{\textit{K}} and then we can see how the weight of the edges changes after we apply a kernel method. Basically, we can see how the kernel method exploits the topology of the network and the emerging relationships between the patients that are closer in the graph. 

\paragraph{Graphs from the Pancreatic cancer dataset} \label{graph_pancreatic}

\begin{figure} [H]
\centering
  \begin{subfigure}[H]{1\textwidth}
  \centering
    \includegraphics[scale=0.40]{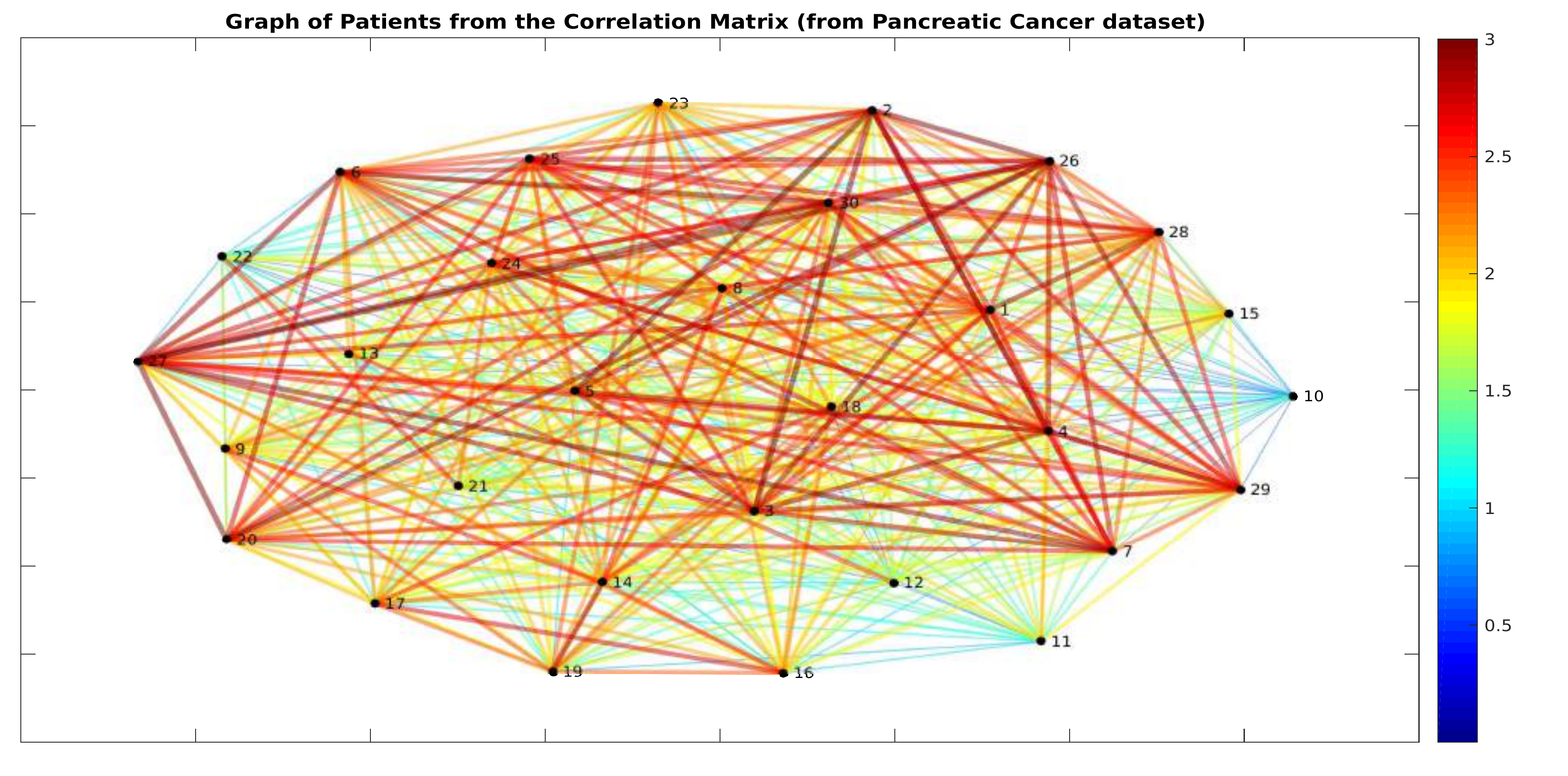}
  \end{subfigure}
  \begin{subfigure}[H]{1\textwidth}
  \centering
    \includegraphics[scale=0.40]{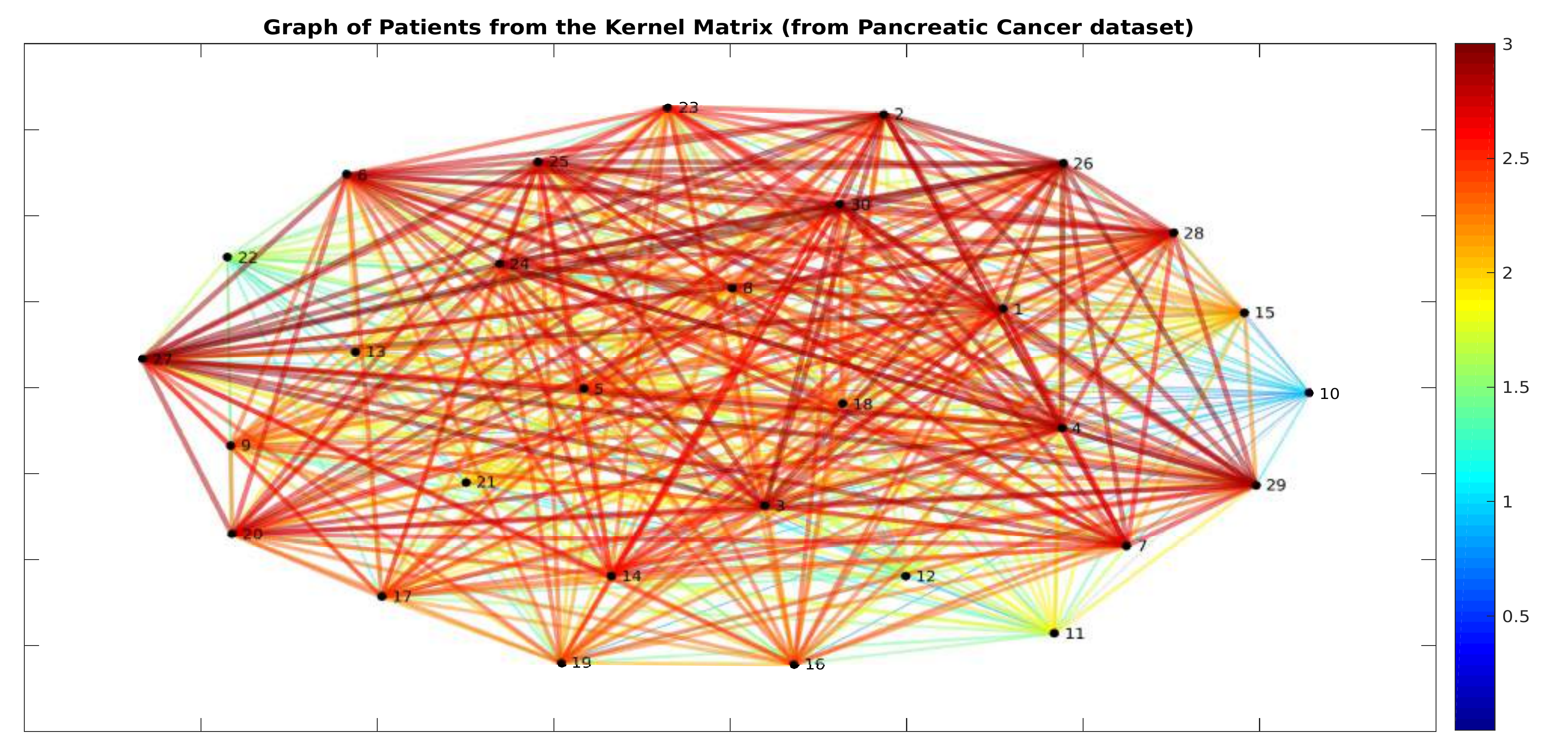}
  \end{subfigure}
  \caption[Graphs from the matrices \textbf{\textit{W}} and \textbf{\textit{K}} of the pancreatic cancer dataset.]{\textbf{Graphs from the matrices \textbf{\textit{W}} and \textbf{\textit{K}} of the pancreatic cancer dataset.} In both the graphs the nodes represent the patients and the edges represent the functional relationships between patients. The colour and thickness of the edges represent the weight of the corresponding edge. Higher is the weight higher is the thickness and the colour is closer to red.}
  \label{fig_graphs}
\end{figure}

The graphs corresponding to the matrices \textbf{\textit{W}} and \textbf{\textit{K}} for the Pancreatic cancer dataset are showed in figure \ref{fig_graphs}, where we applied the 8-step Random Walk Kernel. 

In the graph obtained from the correlation matrix \textbf{\textit{W}} of the pancreatic cancer dataset, we can see that the most part of the nodes is connected by edges with high weights (red edges) and only few nodes present edges with low weights (nodes 10, 11, 12, 15 and 22). We can exploit this graph to make some hypothesis about the possible score assigned to each node/sample. Indeed, we have to remember that both the correlation matrix and the kernel matrix can be inputs for P-Net; the only constraint is that the input matrix has to be a symmetric matrix representing similarities between the examples. For the sake of simplicity, we assume to use P-Net with the \textit{leave-one-out procedure} and to apply the \textit{Nearest Neighbour score function}. For example, we can consider the sample 2 and we can see from the graph that the connected nodes with an high correlation value are nodes 3, 4, 5 and 26. The patients 5 and 26 are labelled as GP so, by definition, they are not taken into account for the computation of the score (see subsection \ref{step4}). The remaining \textit{poor prognosis} neighbours have all an high weight, even if we consider the overall weights of the network. So, we can suppose that an high score will be assigned to the patient 2 and it will be probably labelled as \textit{poor prognosis}. It is important to notice that the translation of the score into a class is not so straightforward because it depends on the \textit{optimal score threshold} computed on the training set, which is necessary to split the patients into the two prognosis classes after the ranking. We can also make easier this kind of considerations painting the nodes with two different colours for each class of labels.

After the application of the 8-step Random Walk Kernel on the matrix \textbf{\textit{W}}, we obtain the kernel matrix \textbf{\textit{K}}. As wee can see, in the corresponding graph the weights of the edges are higher in general. So, it is more difficult make an hypothesis about the most important nodes that influence the score of a given sample. However, we can notice that the nodes connected with the rest of the graph through low weighted edges tend to retain this feature.  

\paragraph{Graphs from Melanoma and Ovarian cancer datasets}

\begin{figure} [H]
\centering
  \begin{subfigure}[H]{1\textwidth}
  \centering
    \includegraphics[scale=0.42]{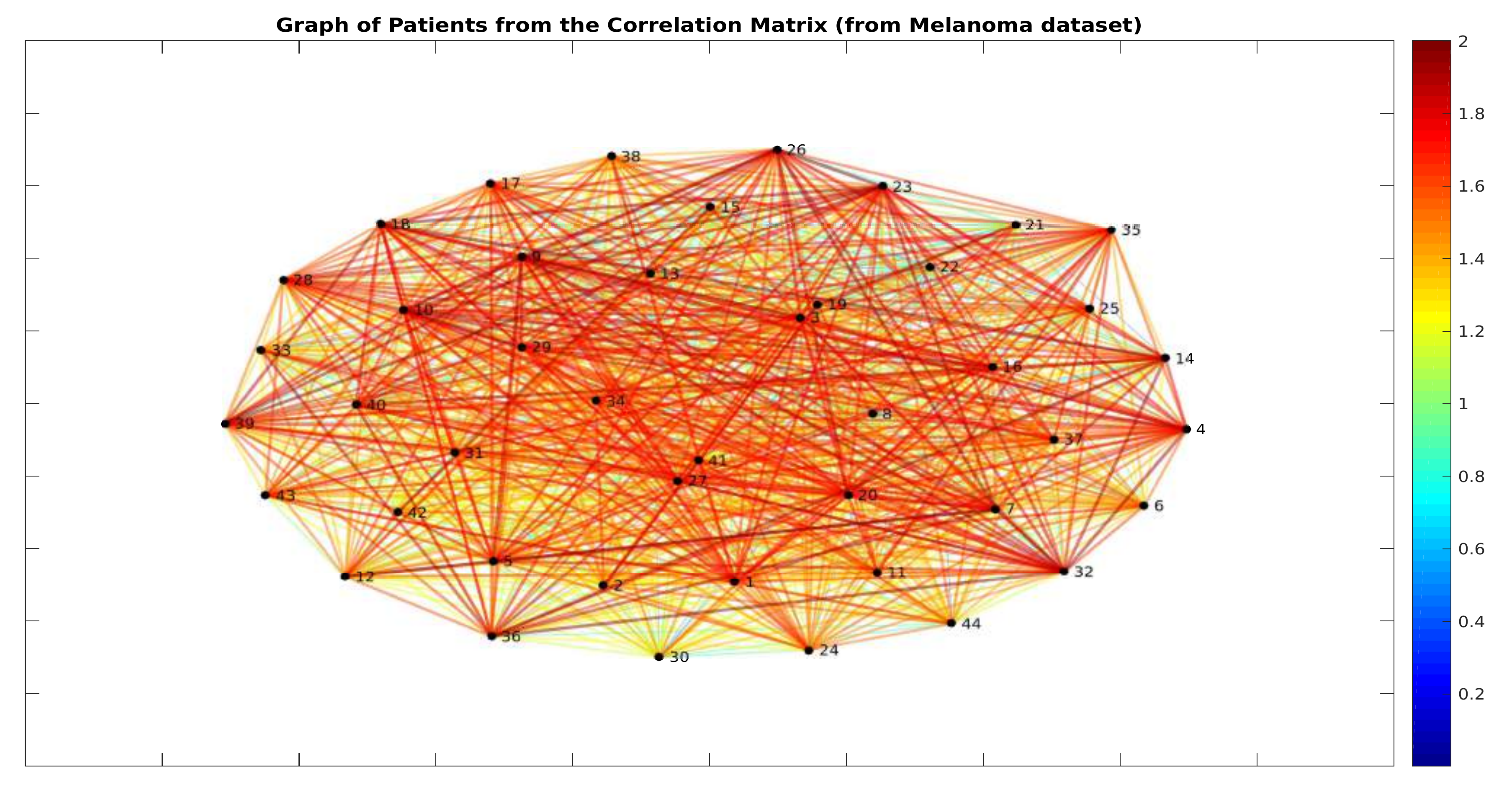}
  \end{subfigure}
  \begin{subfigure}[H]{1\textwidth}
  \centering
    \includegraphics[scale=0.42]{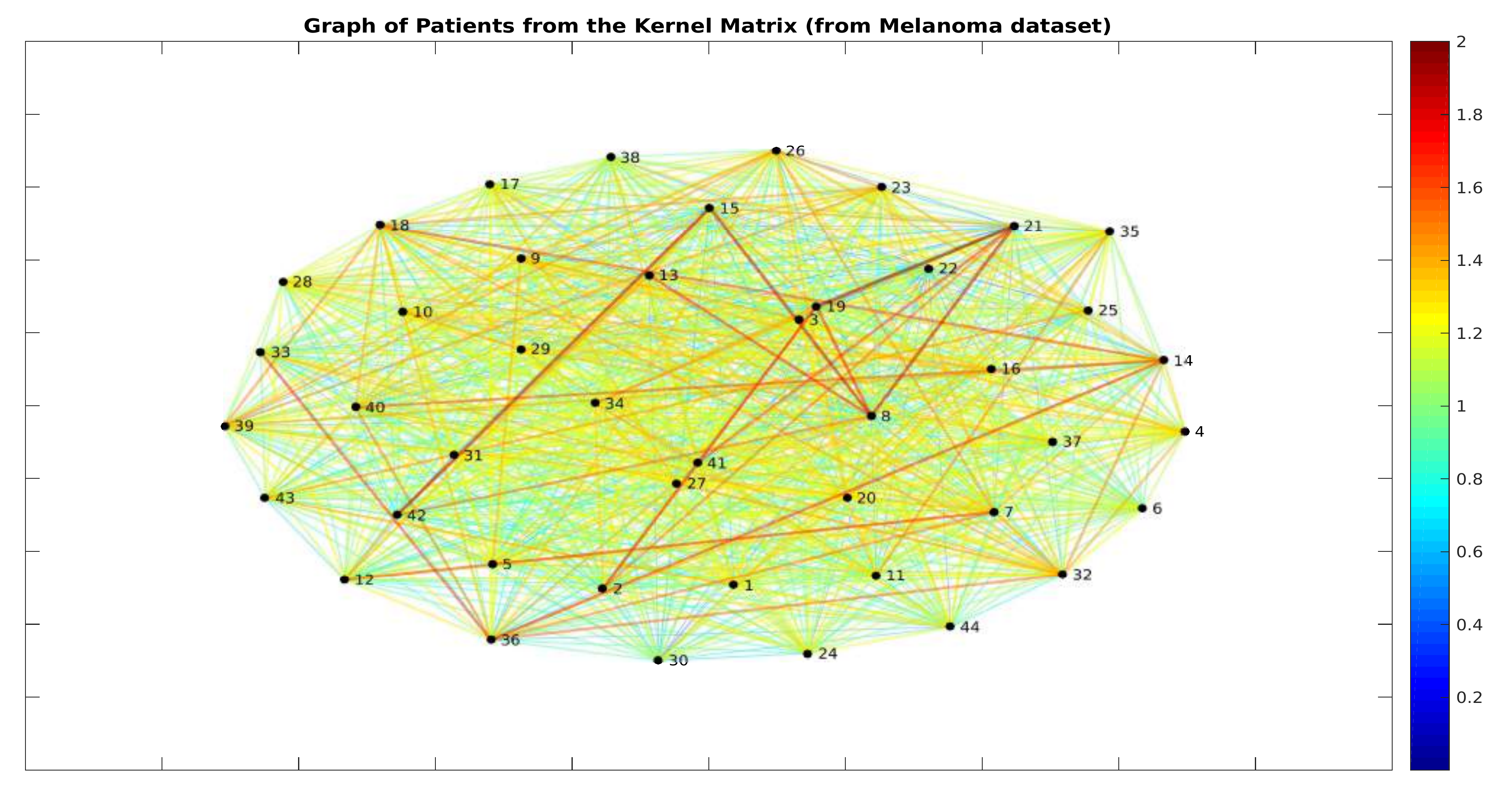}
  \end{subfigure}
  \caption[Graphs from the matrices \textbf{\textit{W}} and \textbf{\textit{K}} of the Melanoma dataset]{\textbf{Graphs from the matrices \textbf{\textit{W}} and \textbf{\textit{K}} of the Melanoma dataset.} In both the graphs the nodes represent the patients and the edges represent the functional relationships between patients. The colour and thickness of the edges represent the weight of the corresponding edge. Higher is the weight higher is the thickness and the colour is closer to red.}
  \label{fig_graphs_melanoma}
\end{figure}

\begin{figure} [H]
\centering
  \begin{subfigure}[H]{1\textwidth}
    \includegraphics[scale=0.39]{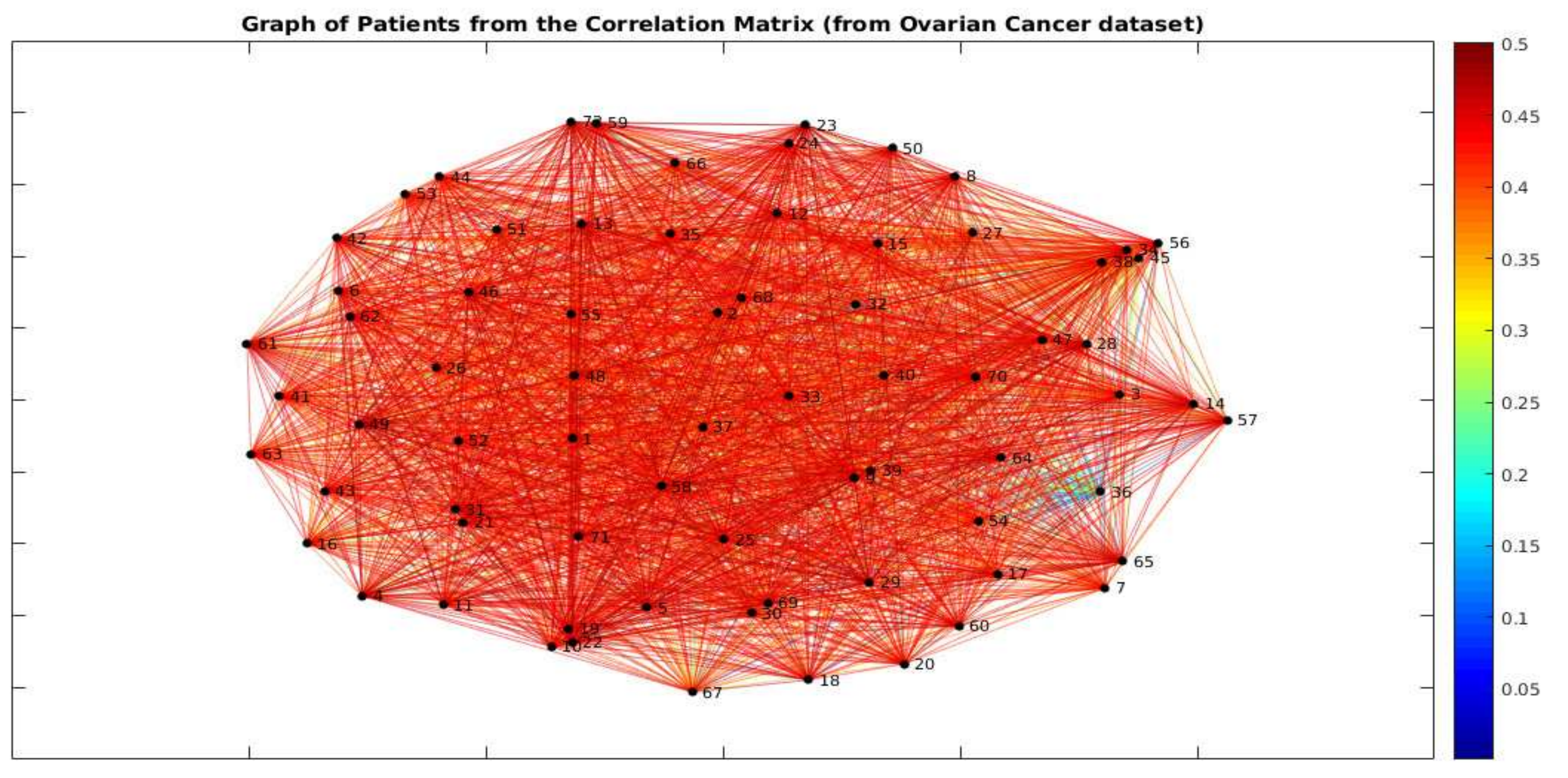}
  \end{subfigure}
  \begin{subfigure}[H]{1\textwidth}
    \includegraphics[scale=0.4]{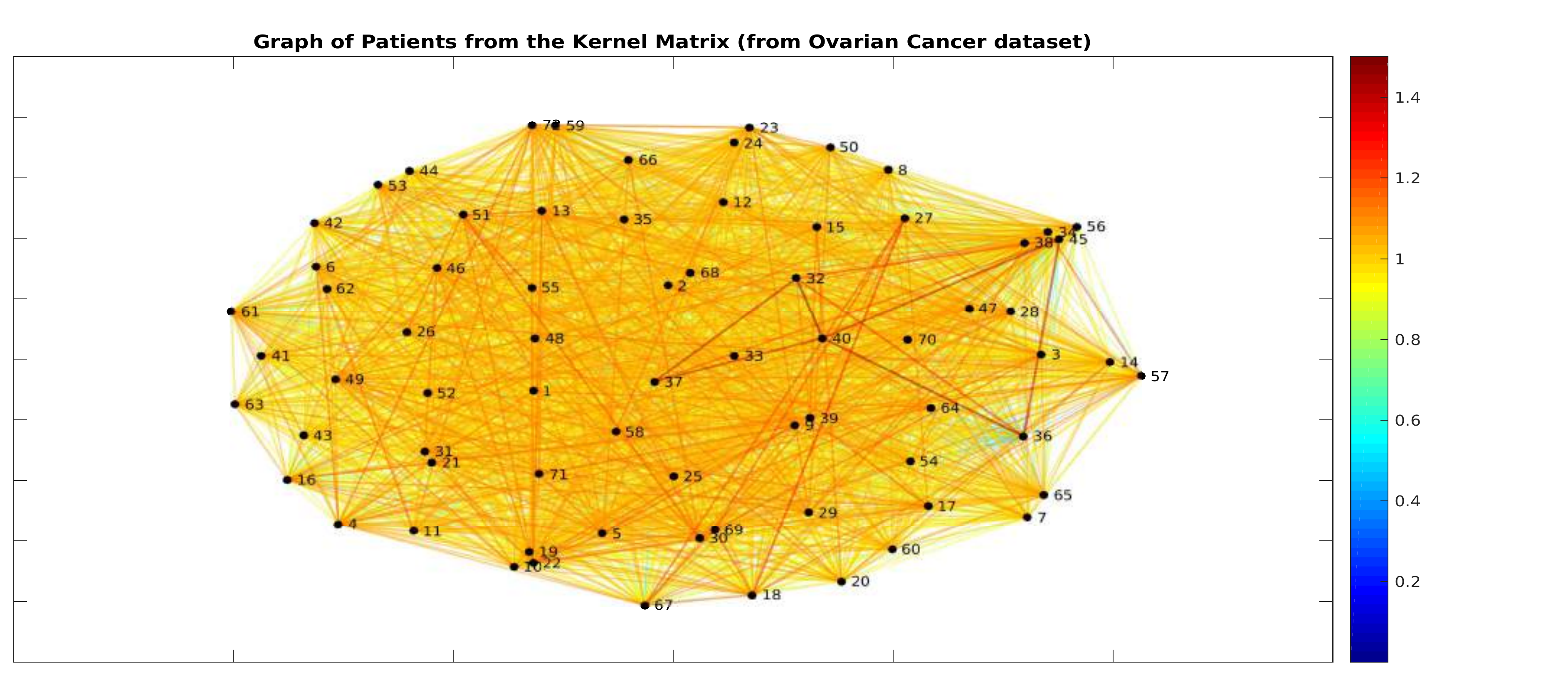}
  \end{subfigure}
  \caption[Graphs from the matrices \textbf{\textit{W}} and \textbf{\textit{K}} of the Ovarian cancer dataset]{\textbf{Graphs from the matrices \textbf{\textit{W}} and \textbf{\textit{K}} of the Ovarian cancer dataset.} In both the graphs the nodes represent the patients and the edges represent the functional relationships between patients. The colour and thickness of the edges represent the weight of the corresponding edge. Higher is the weight higher is the thickness and the colour is closer to red.}
  \label{fig_graphs_ovarian}
\end{figure}

In the figures \ref{fig_graphs_melanoma} and \ref{fig_graphs_ovarian} there are the graphs corresponding to the matrices \textbf{\textit{W}} and \textbf{\textit{K}} for the Melanoma and Ovarian cancer, respectively. To build all the graphs the 1-step Random Walk Kernel was used.

The graph obtained from the correlation matrix of the Melanoma dataset shows that almost all the nodes are connected by edges with high weights. In this way, it is difficult to exploit the graph to make considerations about the relationships between patients just looking at it. Moreover, if we apply P-Net on this graph the classification task could be really difficult for the algorithm. Indeed, the computed scores could be really similar and, in this situation, it is difficult to obtain a good-quality ranking.\\
After the use of the 1-step Random Walk Kernel, two groups of more similar nodes emerge from the graph. The first group involves the samples 2, 8, 13, 15, 19, 21, 41 and 42 and six of these patients are labelled as PP. The second group of patients include the nodes 14, 18, 32, 33, 36, 40 and five of these samples are labelled as GP. So, we found two clusters of patients with the same prognosis and it can be one of the reasons why the 1-step Random Walk kernel outperforms the Identity kernel on this dataset (as mentioned in subsection \ref {res_melov}). Indeed, patients belonging to a cluster are easier to rank and classify. Also in this case, we can try to make some hypothesis about the score of each patient as showed in the precedent paragraph.

Similar considerations can be stated about the graphs obtained from the matrices \textbf{\textit{W}} and \textit{\textbf{K}} for the Ovarian cancer dataset. In the graph from the correlation matrix, it is difficult to make some hypothesis about the relationships between patients because the weights of the edges are too similar each other. Moreover, a further complication in this graph is the number of patients (72 patients) that is higher than the number of samples in the other considered datasets. In this way, the graph becomes too dense and it is hard to distinguish individual edges.\\
However, after the application of the Random Walk Kernel, we can distinguish a group of patients that are closer in the graph (patients 32, 36, 37, 40 and 45). In this case, there are 3 PP patients and 2 GP patients in the group. So, even if there is this cluster it does not show a specific relationship with the prognosis.  

\paragraph{Graph of patients with related predicted scores}

\begin{figure}[H]
\centering
\includegraphics[scale=0.5]{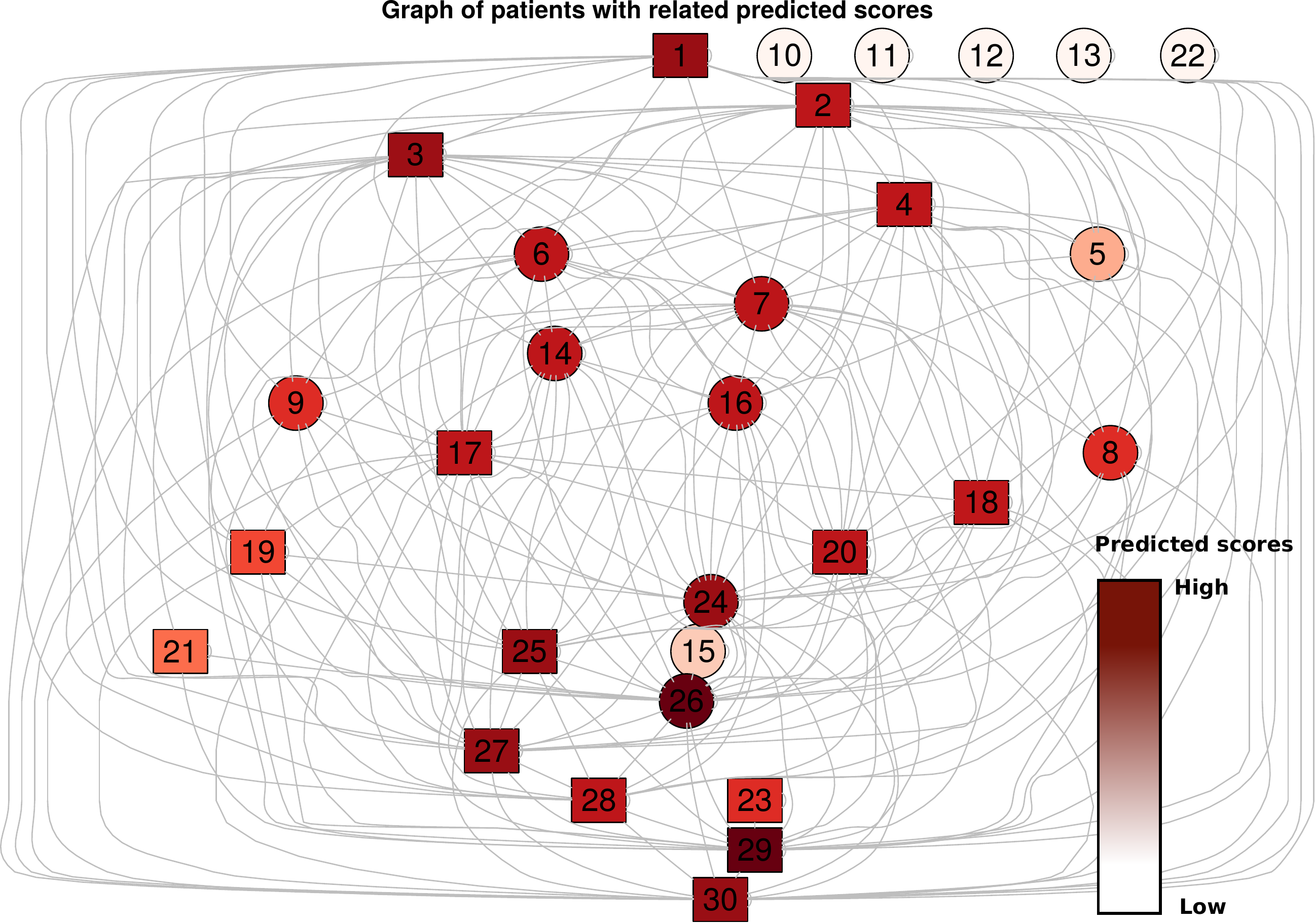}
\caption[Plot of the graph of patients with related predicted scores]{\textbf{Plot of the graph of patients with related predicted scores.} The graph is built from the P-Net results using the 8-steps Random Walk Kernel, training set size of 20 samples and Average kernelize score function. Squared and circle nodes represent respectively patients with poor and good prognosis.} \label{fig_plot_graph_scores}
\end{figure}

The interpretation of the data as a graph \textbf{\textit{G}} of patients gives us the ability to visualise also the score associated with each patient. We can obtain a graph where the nodes represent the patients and the edges represent the functional relationships between patients, e.g. the correlation coefficient between individuals. The true label, associated with each patient, can be  represented by different shapes of the nodes. In our case, the positive patients are represented by squares and the negative ones by circles. Finally, the score computed for each individual is represented through shaded colours and higher is the score more intense is the colour. An example of this kind of representation is showed in the figure \ref{fig_plot_graph_scores}, where the graph of patients and the predicted scores on the Pancreatic cancer dataset are represented.\\
We can exploit this type of representation to make some observations about the relationship between the predicted scores and the corresponding true labels for each patient. For example, we can see that the samples 10, 11, 12, 13, and 22 have a really low score so it is likely that they are labelled as GP, which is the right label. The same reasoning can be followed for the patients 5 and 15, which have a little bit higher scores. Instead, the patient 29 has a dark red shade and probably it will be classified correctly as a positive/poor prognosis patient. 
We can also spot that the patient 26 has a dark red colour but the true label is GP, so this patient it is probably misclassified. If we check the vector of the \textit{predicted labels} we can see that all the precedent hypothesis about the link between scores and predicted labels are correct. However, this kind of considerations are more difficult to do on the patients with intermediate shades of red because their labels are more related to the selection of the \textit{optimal score threshold}.  
 
We can conclude that the direct visual inspection of the \textit{graph of patients} can be useful to better understand how the P-Net algorithm works and it can provide further information about the phenotypic and/or genetic characteristics of the patients (depending on the type of data used to construct the graph). For example, we can have a visual clue about the presence of clusters in the graph. The next step is the use of the labelled graph obtained from P-Net to make further analysis using e.g. unsupervised and supervised clustering algorithms. In this way, we can discover novel significant grouping of patients that could be explained through the quantitative analysis of the features used to construct the network. 

\chapter{Conclusions} \label{Concl}

The primary purpose of this work was to develop and present the new network-based semi-supervised method, \textbf{P-Net}, which is able to rank and classify patients with respect to a specific phenotype or outcome under study (e.g. poor prognosis of patients afflicted with cancer). The choice to develop an algorithm with these particular characteristics comes from the necessity to deal with a series of drawbacks observed in the state-of-the-art for phenotype prediction tasks:

\begin{enumerate}
\item The supervised learning algorithms are the most popular methods used in the context of phenotype prediction and they usually exploit biomarker signatures selected by feature selection methods to address this task \cite{guyon2002gene}. However, recent works showed that this approach is often not statistically significant \cite{venet2011most} or not reliable due to the biased techniques adopted to evaluate the performances of these methods \cite{ambroise2002selection, michiels2005prediction}.  

\item The supervised methods do not take into account the functional or genetic relationships between individuals but these are extremely important has shown in the emerging field of \textit{Network Medicine} \cite{barabasi2011network}, which studies the organism's biology by graph structures, where nodes represent bio-molecules (e.g. genes, proteins) and edges represent functional or genetic relationships between nodes. 

\item Recently, many semi-supervised methods able to exploit network information were developed to rank genes and infer novel candidate disease genes or markers for specific diseases \cite{valentini2014extensive, winter2012google}. However, in these type of methods, the information about specific samples/patients is not taken into account because it is used collectively to define the biomolecular network.   
\end{enumerate}

To consider these problems, we developed the \textit{P-Net} algorithm that is able to consider both the patient's specific information and the functional/genetic relationships between patients. Indeed, the most innovative feature of P-Net is that it builds and uses a graph of patients, where the nodes represent the samples and the edges represent functional relationships between patients (e.g. correlation of the expression profiles). In other words, it builds the network in the ``sample space'' and not in the ``biomarker space'', as usual in state-of-the-art methods. 
To our knowledge, there are only a few methods that exploit networks in the ``sample space'' for phenotype or clinical outcome prediction \cite{park2014integrative, wang2014similarity}.

To compare our proposed method with state-of-the-art algorithms, we repeated the experiments proposed in two reference papers \cite{barter2014network, winter2012google}, where the authors compared the performances of network-based methods in the ``biomarker space'' with  state-of-the-art methods to select gene expression signatures and, at a later stage, used the extracted signatures to classify patients in the two classes \textit{poor prognosis} and \textit{good prognosis} through inductive supervised methods. We followed strictly the experimental set-up described in the two papers on the same publicly available datasets (Pancreatic cancer dataset, Melanoma dataset and Ovarian cancer dataset). Each of these datasets contains microarray data from patients afflicted with a specific type of tumour. P-Net exploits these data to build a patient-patient network, where each node is a patient and the edges express a functional relationship between two individuals. Then, our method uses the network to compute a score for each patient and to rank them exploiting the scores. Samples with an high score have an higher probability of presenting the phenotype under study. In the used datasets, we considered the outcome (\textit{poor prognosis} or \textit{good prognosis}) using a suitable \textit{score threshold}. So, we classify the patients of each dataset and we can also assess the accuracy (or similarly the error rates) of the classification because we know the \textit{true labels} of the samples. Finally, we can compare the results of P-Net and the results achieved from the methods assessed in the reference papers.\\
We show also some ways to visualise the graphs of patients, either from the similarity matrix \textit{\textbf{W}} or from the kernel matrix \textbf{\textit{K}}. 
Indeed, the graph structure can be used to obtain visual clues about the relationships between patients and the relationships between patients and scores (e.g. visualization on the same graph of the \textit{true label} of each patient and the scores predicted by P-Net).

The above-mentioned experimental procedure provided some interesting results that we can summarize as follows:

\begin{enumerate}
\item In the Pancreatic cancer dataset, P-Net achieved competitive results with respect to NetRank, which is the best method assessed in the reference paper \cite{winter2012google}, for almost all the training set dimensions evaluated. The highest accuracies are achieved always using the \textit{8-step Random Walk Kernel} and the \textit{Nearest Neighbour kernelized score function}. We also assessed the impact of the number of features selected in the feature selection step, carried out by \textit{Welch's t-test}, on the performances of P-Net. We discovered that the feature selection step has a significant impact on the accuracy when we consider big training set dimensions and it leads to an important increase in the performances of the algorithm. 

\item In the Melanoma and Ovarian cancer datasets, P-Net with the application of \textit{moderated t-statistic} as feature selection method achieves the best results using the \textit{1-step Random Walk Kernel} and different kernelized score functions. In the Melanoma dataset, the best kernelized score function is the k-Nearest Neighbour with a cross-validation error rate of 39.11\% and in the Ovarian cancer dataset, the best result is 30.38\% obtained with the application of the Differential score. The comparison with the results of the reference paper \cite{barter2014network} (which assesses the performance of three different classifiers (RF, SVM and DLDA) combined with the use of \textit{moderated t-statistic}) shows that P-Net achieves the lowest error rate on the Ovarian cancer dataset and the same error rate of the SVM classifier on the Melanoma dataset.\\
Moreover, from the analysis of the class-specific error rate we discovered that P-Net is able to achieve comparable or better results with respect to the baseline methods on the individual classes GP and PP. In the Ovarian cancer dataset, it outperforms all the other methods with a decreasing of the error rate of 10-12\% (depending on the specific considered classifier) on the GP class.\\
From the accuracy analysis at a patient-level, we can see that P-Net can improve the accuracy for some patients that are generally difficult to classify by the other methods. This is an advantage because we can think to use P-Net combined with other outcome prediction methods to improve the classification performances.\\
Also on these datasets, we evaluated the impact of the number of selected features on the results of P-Net and we discovered that the selection of the first 1000 features with the smallest p-value is a reasonable choice. 

\item In the Pancreatic cancer dataset the increasing of the number of steps \textit{p} used in the \textit{p-step Random Walk Kernel} leads to improved accuracy values until we achieve a convergence point, which we found is between $p=10$ and $p=15$. After this point, there is no improvement in the algorithm performances even if we increase the value of \textit{p}. This is an opposite trend with respect to the results achieved by P-Net in the Melanoma and Ovarian cancer datasets. Indeed, we can observe decreasing performances with the increasing of the value of \textit{p} in these datasets. This observation is explained considering the different topology of the corresponding patient networks.\\
A tuning of the number of steps \textit{p} of the Random Walk Kernel is necessary every time we apply P-Net on a new dataset. Moreover, it is also necessary to carry out some experiments to discover which is the proper \textit{kernelized score function} to be used.  

\item The visualisation of the graph of patients gave us the possibility to better understand how P-Net works. For example, we can make some hypothesis about the score assigned to each node/patient and we can see how the weight of the edges changes after the application of a kernel method. This is useful to understand how the kernel methods exploit the topology of the network leading to better performances of the algorithm. We can also visualise if in the network there are clusters of nodes, involving patients more similar each other, and if these clusters are related to the phenotype/outcome of the patients. Moreover, we can visualise the graph of patients with the predicted scores and the true labels, which can be exploited to unravel the relationship between scores and labels. 
\end{enumerate}

From these experimental results, we can conclude that P-Net is a competitive method with respect to the approaches commonly used by state-of-the-art methods. It has also the advantage to provide a visualisation of the graph of patients which can be exploited to better understand the behaviour of the algorithm and to gain further information about phenotypic and/or genetic characteristics of the patients (depending on the type of data used to build the graph). 

\paragraph{Future perspectives}

We plan to further analyse P-Net and unravel all its features and potentiality. More precisely, we want to apply our new algorithm to different datasets to definitely assess the performances of P-Net. This is necessary to prove that the algorithm can achieve good performance in general and not only on the datasets selected in this work. Moreover, we want to further compare P-Net with other state-of-the-art methods to further evaluate its effectiveness in the current scenario of phenotype and outcome prediction.\\ 
Another possible future perspective of this research is related to the use of different types of biological data. In our work, we exploited only gene expression data obtained by microarray technology but there are several types of data we could use (e.g. allelic configuration of SNPs, clinical data, miRNA). Moreover, it has been shown that the integration of different kinds of data produces synergies between data that can improve the prediction performances~\cite{Vale12a}. So, another possible research path can involve the application or development of data integration systems, to further improve the effectiveness of the proposed algorithms.\\ 
In this work, we presented some ideas to visualise and exploit the graph of patients but we did not extensively analyse this problem. We believe that the graph of patients, provided by P-Net, can be really useful to obtain visual clues about the relationships between patients and to further analyse their phenotypic and genetic characteristics. A more extensive work is required to fully exploit the network representation of the patients and we plan to carry out it in the future. For example, we could try to use the labelled graph of patients to make further analysis using e.g. supervised and unsupervised clustering methods. In this way, we can discover novel significant grouping of patients that could be explained through the quantitative analysis of the features used to obtain the network.

\backmatter

\chapter*{List of abbreviations} \label{abbreviations}
\addcontentsline{toc}{chapter}{List of abbreviations}
\markboth{\MakeUppercase{List of abbreviations}}{\MakeUppercase{List of abbreviations}}
\begin{acronym}
\setlength{\parskip}{0ex}
\setlength{\itemsep}{1ex}

\acro{ANN}{Artificial Neural Networks}
\acro{AUC}{Area Under the Curve}
\acro{BSS/WSS}{Between-to-Within Sum-of-Squares method}
\acro{CNA}{Copy Number Alteration}
\acro{CNV}{Copy Number Variation}
\acro{CV}{Cross-Validation}
\acro{DIFF}{Differential score}
\acro{DLDA}{Diagonal Linear Discriminant Analysis}
\acro{DNORM}{Differential Normalized score}
\acro{DT}{Decision Tree}
\acro{EAV}{Empirical Average score}
\acro{GEO}{Gene Expression Omnibus}
\acro{GP}{Good Prognosis}
\acro{IDs}{Identifiers}
\acro{k-NN}{k-Nearest Neighbour score}
\acro{LDS}{Low Density Separation}
\acro{LOO}{leave-one-out}
\acro{MCCV}{Monte Carlo cross-validation}
\acro{MedExp}{Median Expression}
\acro{miRNA}{microRNA}
\acro{Mod-t}{Moderated t-statistic}
\acro{NN}{Nearest Neighbour score}
\acro{PCC}{Pearson product-moment correlation coefficient}
\acro{P-Net}{PatientNet: Network-based ranking of patients with respect to a given phenotype}
\acro{PPI}{protein-protein interaction}
\acro{PP}{Poor Prognosis}
\acro{RF}{Random Forest}
\acro{RMA}{Robust Multi-Array average expression measure}
\acro{RWK}{Random Walk Kernel}
\acro{SAM}{Significance Analysis of Microarray}
\acro{SEM}{Standard Error of the Mean}
\acro{SNF}{Similarity Network Fusion}
\acro{SNPs}{Single Nucleotide Polymorphisms}
\acro{SSL}{Semi-Supervised Learning}
\acro{SVM}{Support Vector Machine}
\acro{TOT}{Total score}
\acro{t-SNE}{t-Distributed Stochastic Neighbour Embedding}

\end{acronym}


\bibliographystyle{plain}
\bibliography{Bibliography} \label{bib}
\cleardoublepage

\end{document}